\documentclass[10pt,letterpaper]{article}

\usepackage{iftex}
\usepackage{amsmath}
\ifXeTeX
  \usepackage{newtxmath}
  \usepackage[no-math]{fontspec}
\else
  \usepackage[T1]{fontenc}
  \usepackage[utf8]{inputenc}
  \usepackage{newtxtext,newtxmath}
  
\fi
\usepackage[letterpaper,textwidth=5.5in,textheight=9in,centering,headheight=14pt]{geometry}
\usepackage{microtype}
\usepackage{graphicx}
\usepackage{booktabs,array,tabularx,longtable,multirow,makecell}
\usepackage{enumitem}
\usepackage[table]{xcolor}
\usepackage{pifont}
\usepackage{tikz}
\usetikzlibrary{arrows.meta,positioning,fit,calc,shapes.geometric,backgrounds}
\usepackage{pgfplots}
\pgfplotsset{compat=1.18}
\usepackage{algorithm}
\usepackage[noend]{algpseudocode}
\usepackage{caption}
\usepackage{subcaption}
\usepackage[numbers,sort&compress]{natbib}
\usepackage{xurl}
\usepackage{hyperref}
\usepackage[capitalise,noabbrev]{cleveref}
\usepackage{fancyhdr}

\definecolor{vizBlue}{HTML}{2A78D6}
\definecolor{vizOrange}{HTML}{EB6834}
\definecolor{vizAqua}{HTML}{1BAF7A}
\definecolor{vizYellow}{HTML}{EDA100}
\definecolor{vizInk}{HTML}{0B0B0B}
\definecolor{vizInkTwo}{HTML}{52514E}
\definecolor{vizMuted}{HTML}{898781}
\definecolor{vizGrid}{HTML}{E1E0D9}
\definecolor{vizAxis}{HTML}{C3C2B7}
\definecolor{vizSurface}{HTML}{FCFCFB}
\definecolor{vizBlueLight}{HTML}{CDE2FB}
\definecolor{vizBlueMid}{HTML}{86B6EF}
\definecolor{vizBlueDark}{HTML}{184F95}
\definecolor{headerGreen}{HTML}{17824A}

\fancypagestyle{plain}{%
  \fancyhf{}%
  \fancyhead[R]{\small\href{https://jeremycanale.com}{\textcolor{headerGreen}{jeremycanale.com}}}%
  \fancyfoot[C]{\thepage}%
  }
\pgfplotsset{vizaxis/.style={
    axis lines=left, axis line style={vizAxis}, tick style={vizAxis},
    grid=major, grid style={vizGrid, line width=0.4pt, solid},
    tick label style={font=\small, text=vizInkTwo},
    label style={font=\small, text=vizInk},
    legend style={font=\small, draw=none, fill=white, fill opacity=0.9, text opacity=1, text=vizInk},
    legend cell align=left,
    every axis plot/.append style={line width=1.4pt, line join=round, line cap=round},
    clip=false}}
\tikzset{
  viznode/.style={draw=vizAxis, rounded corners=2pt, align=flush center, font=\small, text=vizInk, inner sep=4pt, fill=white},
  vizarrow/.style={-{Stealth[length=2mm]}, semithick, draw=vizInkTwo},
  vizlabel/.style={font=\scriptsize, text=vizInkTwo, align=flush center}}

\newcommand{\cmark}{\ding{51}}
\newcommand{\xmark}{\ding{55}}
\newcommand{\AD}{Agent Detector}
\newcommand{\code}[1]{\texttt{\small #1}}
\setlist{itemsep=2pt,topsep=2pt,parsep=0pt}
\makeatletter
\renewcommand{\section}{\@startsection{section}{1}{\z@}%
  {-2.0ex \@plus -0.5ex \@minus -0.2ex}{1.5ex \@plus 0.3ex \@minus 0.2ex}{\large\bfseries\raggedright}}
\renewcommand{\subsection}{\@startsection{subsection}{2}{\z@}%
  {-1.8ex \@plus -0.5ex \@minus -0.2ex}{0.8ex \@plus 0.2ex}{\normalsize\bfseries\raggedright}}
\renewcommand{\subsubsection}{\@startsection{subsubsection}{3}{\z@}%
  {-1.5ex \@plus -0.5ex \@minus -0.2ex}{0.5ex \@plus 0.2ex}{\normalsize\bfseries\raggedright}}
\renewcommand{\paragraph}{\@startsection{paragraph}{4}{\z@}%
  {1.5ex \@plus 0.5ex \@minus 0.2ex}{-1em}{\normalsize\bfseries}}
\renewcommand{\maketitle}{%
  \thispagestyle{plain}%
  \vbox{\hsize\textwidth \linewidth\hsize
    \vskip 0.1in
    \hrule height 4pt
    \vskip 0.25in
    \centering
    {\LARGE\bfseries \@title\par}
    \vskip 0.22in
    \hrule height 1pt
    \vskip 0.3in
    {\def\and{\end{tabular}\hskip 3em\begin{tabular}[t]{c}}%
     \begin{tabular}[t]{c}\@author\end{tabular}\par}
    \vskip 0.3in}}
\renewenvironment{abstract}{%
  \centerline{\large\bfseries Abstract}%
  \vspace{0.5ex}%
  \begin{quote}}{\par\end{quote}\vskip 1ex}
\makeatother
\hypersetup{colorlinks=true, linkcolor=vizBlueDark, citecolor=vizBlueDark,
  urlcolor=vizBlueDark}

\IfFileExists{eval/results/tex/macros.tex}{
\newcommand{\StdAgents}{12}
\newcommand{\StdComponents}{3}
\newcommand{\StdBenign}{9}

\newcommand{\StdRuns}{5}

\newcommand{\StdDOneFp}{4}

\newcommand{\StdDOnePrecision}{0.69}

\newcommand{\StdDOneFOne}{0.72}

\newcommand{\StdDTwoFp}{4}

\newcommand{\StdDTwoPrecision}{0.71}

\newcommand{\StdDTwoFOne}{0.77}

\newcommand{\StdADTp}{12}

\newcommand{\StdADFp}{0}

\newcommand{\StdADPrecision}{1.00}
\newcommand{\StdADRecall}{1.00}
\newcommand{\StdADFOne}{1.00}

\newcommand{\StdADComponents}{3}

\newcommand{\StdOsEdition}{Home}
\newcommand{\StdOsVersion}{26H2}
\newcommand{\StdOsBuild}{26300.9711}

\newcommand{\StdBinarySha}{396416552dca}
\newcommand{\StdToolVersion}{0.3.2}
\newcommand{\StdCatalogVersion}{2026.11.09}
\newcommand{\StdRunFirst}{20261109-132251}
\newcommand{\StdRunLast}{20261109-142759}
\newcommand{\StdDurationmsMedian}{154}
\newcommand{\StdDurationmsPNineFive}{166}

\newcommand{\StdWallmsMedian}{177}

\newcommand{\StdCputimemsMedian}{406}

\newcommand{\StdPeakworkingsetmbMedian}{22.3}

\newcommand{\StdProcessesMedian}{352}

\newcommand{\StdConnectionsMedian}{62}

\newcommand{\StdOverheadRuns}{100}
\newcommand{\StdInsituCorrect}{9}
\newcommand{\StdInsituTotal}{9}
\newcommand{\StdTtlOpenai}{60}

\newcommand{\StdTtlLastMaxOpenai}{55}

\newcommand{\StdTtlMistral}{300}

\newcommand{\StdTtlLastOverall}{295}
\newcommand{\StdTtlAttributedAll}{168}
\newcommand{\StdTtlScansAll}{288}
\newcommand{\StdTtlReconnectAll}{25}
\newcommand{\StdTtlReconnectOfAll}{31}
\newcommand{\StdIntScans}{40}
\newcommand{\StdIntConnected}{18}

\newcommand{\StdIntProbableagent}{4}

\newcommand{\StdIntActionable}{15}
\newcommand{\StdIntActionableRatio}{0.38}
\newcommand{\StdIntActionableLo}{0.23}
\newcommand{\StdIntActionableHi}{0.54}

\newcommand{\AdmAgents}{12}
\newcommand{\AdmComponents}{3}

\newcommand{\AdmRuns}{5}

\newcommand{\AdmADTp}{11}

\newcommand{\AdmADFp}{0}

\newcommand{\AdmADPrecision}{1.00}
\newcommand{\AdmADRecall}{0.92}

\newcommand{\AdmADComponents}{3}

\newcommand{\AdmRunFirst}{20261109-112950}
\newcommand{\AdmRunLast}{20261109-123502}
\newcommand{\AdmDurationmsMedian}{168}
\newcommand{\AdmDurationmsPNineFive}{174}

\newcommand{\AdmCputimemsMedian}{469}

\newcommand{\AdmPeakworkingsetmbMedian}{22.5}

}{}

\title{Detecting Shadow AI: Unsanctioned LLM Agents\\
  on Enterprise Endpoints\\[6pt]
  {\Large\mdseries A Taxonomy, an Open Detector and a Controlled Evaluation}}
\author{\textbf{Jeremy Canale}\\
  \texttt{contact@jeremycanale.com}
  \and
  \textbf{Fabien Soulis}\\
  \texttt{soulis.fabien@gmail.com}}
\hypersetup{
  pdftitle={Detecting Shadow AI: Unsanctioned LLM Agents on Enterprise Endpoints. A Taxonomy, an Open Detector and a Controlled Evaluation},
  pdfauthor={Jeremy Canale, Fabien Soulis}}

\begin{document}
\maketitle

\begin{abstract}
Shadow IT, the use of technology that the IT organisation has not approved,
is a long-standing governance problem. We argue that autonomous AI agents can
make it grow much faster: agents make building tools nearly free, act with
their user's privileges, and leave behind scripts, automations and MCP servers
that are themselves unsanctioned. A simple compounding model shows that,
under its assumptions, shadow IT grows exponentially once agentic artifacts
reproduce faster than the organisation retires them; retiring them starts
with detecting them. We then ask whether unsanctioned agents can be detected
on Windows endpoints. We propose a taxonomy of nine classes of agents and six
layers of endpoint footprint, and argue that rules based only on the
structure of processes and connections cannot separate an agent from its
\emph{behavioural twin}, the same program talking to a non-AI service. We
present \AD{}, an open-source Windows detector that anchors behaviour to the
identity of the remote service and correlates it with configuration,
installation and execution history, using a catalog of 39 tools and 184 LLM
endpoint entries; it informs users through a notice they must acknowledge
instead of blocking. On a controlled bench of \StdAgents{} synthetic agents,
\StdComponents{} AI components and \StdBenign{} benign workloads, repeated
\StdRuns{} times on one workstation, two behavioural rules flagged every
behavioural twin, while \AD{} flagged none and detected every agent in the
majority of runs. Its limits are temporal: a connection to a provider behind
a shared content-delivery network is attributable only while the DNS cache
holds the provider's name, one to five minutes, and an intermittent agent was
seen by about one scan in three. A scan costs about
\StdDurationmsMedian{}\,ms. The bench is too small to rank detectors; it maps
what each kind of evidence can see: unsanctioned agents can be partly
detected, and not by behaviour alone.
\end{abstract}

\section{Introduction}\label{sec:intro}

\paragraph{Shadow IT.}
Every organisation runs software that its IT department never approved. The
phenomenon has a name, \emph{shadow IT}: ``all hardware, software, or any
other solutions used by employees inside of the organisational ecosystem which
have not received any formal IT department approval''~\cite{silic2014shadow}.
It is old, it is mostly well-intentioned, since employees adopt tools to get
their work done, and it is persistently underestimated: a systematic review of
the literature concludes that ``the true extent of Shadow IT usage in
companies is vastly underestimated by CIOs''~\cite{klotz2019causing}. Its
risks are those of any system nobody has listed: data held by unknown
providers, security controls that do not apply, compliance obligations that
cannot be met. Its governance starts with finding
it~\cite{furstenau2021leaving}.

\paragraph{Shadow AI.}
Generative AI produced the fastest wave of shadow IT so far. In 2024, 75\% of
knowledge workers used generative AI at work and ``78\% of AI users are
bringing their own AI tools to work''~\cite{microsoft2024worktrend}; in a
2025 global study, 44\% of employees admitted using AI in ways that contravene
their organisation's policies and 61\% avoided revealing their
use~\cite{gillespie2025trust}. Among the breached organisations that IBM
studied, one in five reported a breach involving shadow AI, and only 37\% had
policies to manage AI or detect shadow AI~\cite{ibm2025breach}. This first wave was conversational: employees pasted
text into a chatbot, in a browser, where network controls could see it.

\paragraph{Agents change the rate, not just the scale.}
The second wave is made of \emph{agents}: software in which a language model,
in a loop, decides on actions that tools then execute on the user's machine,
with the user's privileges. Coding agents such as Claude Code, Codex and
GitHub Copilot's agent mode, desktop agents, AI-native IDEs, local model
runtimes and home-made scripts now run on enterprise laptops, often outside
any inventory~\cite{silva2026agentic,snyk2026volume2}. We argue that this wave
does not merely add a category to shadow IT; it changes how fast shadow IT
grows. Agents make building a tool nearly free; they act without asking; and
the scripts, automations, packages and MCP servers they leave behind are
themselves unsanctioned, and sometimes agentic in turn. We make the argument
explicit with a compounding model in which, under stated assumptions, shadow
IT grows exponentially as soon as agentic artifacts reproduce faster than the
organisation retires them (\cref{sec:acceleration}). The model does not
measure anything; it shows which quantities matter. Since deliberately
retiring an artifact requires finding it first, detection is a precondition
of governance, though not a sufficient one.

\paragraph{Why detecting agents is hard.}
Detecting agents is harder than detecting the chatbots of the first wave.
The network sees an encrypted connection to an API, often through a
content-delivery network shared with unrelated services, sometimes through an
aggregator or a relay with an innocuous name; it does not see which process
opened it, it does not see laptops outside the corporate network, and it does
not see local models at all. Inventories see installed products, but a
script of a few dozen lines has no installer. Behavioural detection, the usual answer
to software without a signature, looks promising: an agent holds a connection
to a remote service and runs commands. But so does \code{npm install}, so does
an IDE with an integrated terminal, so does an administrator's automation
script. We argue that for every agent there is a \emph{behavioural twin}, the
same program talking to a non-AI service, that a detector using only the
structure of processes and connections cannot separate from it, and we
measure the consequence for two simple behavioural rules
(\cref{sec:behavioral,sec:evaluation}). What separates an agent from its twin
is the identity of the remote service and of the software.

\paragraph{Approach.}
We therefore detect agents on the endpoint by \emph{correlating} weak signals
around identity: which process talks to which LLM service (attributed through
the Windows DNS cache or through address ranges dedicated to a provider),
whether that process also runs shells, what configuration it carries (base
URLs, MCP server definitions), what is installed, and what Windows remembers
having executed. We implement this in \AD{}, an open-source Windows detector
deployed with Microsoft Intune, which runs every hour as SYSTEM, reads only
local state, never reads content, and never blocks. When it finds an agent, it
shows the user a notice that they must acknowledge, and it records the
acknowledgement: the user is informed rather than silently observed.

\paragraph{Contributions.}
This paper makes six contributions.

\begin{enumerate}[label=\textbf{C\arabic*.}]
  \item An argument, with a simple compounding model and the available
    evidence, that agents accelerate shadow IT through second-order
    artifacts, and that detection is a precondition, though not a
    sufficient one, for keeping an organisation below the threshold of
    exponential growth (\cref{sec:acceleration}).
  \item An analysis of why behaviour-only detection of agents falls short,
    centred on the behavioural-twin argument, with an operational boundary
    between behaviour and identity (\cref{sec:behavioral}).
  \item A taxonomy of nine classes of agents and six layers of endpoint
    footprint, arguing that product-less agents are visible only at run
    time, in the network and in configuration (\cref{sec:taxonomy}).
  \item An open, dated catalog of 39 tool signatures and 184 LLM endpoint
    entries in nine categories, including aggregators, inference platforms,
    gateways and relays (\cref{sec:taxonomy}, \cref{app:catalog}).
  \item The design and implementation of \AD{}, which correlates these layers,
    protects privacy by design, informs users instead of blocking, and deploys
    with standard enterprise tooling
    (\cref{sec:design,sec:implementation}).
  \item A controlled evaluation with synthetic agents in three languages,
    behavioural twins, AI components, developer workloads and an in-situ
    cross-check, repeated \StdRuns{} times as standard user and as
    administrator against a signature baseline and two behavioural
    baselines, with two experiments on time: the expiry of DNS cache entries
    and intermittent agents (\cref{sec:evaluation}). The bench is small and
    run on one workstation; it shows where each kind of evidence fails
    rather than ranking detectors. The behavioural rules flag every twin;
    \AD{} separates agents from their twins, but attributes connections to
    providers behind a shared content-delivery network only while the DNS
    cache holds their names, sees an intermittent agent at about one scan in
    three, and misses ephemeral agents, undeclared relays and agents inside
    WSL~2; a scan takes about \StdDurationmsMedian{}\,ms.
\end{enumerate}

\paragraph{Organisation.}
\Cref{sec:shadow} reviews shadow IT and shadow AI, and \cref{sec:agents}
introduces agents and their ecosystem. \Cref{sec:acceleration} develops the
acceleration argument. \Cref{sec:threat} states the threat model and the
requirements. \Cref{sec:behavioral} explains why behaviour is not enough.
\Cref{sec:taxonomy} presents the taxonomy and the catalog, and
\cref{sec:landscape} the landscape of existing approaches.
\Cref{sec:design,sec:implementation} describe \AD{}, and
\cref{sec:evaluation} evaluates it. \Cref{sec:discussion} discusses the
inform-not-block design, the legal framework and the limitations;
\cref{sec:related} reviews related work and \cref{sec:conclusion} concludes.

\section{Shadow IT and Shadow AI}\label{sec:shadow}

\subsection{Shadow IT: an old problem}

Shadow IT is as old as end-user computing. Silic and Back define it as ``all
hardware, software, or any other solutions used by employees inside of the
organisational ecosystem which have not received any formal IT department
approval''~\cite{silic2014shadow}; earlier work describes it as the
``user-driven fulfilment of requirements'' and calls it ``a widespread
challenge amongst IT departments''~\cite{gyory2012exploring}. The
information-systems literature has studied its nature, its management and its
outcomes for more than a decade~\cite{haag2017shadow,zimmermann2017multiple}.
A systematic review of 107 studies characterises it by its covertness, ``i.e.,
without alignment with the IT organization'', and concludes that ``the true
extent of Shadow IT usage in companies is vastly underestimated by
CIOs''~\cite{klotz2019causing}.

The causes are well understood: business units need tools faster than central
IT can provide them; cloud services, consumer applications and personal
devices made acquisition trivial; and the people who adopt them are rarely
malicious. The consequences are also well understood: data held by providers
that the organisation does not know, licences and costs outside any budget,
security controls that do not apply, and compliance obligations that cannot be
met for systems no one has listed. Analysts have long expected the share of
employees who ``acquire, modify or create technology outside IT's visibility''
to keep growing, from 41\% in 2022 to 75\% by 2027 according to a Gartner
prediction reported in the press~\cite{nation2023gartner}.

Two findings from this literature frame our work. The first is that shadow IT
is not only a risk: the governance literature distinguishes control-oriented,
user-oriented and user-driven approaches, and treats user-driven innovation as
something to channel rather than suppress~\cite{gyory2012exploring}. The second
is that every governance outcome starts with identification. A study of 27
shadow systems traces what happens once they are found: they are phased out,
replaced, or continued as IT-managed or business-managed
systems~\cite{furstenau2021leaving}. None of these outcomes is available for a
system that has not been found.

\subsection{Shadow AI}

Generative AI produced the most rapid wave of shadow IT so far. Silic et~al.\
define shadow AI as ``the unsanctioned use of AI systems outside approved
governance frameworks'' and frame it as a ``sociotechnical governance
failure''~\cite{silic2025shadowai}; recent work documents it in critical
infrastructure~\cite{baruwalchhetri2026frontier} and proposes maturity models
for ``bring your own AI''~\cite{bello2026governing}.

Survey data show both its scale and its covertness. In Microsoft and
LinkedIn's 2024 study of 31{,}000 people, 75\% of knowledge workers used
generative AI at work, ``78\% of AI users are bringing their own AI tools to
work'', and 52\% of those who used AI at work were reluctant to admit it for
their most important tasks~\cite{microsoft2024worktrend}. A 2025 global study
of 48{,}000 respondents found that 58\% of employees regularly used AI tools
for work, mostly ``free public options (70\%) rather than employer-provided
tools (42\%)''; 44\% admitted using AI in ways that contravene their
organisation's policies, and 61\% avoided revealing when they had used
it~\cite{gillespie2025trust}.

The risk is not hypothetical. In 2023, Samsung banned generative AI tools for
its staff after discovering that employees had uploaded sensitive source code
to a public chatbot~\cite{gurman2023samsung}. IBM's 2025 breach study reports
that one organisation in five suffered a breach due to shadow AI, that
organisations with high levels of shadow AI paid on average \$670{,}000 more per
breach, and that only 37\% had policies to manage AI or detect shadow
AI~\cite{ibm2025breach}. Prohibition alone does not solve the problem: a
parameterised model of bring-your-own-AI governance finds that
``prohibition-based solutions will result in residual risk levels close to
those achieved through baseline solutions''~\cite{bello2026governing}.

\subsection{From chatbots to agents}

The first wave of shadow AI was conversational: employees pasted text into a
web chatbot. That wave is visible where web traffic is visible, and it is the
target of the network and data-centric controls reviewed in
\cref{sec:landscape}. The second wave is different. Agents run on the
endpoint, act with the user's privileges, and talk to models through APIs,
aggregators, relays or local runtimes rather than through a web page.

Industry measurements of developer machines show how far this wave has already
gone. In nearly 10{,}000 developer environments, 50.8\% of developers had at
least one MCP server installed, 43\% ran two or more AI coding environments,
and MCP servers were found ``installed on developer machines outside the
centralized inventory''~\cite{silva2026agentic}. The same vendor reports that
security programs see ``roughly a third of their organization's real AI
footprint''~\cite{snyk2026volume2}. Security researchers describe personal
agents that ``can run shell commands, read and write files, and execute
scripts'', and warn that such agents ``can become covert data-leak channels
that bypass traditional data loss prevention''~\cite{chang2026openclaw}.

Shadow AI has therefore moved from the browser, where organisations already
look, to the endpoint, where they mostly do not. \Cref{sec:acceleration} argues
that this move does not just add one more category of shadow IT: it changes
the rate at which shadow IT grows.

\section{Autonomous LLM Agents}\label{sec:agents}

This section summarises what an AI agent is, how agents reached the
enterprise desktop, and the two ecosystem features that matter most for their
detection: the Model Context Protocol and the many routes from an agent to a
model.

\subsection{What an agent is}

A chatbot answers; an agent acts. The step from one to the other is a control
loop in which the model's output is parsed into actions, the actions are
executed by tools, and their results are fed back to the model. ReAct showed
that interleaving reasoning traces and actions lets a model ``interface with
external sources, such as knowledge bases or environments''~\cite{yao2023react};
Toolformer showed that a model can learn ``which APIs to call, when to call
them, what arguments to pass''~\cite{schick2023toolformer}. Surveys now
describe LLM-based agents through a common architecture of profile, memory,
planning and action modules~\cite{wang2024survey}.

For detection purposes we use a deliberately operational definition. An
\emph{agent} is software that (i) reaches a language model, through a remote
API or a local runtime; (ii) runs a loop that turns the model's output into
actions; and (iii) owns tools that act on its environment: a shell, the file
system, a browser, network services or MCP servers. This definition is
behavioural on purpose: it covers a commercial coding assistant as well as a
script of a few dozen lines. \AD{}'s \code{probable-agent} rule tests a
narrower proxy of it on the endpoint: a process that talks to a model
\emph{and} has a shell among its descendants. Agents whose tools run inside
their own process, such as file edits, HTTP calls or a browser driven over a
pipe, start no shell and escape that rule; other evidence must find them.

\subsection{Coding agents on the desktop}

Software engineering is where agents first became everyday tools. On
SWE-bench, a benchmark of 2{,}294 real GitHub issues, the best-performing
model at the time, Claude~2, solved ``a mere 1.96\% of the
issues''~\cite{jimenez2024swebench}; SWE-agent
reached 12.5\% by giving the model an agent--computer
interface~\cite{yang2024sweagent}, and open platforms such as OpenHands let
agents work ``by writing code, interacting with a command line, and browsing
the web''~\cite{wang2025openhands}. The tools followed. Claude Code, Codex,
GitHub Copilot's agent mode, Cursor and their peers now run on developers'
laptops with the developer's privileges.

Adoption data come from repositories and surveys. A study of 128{,}018 GitHub
projects estimates that between 22.2\% and 28.7\% of them already use coding
agents~\cite{robbes2026agentic}; a dataset of more than 456{,}000 pull
requests documents five agents operating ``in the wild''~\cite{li2025aiteammates};
and a study of pull requests generated with Claude Code reports that 83.8\%
of them were eventually merged~\cite{watanabe2025agenticprs}. The 2026 Stack
Overflow developer survey, with more than 30{,}000 respondents, names Claude
Code (66\%) and GitHub Copilot (59\%) as ``the leaders for coding
agents''~\cite{stackoverflow2026survey}. These
numbers describe developers, but the same tools, and general-purpose desktop
agents such as Claude Desktop's Cowork mode, are spreading to analysts,
marketers and managers.

Not every agent runs on the endpoint. GitHub's cloud coding agent, for
instance, edits files and runs tests ``in a cloud development environment,
rather than your local IDE''~\cite{github2026copilotcloud}. Such agents leave
no footprint on the laptop; we treat them as out of scope
(\cref{sec:threat}).

\subsection{The Model Context Protocol}

The Model Context Protocol (MCP), open-sourced by Anthropic in November
2024~\cite{anthropic2024mcp}, standardises how an agent (the \emph{host}) talks
to \emph{servers} that expose tools, resources and prompts. Messages are
JSON-RPC~2.0; with the \code{stdio} transport, a server is a
``client-launched subprocess'' that exchanges newline-delimited messages over
its standard streams~\cite{mcp2026specification}. The specification itself
warns that ``tools represent arbitrary code execution and must be treated with
appropriate caution'', and the documentation of local servers notes that a
server ``runs with your user account permissions''~\cite{mcp2026localservers}.

MCP grew fast. An analysis of 177{,}436 agent tools created between November
2024 and February 2026 finds that software development accounts for 67\% of
the tools and 90\% of MCP server downloads, and that the share of ``action''
tools, which change the state of the world, rose from 27\% to 65\% of
usage~\cite{stein2026mcptools}.

For an endpoint detector, MCP has two consequences. First, an agent at work is
often a \emph{process tree}: the host, its MCP servers started through
\code{npx}, \code{uvx} or \code{python}, and the shells and browsers they
drive. Second, MCP servers are declared in configuration files that live in
the user profile: \code{\%APPDATA\%\textbackslash Claude\textbackslash
claude\_desktop\_config.json} for Claude Desktop~\cite{mcp2026localservers},
\code{\textasciitilde/.claude.json} for Claude
Code~\cite{anthropic2026claudecodesettings},
\code{\textasciitilde/.codex/config.toml} for Codex~\cite{openai2026codexmcp},
\code{mcp.json} files for VS Code and Copilot~\cite{microsoft2026vscodemcp}, and
\code{\textasciitilde/.cursor/mcp.json} for Cursor~\cite{cursor2026mcp}. These
files are evidence that an agent is installed and configured, even when it is
not running. Many clients also read project-scoped files, such as an
\code{.mcp.json} or a \code{.vscode/mcp.json} inside a repository; the
evaluated catalog reads only the files in the user profile, so servers
declared per project escape it. They are also a target: MITRE ATLAS lists ``Discover AI Agent
Configuration'' and ``Modify AI Agent Configuration'' among adversary
techniques~\cite{mitre2026atlas}.

\subsection{Routes to a model}

An agent is only as visible as its route to a model. The routes have
multiplied:

\begin{itemize}
  \item \textbf{Model providers} expose first-party APIs (OpenAI, Anthropic,
    Google, Mistral) and are resold by cloud platforms such as Azure OpenAI or
    Amazon Bedrock under the customer's own subdomain.
  \item \textbf{Aggregators} such as OpenRouter resell hundreds of models from
    many providers behind a single, OpenAI-compatible API; the agent's traffic
    then goes to the aggregator, whatever the model.
  \item \textbf{Inference platforms} host open-weight models (Groq, Together
    AI, Fireworks AI and others).
  \item \textbf{Gateways and relays}, hosted (Cloudflare AI Gateway, Portkey)
    or self-hosted (LiteLLM), sit between the agent and the provider; relays
    re-expose a provider's API under an arbitrary domain.
  \item \textbf{Local runtimes} such as Ollama or LM Studio serve open-weight
    models on a local port, with no external traffic at all.
\end{itemize}

Most agents and SDKs let the user switch between these routes with a single
environment variable, such as \code{OPENAI\_BASE\_URL} or
\code{ANTHROPIC\_BASE\_URL}. That convenience makes the route itself
unreliable as a sole signal, but the variable that sets it is a signal in its
own right, which \AD{} reads (\cref{sec:design}).

\subsection{Agents as a security concern}

The security literature on agents focuses on what an attacker can make an
agent do: indirect prompt injection turns retrieved data into
instructions~\cite{greshake2023indirect}; benchmarks show tool-integrated
agents following injected instructions in a substantial share of
cases~\cite{zhan2024injecagent,debenedetti2024agentdojo}; poisoned MCP tool
descriptions hijack agents~\cite{beurerkellner2025toolpoisoning,wang2026mcptox};
and frameworks name the risk of ``excessive
agency''~\cite{owasp2024llmtop10,owasp2026llmtop10}, and a separate list now
covers agentic applications~\cite{owasp2025agentictop10}. All of
these presuppose that the organisation knows which agents run where. The
question of this paper comes first: \emph{which agents are there at all?}

\section{Why Agents Accelerate Shadow IT}\label{sec:acceleration}

This section argues that AI agents will make shadow IT grow faster than any
previous wave, and that under conditions we make explicit the growth becomes
exponential. The argument rests on three mechanisms, a simple model that shows
how they combine, and the available evidence. It is an argument, not a
measurement: we state the assumptions, and we mark which numbers are
illustrative.

\subsection{Three mechanisms}

\paragraph{1. Creating a tool becomes nearly free.}
Classic shadow IT is \emph{acquired}: an employee subscribes to a service or
installs an application that someone else built. Agents let employees
\emph{build}. Research has shown models that ``create their own reusable tools
for problem-solving'', with ``continual tool generation and caching as new
requests emerge''~\cite{cai2024toolmakers}. Practice has followed: more than
1.1 million public repositories use an LLM SDK, and GitHub's coding agent
created more than one million pull requests in its first five
months~\cite{github2025octoverse}. In McKinsey's 2026 survey, as reported in
the trade press, nearly a third of respondents said their organisation had
decided against buying a software product because it ``could be built
internally with agentic coding tools''~\cite{ring2026mckinsey}. Productivity
evidence is mixed for experienced developers working on mature code bases: a
randomised trial found that AI assistance \emph{increased} completion time by
19\%~\cite{becker2025measuring}, while an earlier controlled experiment found a
55.8\% speed-up on a well-defined task~\cite{peng2023impact}. What matters for
shadow IT, however, is not the speed of expert maintenance work but the cost
of producing a new, small, good-enough tool for oneself. For that, the
evidence points one way.

\paragraph{2. Agents act with the user's privileges.}
An agent does not wait for IT. It installs packages, starts MCP servers, writes
and schedules scripts, configures other tools and calls external services,
all with the rights of the user who runs it. The MCP documentation is explicit
that a local server ``runs with your user account
permissions''~\cite{mcp2026localservers}, and security frameworks list
``excessive agency'' among the main risks of LLM
applications~\cite{owasp2024llmtop10,owasp2026llmtop10}. Each of these actions
can create an artifact that no one approved.

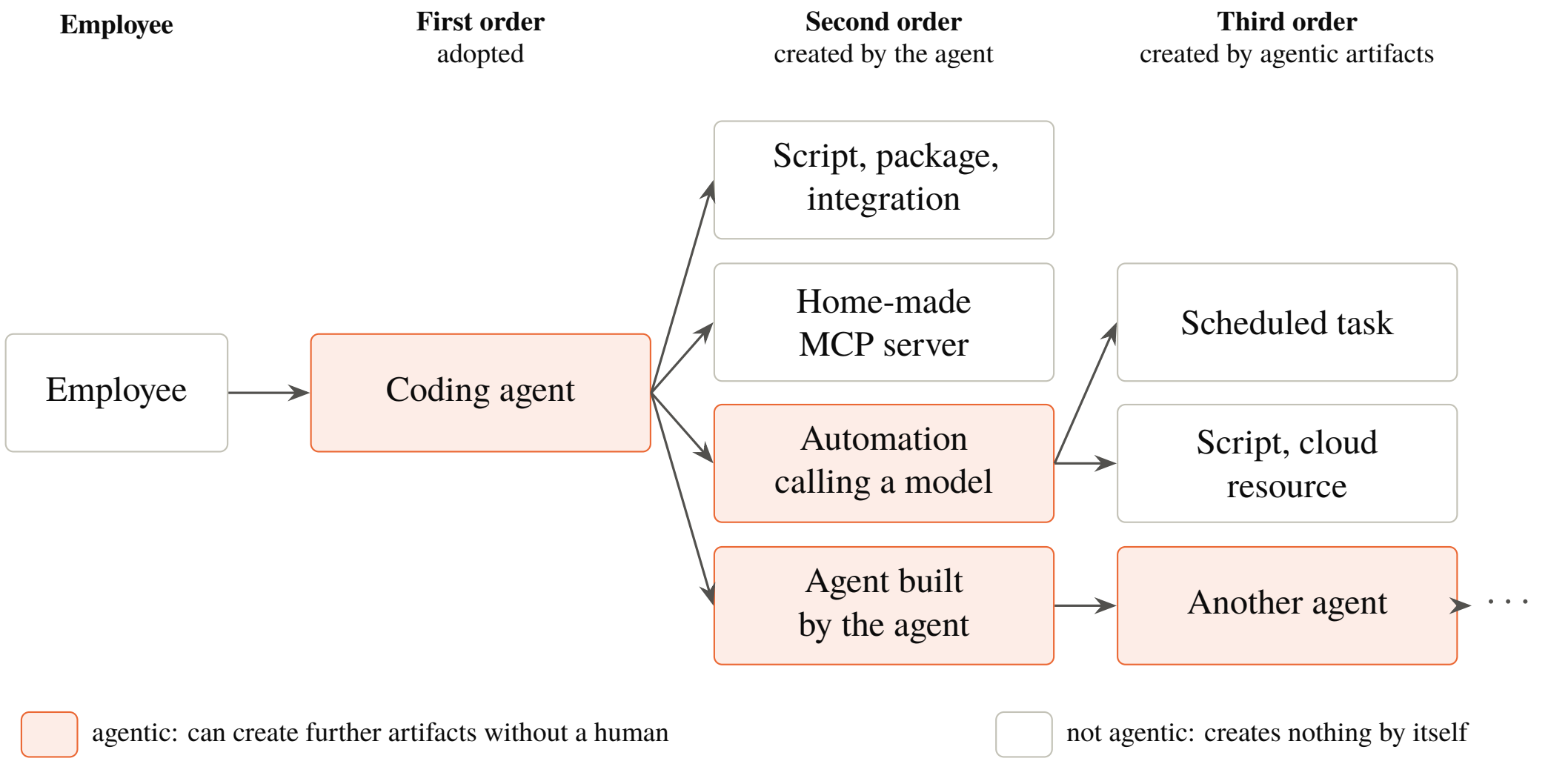
\begin{figure}[t]
  \centering
\begin{tikzpicture}[
    box/.style={viznode, text width=2.75cm, minimum height=10.5mm},
    person/.style={viznode, text width=1.7cm, minimum height=10.5mm},
    agentic/.style={box, fill=vizOrange!12, draw=vizOrange},
    plainbox/.style={box, fill=white, draw=vizAxis},
    head/.style={vizlabel, text=vizInk, font=\scriptsize}]
  \node[head] at (0, 3.15) {\textbf{Employee}\\\phantom{x}};
  \node[head] at (3.25, 3.15) {\textbf{First order}\\adopted};
  \node[head] at (6.85, 3.15) {\textbf{Second order}\\created by the agent};
  \node[head] at (10.45, 3.15) {\textbf{Third order}\\created by agentic artifacts};
  \node[person] (emp) at (0, 0) {Employee};
  \node[agentic] (ag) at (3.25, 0) {Coding agent};
  \node[plainbox] (s1) at (6.85, 1.9) {Script, package,\\integration};
  \node[plainbox] (s2) at (6.85, 0.63) {Home-made\\MCP server};
  \node[agentic] (s3) at (6.85, -0.63) {Automation\\calling a model};
  \node[agentic] (s4) at (6.85, -1.9) {Agent built\\by the agent};
  \node[plainbox] (t1) at (10.45, 0.63) {Scheduled task};
  \node[plainbox] (t2) at (10.45, -0.63) {Script, cloud\\resource};
  \node[agentic] (t3) at (10.45, -1.9) {Another agent};
  \node[vizlabel, font=\normalsize] (more) at (12.45, -1.9) {$\cdots$};
  \draw[vizarrow] (emp) -- (ag);
  \foreach \t in {s1, s2, s3, s4} \draw[vizarrow] (ag.east) -- (\t.west);
  \draw[vizarrow] (s3.east) -- (t1.west);
  \draw[vizarrow] (s3.east) -- (t2.west);
  \draw[vizarrow] (s4.east) -- (t3.west);
  \draw[vizarrow] (t3.east) -- (more.west);
  \node[agentic, text width=5mm, minimum height=4mm, inner sep=0pt] (la) at (-0.6, -3.05) {};
  \node[anchor=west, vizlabel, text=vizInk] at (la.east) {agentic: can create further artifacts without a human};
  \node[plainbox, text width=5mm, minimum height=4mm, inner sep=0pt] (lp) at (8.1, -3.05) {};
  \node[anchor=west, vizlabel, text=vizInk] at (lp.east) {not agentic: creates nothing by itself};
\end{tikzpicture}
  \caption{Orders of shadow IT. Classic shadow IT stops at the first order:
    a tool adopted by an employee. Agents add a second order, the artifacts
    they create, and a third, the artifacts created by those of them that are
    agentic themselves. Only the agentic branches can keep growing without a
    human.}
  \label{fig:orders}
\end{figure}

\paragraph{3. Agents produce shadow IT, and some of it is agentic.}
The artifacts an agent leaves behind are themselves shadow IT: scripts,
automations, integrations, MCP servers, packages, cloud resources. We call
them \emph{second-order} shadow IT. Some of them are agentic in turn, such as
an automation that calls a model or an agent built by an agent: they can
create further artifacts without a human in the loop. Others, such as a
home-made MCP server, are not agentic themselves but extend what the agents
that use them can do. The ecosystem data are consistent with this: half of the developers in a large
sample already had an MCP server installed~\cite{silva2026agentic}, and the
share of ``action'' tools in MCP usage rose from 27\% to 65\% in little more
than a year~\cite{stein2026mcptools}. \Cref{fig:orders} illustrates the
three orders.

\subsection{A compounding model}

The three mechanisms combine multiplicatively. Consider an organisation of
$N$ employees and let $A_t$ be the number of employees who use an
unapproved agent in month $t$. In \emph{classic} shadow IT, adopters acquire
unapproved tools at a rate $\kappa$ per month and the organisation retires a
share $\delta$ of them each month, because it finds and phases them out or
because they are abandoned. The stock $S_t$ of unapproved tools follows
\begin{equation}
  S_{t+1} = (1-\delta)\,S_t + \kappa\,A_t ,
\end{equation}
and converges to $\kappa N/\delta$ once adoption saturates: shadow IT grows
with the number of adopters, and stops.

With agents, two terms change. Agent users create artifacts at a rate $\rho$
that is much larger than $\kappa$ (mechanism~1), and a fraction $\phi$ of the
stock $X_t$ of artifacts is agentic and creates new artifacts at a rate
$\sigma$ (mechanism~3):
\begin{equation}
  X_{t+1} = (1-\delta+\sigma\phi)\,X_t + \rho\,A_t .
\end{equation}
Write $g = 1-\delta+\sigma\phi$. When $\sigma\phi < \delta$, $g<1$ and the
stock converges to $\rho N/(\delta-\sigma\phi)$: larger than in classic
shadow IT by a factor that grows as $\sigma\phi$ approaches $\delta$, but
bounded. When
\begin{equation}
  \sigma\phi > \delta ,
\end{equation}
$g>1$ and the stock grows exponentially, as $g^t$, whatever the number of
employees. The condition has a plain reading: \emph{shadow IT explodes when
agentic artifacts reproduce faster than the organisation retires them}.

\begin{figure}[t]
  \centering
  \begin{tikzpicture}
    \begin{semilogyaxis}[vizaxis, clip=true,
        width=0.74\linewidth, height=7.2cm,
        xlabel={Month}, ylabel={Unapproved artifacts per employee},
        xmin=0, xmax=48, ymin=0.05, ymax=100, xtick={0,12,24,36,48},
        legend columns=2,
        legend style={at={(0.5,-0.2)}, anchor=north,
          /tikz/every even column/.append style={column sep=12pt}}]
      \addplot[vizMuted] table[x=month, y=classic] {eval/results/tex/acceleration.dat}
        coordinate[pos=1] (endClassic);
      \addlegendentry{Classic shadow IT ($\kappa=0.05$)}
      \addplot[vizBlue] table[x=month, y=subcritical] {eval/results/tex/acceleration.dat}
        coordinate[pos=1] (endSub);
      \addlegendentry{Agents, $\sigma\phi=0.03<\delta=0.05$}
      \addplot[vizOrange] table[x=month, y=supercritical] {eval/results/tex/acceleration.dat}
        coordinate[pos=1] (endSuper);
      \addlegendentry{Agents, $\sigma\phi=0.10>\delta=0.05$}
      \addplot[vizAqua] table[x=month, y=detected] {eval/results/tex/acceleration.dat}
        coordinate[pos=1] (endDetected);
      \addlegendentry{Agents, $\sigma\phi=0.10$, detection: $\delta=0.15$}
    \end{semilogyaxis}
    \foreach \c/\t in {endClassic/Classic, endSub/Sub-critical,
        endSuper/Super-critical, endDetected/With detection}
      \node[anchor=west, font=\scriptsize, text=vizInkTwo, xshift=2pt] at (\c) {\t};
  \end{tikzpicture}
  \caption{Stock of unapproved artifacts per employee under the model of
    \cref{sec:acceleration}, with adoption $A_t/N = 1-e^{-0.1t}$ and agent
    users creating $\rho=0.5$ artifacts per month. All parameter values are
    \emph{illustrative}, chosen to show the regimes, not fitted to data
    (logarithmic scale).}
  \label{fig:acceleration}
\end{figure}
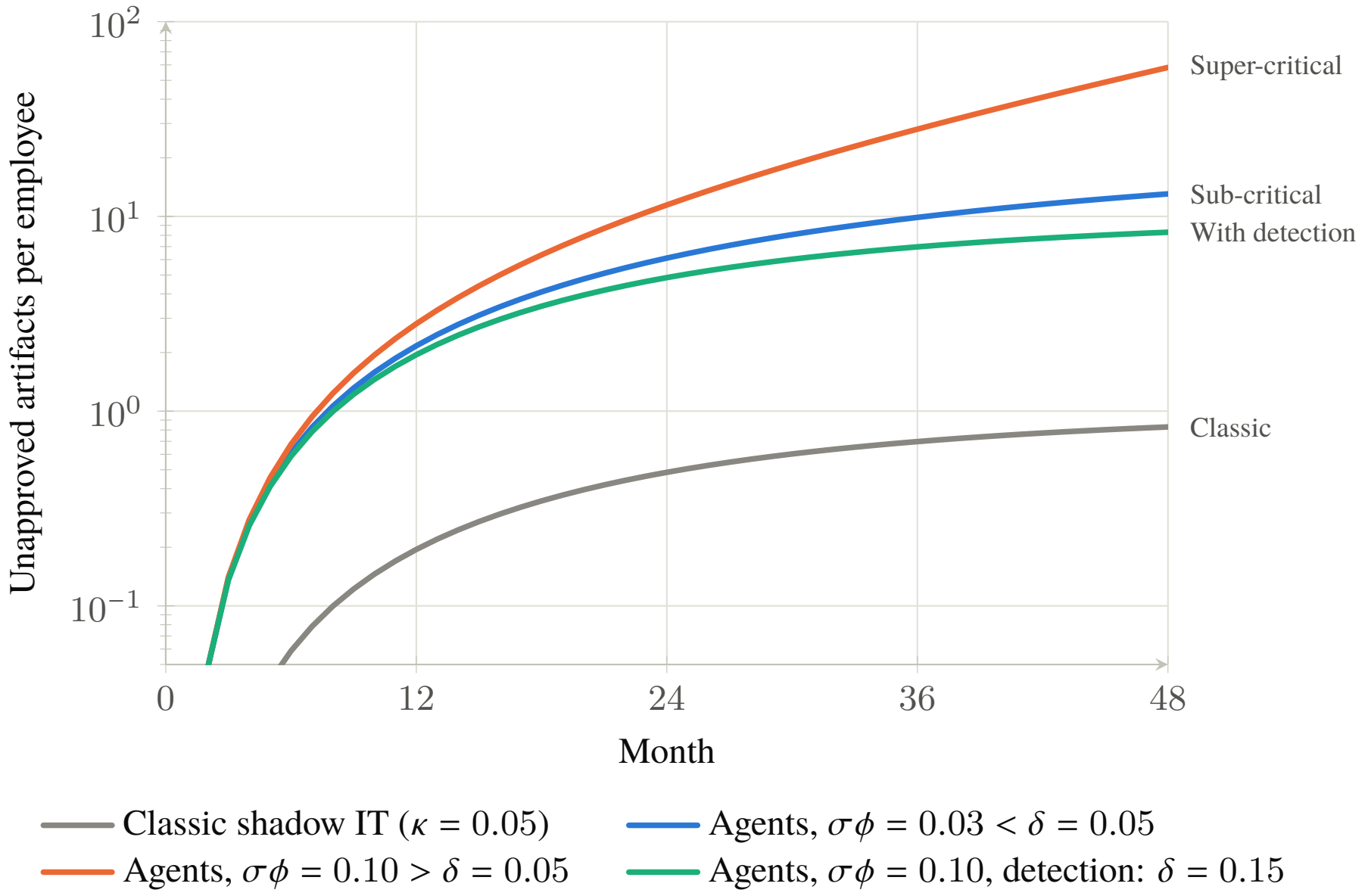

\Cref{fig:acceleration} plots the four regimes with illustrative parameters.
The point of the figure is qualitative, not numerical: the classic curve
saturates; the agentic curves start higher because $\rho\gg\kappa$; the
super-critical curve diverges; and raising the retirement rate $\delta$ above
$\sigma\phi$ brings it back to a bounded regime.

\subsection{What the model says about detection}

The organisation has several levers. Sanctioned alternatives lower adoption
$A_t$; application control, least privilege and allowlists of MCP servers
lower the creation rates $\rho$ and $\sigma$; a policy on automations lowers
the agentic share $\phi$; and retirement raises $\delta$. Detection does not
act on any of these parameters by itself. It is the precondition of the last
lever and the measurement that the others need. Abandonment retires
artifacts without anyone finding them, but every deliberate retirement starts
with identification: an artifact that is not found cannot be phased out,
replaced or brought under management~\cite{furstenau2021leaving}. Today most
of it is not found: a vendor study of its own discovery customers reports that
security programs see roughly a third of the real AI
footprint~\cite{snyk2026volume2}, and most breached organisations in IBM's
sample had no policy to detect shadow AI~\cite{ibm2025breach}. With invisible
artifacts, the deliberate part of $\delta$ is close to zero.

Detection is therefore necessary but not sufficient. In \cref{fig:acceleration},
raising $\delta$ from 0.05 to 0.15 bounds the super-critical curve, but its
stock still settles at $\rho/(\delta-\sigma\phi)=10$ artifacts per
employee, ten times the classic level: retirement alone does not bring shadow
IT back to its pre-agent size, and the other levers remain necessary. We
therefore treat agent detection as the first step of agent governance, not as
its whole.

\subsection{Evidence and caveats}

Four caveats apply. First, the model is linear: once $g>1$, exponential
growth follows by construction, and a real organisation would hit limits
that the model omits (the attention of its employees, compute budgets, the
saturation of useful tasks). The model makes the threshold explicit; it does
not predict growth. Second, the parameters are not known; we do not claim
that any organisation is above the threshold today. In particular, we know
of no measurement of $\sigma$, the rate at which agentic artifacts create
further artifacts without a human in the loop: the ecosystem data cited
above show that agents increasingly act, not that their artifacts reproduce.
Third, Second, the macro trends that make the
super-critical regime plausible come from analyst predictions and vendor
surveys, which are gray literature: Gartner predicts, as reported by Reuters,
that a third of enterprise software applications will include agentic AI by
2028 and that 15\% of day-to-day work decisions will then be made
autonomously, but also that more than 40\% of agentic AI projects will be
cancelled by the end of 2027~\cite{reuters2025agentic}; these forecasts mostly
concern sanctioned enterprise software, so they indicate the spread of agentic
capability rather than of shadow artifacts. Fourth, adoption is
moving fast and unevenly: in the 2025 Stack Overflow survey, AI agents were
``not yet mainstream'' among developers~\cite{stackoverflow2025survey}, while
one year later the survey reported clear leaders among coding
agents~\cite{stackoverflow2026survey}, and repository studies estimate that a
quarter of active GitHub projects already use them~\cite{robbes2026agentic}.

The argument therefore stands on its mechanisms rather than on its numbers:
agents lower the cost of creating tools, act without asking, and create
artifacts that act in turn. The first two mechanisms are documented; the
third is plausible but unmeasured; their combination is compounding by
construction. Whether a given organisation is above or below
the threshold is an empirical question, and answering it requires the
measurement capability that the rest of this paper builds.

\section{Threat Model and Requirements}\label{sec:threat}

\subsection{Setting}

We consider an organisation that manages a fleet of Windows laptops with a
mobile device management service, here Microsoft Intune, and that has adopted
an AI usage policy listing approved tools. Employees, and in particular
developers, have user accounts that may or may not hold administrator rights.
The organisation operates a SIEM that collects Windows events. Its question is
not whether an employee \emph{could} use an unapproved agent, but which agents
are actually present or in use, on which machine, for which user, and whether
the user was informed of the policy.

\subsection{Actors}

\begin{description}[leftmargin=1.2em,style=nextline]
  \item[A1 -- The well-intentioned employee.] The primary actor. They install
    or run an agent to be more productive, without asking IT Security and
    often without knowing that a policy exists. They do not try to hide the
    agent. This is the classic profile of shadow IT
    (\cref{sec:shadow}), and the one that matters most by volume.
  \item[A2 -- The builder.] A developer or analyst who writes their own agent:
    a script that calls a model API, possibly through an aggregator or a
    company-unknown relay, with tools that run commands. They do not hide
    either, but their agent has no product name, no installer and no
    signature.
  \item[A3 -- The agent itself.] An agent acts with the privileges of its user
    and can, by itself, install packages, start MCP servers, write scripts,
    schedule tasks or configure other tools. It creates shadow IT of its own
    (\cref{sec:acceleration}).
  \item[A4 -- The evader.] A user who deliberately hides an agent, for
    example by renaming binaries, running it in a virtual machine or a
    container, or routing it through an obscure domain. A4 is not the design
    target: an endpoint detector cannot win against a determined local
    administrator. \Cref{sec:evaluation} nevertheless measures how simple
    evasions affect detection, and \cref{sec:discussion} discusses the
    complementary controls.
\end{description}

\subsection{Risks addressed}

The risks of unapproved agents are those of shadow IT, amplified by
autonomy: company data sent to providers that have no contract with the
organisation; code and commands executed by a model on a machine that holds
credentials and source code; supply-chain exposure through MCP servers and
agent extensions~\cite{hou2025mcplandscape,beurerkellner2025toolpoisoning};
and the loss of accountability when no one can say which agent did what. The
regulatory dimension adds to these: an organisation subject to the GDPR must
know which processors receive personal data, and the EU AI Act asks deployers
to support the AI literacy of their staff~\cite{eu2024aiact,eu2026aiomnibus}.

\subsection{Goal and non-goals}

The goal is \emph{visibility with informed acknowledgement}: to know, per machine and
per user, which agents are installed, used recently or active, with evidence
an analyst can trust; and to make sure that each user who runs one has been
shown the policy. The non-goals are as important. The detector does not block
agents, because blocking without visibility only pushes usage to less visible
channels, and because blocking decisions belong to existing controls such as
application control. It does not inspect content, because the question is
which agent is used, not what it is asked. And it does not claim to see what
happens outside the endpoint: agents that run entirely in the cloud are the
job of network and cloud controls.

\subsection{Requirements}

\begin{description}[leftmargin=1.2em,style=nextline]
  \item[R1 -- Coverage beyond products.] Detect known tools and product-less
    agents (A1 and A2), including agents reached through aggregators, relays
    and local runtimes.
  \item[R2 -- Precision.] Do not flag ordinary developer activity. Every false
    alert erodes the trust of users and analysts, and with a notice that the
    user must acknowledge, a false positive is also a usability defect.
  \item[R3 -- Privacy.] Collect no content and no secret; minimise what is
    collected to what identifies a tool.
  \item[R4 -- Low cost.] Run on every laptop every hour without noticeable
    impact.
  \item[R5 -- Explainability.] Attach evidence to every finding.
  \item[R6 -- Deployability.] Deploy and update with the MDM the
    organisation already uses, including on locked-down machines.
  \item[R7 -- Graceful degradation.] Report partial results and coverage gaps
    instead of failing, and never interfere with the user's work beyond the
    notice.
\end{description}

\subsection{Out of scope}

Agents that run entirely in a provider's cloud and never touch the endpoint;
agents inside virtual machines, containers or WSL, whose processes and sockets
are invisible from the host (we measure one such case); non-Windows endpoints;
unmanaged personal devices; and deliberate evasion by a local administrator.

\section{Why Behavioral Detection Falls Short}\label{sec:behavioral}

Behavioural detection is the obvious answer to agents that have no signature.
Endpoint detection and response (EDR) products have long caught unknown malware
by what it does rather than by what it is: a document editor that starts a
shell, a process that injects code into another. An agent has an equally
distinctive behaviour, or so it seems: it keeps a connection to a remote
service and, between two messages, runs commands. This section argues that
behaviour alone cannot separate agents from ordinary software on a developer's
workstation, explains why, and derives the design principle that \AD{}
follows. \Cref{sec:eval-rq6} measures the claim.

\subsection{What behavioural detection means here}

We draw the boundary operationally. A feature belongs to \emph{identity},
written $\iota$, when it is a string that names a piece of software or a
service: an image name or path that names a product, a publisher or signer,
a package or module name, a command-line argument that names a package, a
host name, an address range, the value of a configuration variable. Every
other observable belongs to \emph{behaviour}, written $\beta$: the shape of
the process tree, the kind of process (shell, interpreter), file and registry
activity, and the number, duration, timing and volume of network
connections. We treat the names of generic system components such as
\code{cmd.exe} or \code{python.exe} as behavioural, because they say what a
process does, not which product it is. A detector is \emph{behavioural} when
its decision depends only on $\beta$. Ignoring identity is what makes
behavioural detection attractive: it promises to catch agents that no catalog
lists.

The canonical structural rule for agents, which we evaluate, is: \emph{an
interpreter or runtime that holds an outbound connection on port 443 and has
a shell among its descendants}. A broader variant drops the interpreter condition.

\subsection{Seven reasons it falls short}

\paragraph{1. Agents make the same gestures as developers.}
An agent edits files, runs \code{git}, installs packages with \code{npm} or
\code{pip}, runs tests and opens HTTPS connections. So does a developer, and
so do the tools a developer uses every minute. \code{npm install} downloads
packages over TLS and runs their install scripts through \code{cmd.exe}; test
runners and build systems start shells; an administrator's PowerShell script
calls the Microsoft Graph API and runs commands. On a developer workstation,
``an interpreter with a TLS connection that runs a shell'' is not a rare
event; it is the background.

\paragraph{2. The host processes are legitimate.}
Commercial agents run inside processes that are themselves legitimate and
signed: the extension host of VS Code (a Node.js process), \code{python.exe},
\code{node.exe}, a packaged desktop application. A signature identifies the
publisher of the host, not what the host is doing. Flagging \code{node.exe}
because it runs shells flags the IDE.

\paragraph{3. The traffic does not say what it is.}
Model APIs use TLS like everything else, often behind content-delivery
networks that also serve unrelated sites. Aggregators and relays put another
name in front of the provider, and Encrypted Client Hello hides the server
name~\cite{rescorla2026ech}. Traffic analysis can recover a great deal from
packet sizes and timing~\cite{weiss2024prompt,mcdonald2025whisperleak,zhang2025agentprint,pouryousef2026llmtraffic},
but it needs a capture point on the network path, trained models, and it
answers with a probability, not with a process and a user.

\paragraph{4. Agents can be tiny and short-lived.}
A working agent fits in a few dozen lines of Python and can run for a few seconds.
There is no installer, no service and no long-lived window. A periodic
observer misses it unless the agent happens to run at observation time;
continuous telemetry catches it, but must then decide, in real time and
without context, what a short burst of HTTPS and shell activity means.

\paragraph{5. Local models are silent on the network.}
An agent that uses a local runtime such as Ollama talks to a loopback port.
Every rule that keys on outbound TLS misses it entirely.

\paragraph{6. Agents are benign by nature.}
Behavioural engines are trained and tuned on malicious behaviour. An agent's
actions are, by construction, the actions its user asked for, executed with
the user's rights. The relevant question is not whether the behaviour is
malicious but whether the tool is \emph{authorised}, which is a property of
identity and policy, not of behaviour.

\paragraph{7. False positives are expensive.}
For a security analyst, a false positive costs a few minutes. When the
response is a notice that a user must acknowledge, a false positive interrupts
a person who did nothing wrong, and a detector that interrupts every build
loses its users' trust within a day. Precision is therefore a first-class
requirement (R2 in \cref{sec:threat}).

\subsection{The behavioural-twin argument}

The first three reasons can be stated precisely. Describe the observable
footprint of a running program $p$ as a pair
\[
  x(p) = \big(\beta(p),\ \iota(p)\big),
\]
where $\beta(p)$ collects the behavioural features (process tree, command
patterns, number and timing of connections) and $\iota(p)$ collects the
identities (software identity and the identity of the remote services). Call
two programs $a$ and $b$ \emph{behavioural twins} when $\beta(a)=\beta(b)$ and
$\iota(a)\neq\iota(b)$.

\begin{quote}
  \emph{For any detector $f$ that depends only on $\beta$, and any pair of
  behavioural twins $(a,b)$, $f(a)=f(b)$. If $a$ is an agent and $b$ is not,
  $f$ either misses $a$ or falsely flags $b$.}
\end{quote}

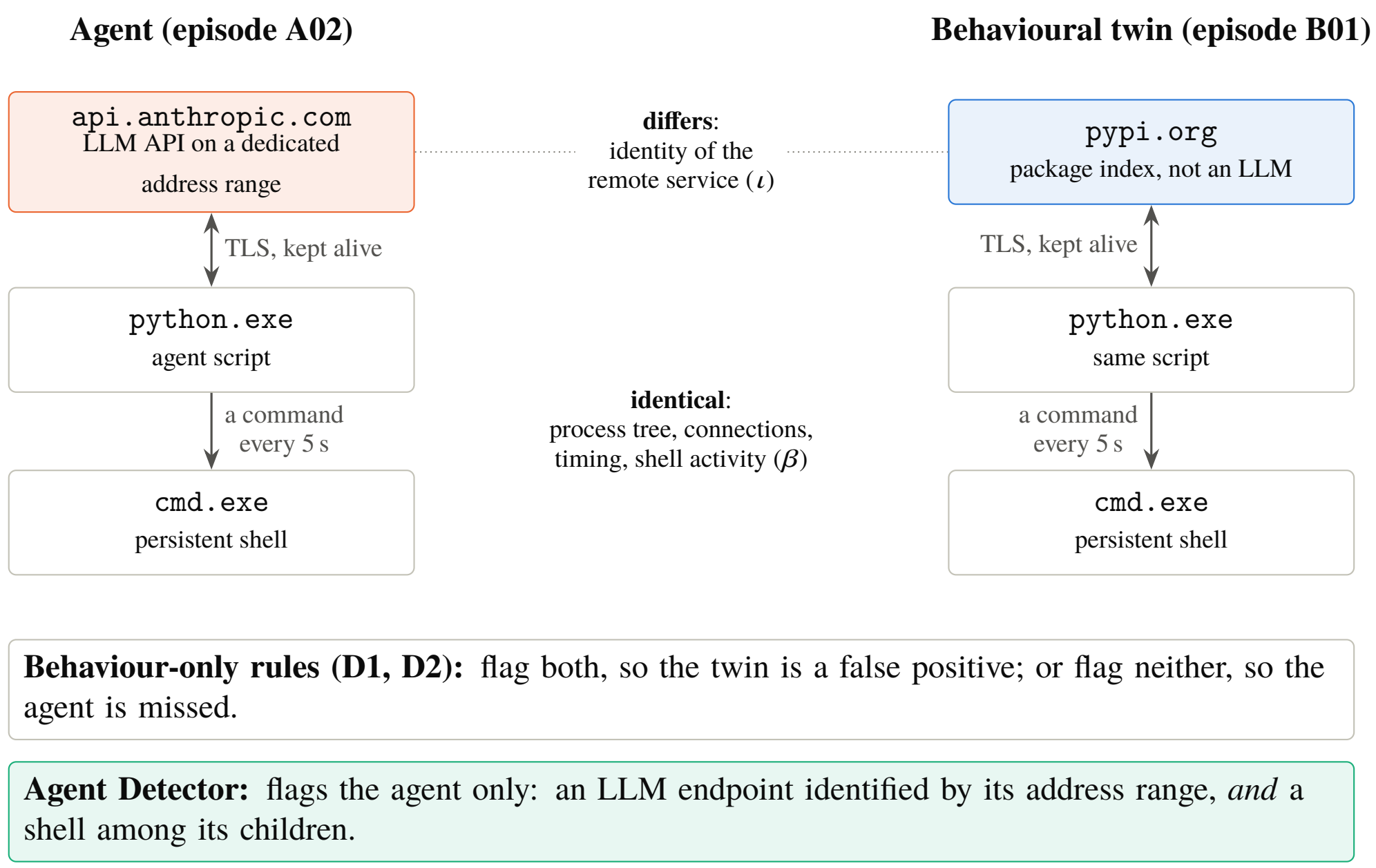
\begin{figure}[t]
  \centering
\begin{tikzpicture}[
    proc/.style={viznode, text width=3.6cm, minimum height=10mm},
    llm/.style={proc, fill=vizOrange!12, draw=vizOrange},
    other/.style={proc, fill=vizBlue!10, draw=vizBlue},
    tls/.style={{Stealth[length=2mm]}-{Stealth[length=2mm]}, semithick, draw=vizInkTwo},
    verdict/.style={viznode, text width=12.6cm, align=left, minimum height=9mm}]
  \node[font=\small\bfseries, text=vizInk] at (0, 2.95) {Agent (episode A02)};
  \node[llm] (ra) at (0, 1.8) {\texttt{api.anthropic.com}\\[-1pt]{\scriptsize LLM API on a dedicated\\address range}};
  \node[proc] (pa) at (0, 0) {\texttt{python.exe}\\[-1pt]{\scriptsize agent script}};
  \node[proc] (sa) at (0, -1.75) {\texttt{cmd.exe}\\[-1pt]{\scriptsize persistent shell}};
  \draw[tls] (pa) -- node[right, vizlabel] {TLS, kept alive} (ra);
  \draw[vizarrow] (pa) -- node[right, vizlabel] {a command\\every 5\,s} (sa);
  \node[font=\small\bfseries, text=vizInk] at (9.0, 2.95) {Behavioural twin (episode B01)};
  \node[other] (rb) at (9.0, 1.8) {\texttt{pypi.org}\\[-1pt]{\scriptsize package index, not an LLM}};
  \node[proc] (pb) at (9.0, 0) {\texttt{python.exe}\\[-1pt]{\scriptsize same script}};
  \node[proc] (sb) at (9.0, -1.75) {\texttt{cmd.exe}\\[-1pt]{\scriptsize persistent shell}};
  \draw[tls] (pb) -- node[left, vizlabel] {TLS, kept alive} (rb);
  \draw[vizarrow] (pb) -- node[left, vizlabel] {a command\\every 5\,s} (sb);
  \node[vizlabel, text=vizInk] (diff) at (4.5, 1.8) {\textbf{differs}:\\identity of the\\remote service ($\iota$)};
  \node[vizlabel, text=vizInk] at (4.5, -0.875) {\textbf{identical}:\\process tree, connections,\\timing, shell activity ($\beta$)};
  \draw[vizMuted, densely dotted] (ra.east) -- (diff.west);
  \draw[vizMuted, densely dotted] (diff.east) -- (rb.west);
  \node[verdict, fill=white] (v1) at (4.5, -3.35) {\textbf{Behaviour-only rules (D1, D2):} flag both, so the twin is a false positive; or flag neither, so the agent is missed.};
  \node[verdict, fill=vizAqua!10, draw=vizAqua, below=2mm of v1] (v2) {\textbf{\AD{}:} flags the agent only: an LLM endpoint identified by its address range, \emph{and} a shell among its children.};
\end{tikzpicture}
  \caption{A behavioural twin from the evaluation. The agent and its twin run
    the same script with the same shell and the same connection pattern; only
    the remote service differs. Any detector that ignores the identity of the
    remote service must treat them alike.}
  \label{fig:twin}
\end{figure}

The statement is trivial, and that is its point: twins are not a corner case
to be engineered around (\cref{fig:twin}). Take an agent that polls a model API over a
keep-alive connection and pipes commands into a persistent shell; replace the
model API with any other HTTPS API, such as a package index, and the result is
an ordinary automation script with exactly the same behaviour. Real benign
workloads are not exact twins, but they are close, and they are numerous: the
build tools of point~1 differ from an agent in timing and volume, not in
kind. A behavioural detector can only trade recall for precision along that
continuum.

The argument has a clear limit. Exact twins exist, but real agents and real
automation differ in features that a richer behavioural detector could use:
the token-streaming shape of LLM responses, which traffic analysis exploits
on the wire, or the coupling between the end of a model response and the
next command an agent runs. Such features are behavioural under our
definition, and nothing in the argument rules them out. What we test is
narrower: the structural rules D1 and D2, which use only the process tree and
the presence of connections (\cref{sec:eval-rq6}). A stronger behavioural
baseline, built on timing and volume and evaluated on real agents against
real automation, is future work. On the endpoint, however, the identity of
the remote service is usually available directly, and reading it is cheaper
and more explainable than inferring it.

\subsection{Design principle: anchor behaviour to identity}

The argument does not make behaviour useless. It makes it insufficient
\emph{alone}. Identity alone has the symmetric weakness: a connection to a
model API says that a process talks to a model, not that it acts; a chat
client and an agent look the same. The combination is what discriminates:

\begin{itemize}
  \item \emph{talks to a model service} (identity of the remote, from the DNS
    cache or a dedicated address range) \emph{and}
  \item \emph{acts on the machine} (behaviour: shells and interpreters among
    its descendants, MCP servers, automated browsers),
\end{itemize}

completed by the evidence that does not depend on the process being alive:
configuration (base URLs, MCP definitions), installation and history. This is
the \code{probable-agent} rule of \cref{sec:correlation} and, more generally,
the reason why \AD{} correlates layers instead of scoring behaviour.
\Cref{sec:eval-rq6} tests the principle on behavioural twins and on real
developer workloads, against three baselines: a signature detector and the two
behavioural rules above.

\section{A Taxonomy of Agent Footprints}\label{sec:taxonomy}

To detect agents without depending on any single product, we first
characterise what agents leave behind on a Windows endpoint. This section
defines nine classes of agents and six layers of observable footprint, and
maps one onto the other. The taxonomy is the result of building and
maintaining the catalog of \AD{}: every class is represented in it, and every
layer is read by at least one collector.

\subsection{Classes of agents}

\Cref{tab:classes} lists the classes, with examples taken from the catalog.
They differ in how the agent is packaged, which process hosts it, how it
reaches a model, and what state it keeps on disk.

\begin{table}[p]
  \centering
  \caption{Classes of AI agents found on enterprise endpoints, with the
    catalog examples, the process that hosts them and their usual route to a
    model.}
  \label{tab:classes}
  \small
  \begin{tabularx}{\linewidth}{@{}>{\raggedright\arraybackslash}p{2.7cm} >{\raggedright\arraybackslash}X >{\raggedright\arraybackslash}p{3.2cm} >{\raggedright\arraybackslash}p{3.3cm}@{}}
    \toprule
    Class & Examples & Host process & Route to a model \\
    \midrule
    C1 Command-line coding agents & Claude Code, OpenAI Codex CLI, Gemini CLI, GitHub Copilot CLI, OpenCode, Aider, Goose, Open Interpreter
      & Native executable, or Node.js / Python runtime & Vendor API, aggregator, or a base URL set by the user \\
    C2 Desktop agents & Claude Desktop (including its Cowork mode), the ChatGPT and Codex desktop applications
      & Packaged (MSIX) application, sometimes with a SYSTEM service & Vendor API \\
    C3 AI-native IDEs & Cursor, Windsurf, Kiro, Trae & Electron application & Vendor back end \\
    C4 IDE agents and assistants & GitHub Copilot, Cline, Roo Code, Kilo Code, Continue, Amazon Q, Gemini Code Assist, Augment, Tabnine, Qodo, JetBrains AI and Junie
      & IDE extension host (VS Code) or JVM (JetBrains) & Vendor back end or any API configured by the user \\
    C5 Local LLM runtimes & Ollama, LM Studio, Jan, GPT4All, AnythingLLM, llama.cpp and llamafile, Hugging Face models
      & Own server process; in-process inference libraries & Local port (loopback) \\
    C6 Agentic browsers and extensions & Perplexity Comet; AI browser extensions & Browser & Vendor back end \\
    C7 Frameworks and SDKs & LangChain-style frameworks, CrewAI, AutoGen, OpenAI Agents SDK, Claude Agent SDK, Browser Use, LiteLLM proxy
      & Python or Node.js interpreter & Any API \\
    C8 Home-made agents & A script that calls a model API in a loop and runs commands & Interpreter & Any API, often through an aggregator or relay \\
    C9 Agent infrastructure & MCP servers, self-hosted gateways and relays, automated browsers & Interpreter, browser & Serves an agent \\
    \bottomrule
  \end{tabularx}
\end{table}

Two properties of this list shape the detection problem. First, the classes
are not disjoint in practice: a C1 agent such as Claude Code may run inside a
C4 host as an IDE extension, start C9 MCP servers, and reach its model through
an aggregator. Second, the classes span a gradient of \emph{product-ness},
from signed, packaged desktop applications (C2) to agents that exist only as a
text file run by a general-purpose interpreter (C8). The more an agent looks
like C8, the fewer stable identifiers it offers.

\subsection{Layers of footprint}

We distinguish six layers of evidence that an agent can leave on an endpoint.

\begin{description}[leftmargin=1.2em,style=nextline]
  \item[L1 -- Software identity.] Image name, install path, version resource
    (product and company), Authenticode signer, MSIX package family.
  \item[L2 -- Runtime structure.] Command line, parent and child processes,
    loaded modules, environment variables of the running process.
  \item[L3 -- Network.] Established connections and the identity of the remote
    service, listening ports, connections to local LLM ports.
  \item[L4 -- Installation.] Installed-program records, packaged-application
    registrations, services, IDE and browser extensions, global npm and pip
    packages.
  \item[L5 -- Configuration and state.] Configuration folders, MCP server
    definitions~\cite{anthropic2026claudecodesettings,openai2026codexmcp,microsoft2026vscodemcp,cursor2026mcp,mcp2026localservers},
    base-URL variables, the presence of API-key variables, downloaded models.
  \item[L6 -- History.] Execution traces kept by Windows (BAM and
    Prefetch~\cite{velociraptor2026bam,neyaz2022prefetch}) and the
    modification time of an agent's history folders.
\end{description}

Layers L1, L4, L5 and L6 describe \emph{what is installed or was used};
layers L2 and L3 describe \emph{what is happening now}. Signature-based
inventories, such as the software inventory of an EDR
platform~\cite{microsoft2026mdvminventory}, live in L1 and L4. Behavioural
detection lives in L2 and, from the network vantage point, in L3.

\subsection{Mapping classes to layers}

\Cref{fig:matrix} maps each class to the layers in which it is usually
visible. The pattern that matters is the last rows: home-made agents (C8)
and most agent infrastructure (C9) have no software identity, no installation
record and no history of their own. A script run by \code{python.exe} leaves,
in L1, L4 and L6, only the trace of \code{python.exe}. For these classes, the
only evidence is what the process does now (L2), what it talks to (L3) and,
sometimes, how it was configured (L5). This is why product-less agents defeat
inventories, and why, as \cref{sec:behavioral} argues, behaviour alone cannot
replace the missing identity: it must be anchored to the identity of the
remote service.

\begin{figure}[t]
  \centering
\newcommand{\vizsymP}{\ding{108}}
\newcommand{\vizsymS}{\ding{109}}
\newcommand{\vizsymN}{--}
\begin{tikzpicture}[x=1.62cm, y=0.66cm,
    cellP/.style={fill=vizBlue, text=white},
    cellS/.style={fill=vizBlueLight, text=vizBlueDark},
    cellN/.style={fill=white, draw=vizGrid, text=vizMuted}]
  \foreach \lab [count=\j] in {{L1\\Identity}, {L2\\Runtime}, {L3\\Network}, {L4\\Install}, {L5\\Config}, {L6\\History}}
    \node[font=\scriptsize, align=center, text=vizInk] at (\j, 0.95) {\lab};
  \foreach \name/\cells [count=\i] in {
      {C1 Command-line agents}/{P,P,P,S,P,P},
      {C2 Desktop agents}/{P,S,P,P,P,P},
      {C3 AI-native IDEs}/{P,S,P,P,P,P},
      {C4 IDE agents}/{S,S,P,P,P,S},
      {C5 Local LLM runtimes}/{P,P,P,P,P,P},
      {C6 Agentic browsers}/{P,S,N,P,S,P},
      {C7 Frameworks and SDKs}/{N,S,P,P,S,N},
      {C8 Home-made agents}/{N,P,P,N,S,N},
      {C9 Agent infrastructure}/{S,P,S,S,P,N}} {
    \node[anchor=east, font=\small, text=vizInk] at (0.45, -\i) {\name};
    \foreach \c [count=\j] in \cells
      \node[cell\c, rounded corners=1.5pt, minimum width=1.52cm, minimum height=0.56cm, font=\small]
        at (\j, -\i) {\csname vizsym\c\endcsname};
  }
\end{tikzpicture}
  \caption{Where each class of agent is visible. Dark cells (\ding{108}):
    primary signal; light cells (\ding{109}): secondary or occasional signal;
    blank cells (--): no signal of its own. The classes are ordered from
    packaged products (top) to product-less agents and infrastructure
    (bottom). Browsers are deliberately excluded from network attribution
    (L3) in \AD{}.}
  \label{fig:matrix}
\end{figure}

\subsection{Anatomy of two footprints}

\Cref{tab:anatomy} makes the gradient concrete with two agents observed during
the evaluation (\cref{sec:evaluation}). Claude Code, a commercial coding agent
installed on the evaluation workstation, left evidence in all six layers: its process, identified by its install path; a
connection to the address range dedicated to its provider's API; the shells it
ran; a VS Code extension in two versions, a global npm package and a launcher
script; its configuration folders; and the modification time of its project
history. The home-made agent of the bench, a short Python script that reproduces the
same behaviour, left two: a connection and a shell. Remove either,
and nothing remains.

\begin{table}[t]
  \centering
  \caption{Evidence collected for a commercial agent and for a home-made agent
    on the same workstation (standard-user scans; paths anonymised).}
  \label{tab:anatomy}
  \small
  \begin{tabularx}{\linewidth}{@{}l >{\raggedright\arraybackslash}X >{\raggedright\arraybackslash}X@{}}
    \toprule
    Layer & Claude Code (in-situ) & Home-made Python agent (episode A02) \\
    \midrule
    L1 Identity & \code{claude.exe}, matched by its install path & -- (\code{python.exe}) \\
    L2 Runtime & runs \code{cmd.exe} and \code{powershell.exe} & runs \code{cmd.exe} \\
    L3 Network & connected to the Anthropic API range & connected to the Anthropic API range \\
    L4 Installation & VS Code extension (two versions), global npm package, launcher script & -- \\
    L5 Configuration & configuration and data folders under the profile & -- \\
    L6 History & last activity in its project-history folder & -- \\
    \midrule
    Finding & Claude Code: active, high confidence & LLM API used by \code{python.exe}, and probable custom agent: active, medium confidence \\
    \bottomrule
  \end{tabularx}
\end{table}

\subsection{The endpoint catalog}

Network attribution needs a list of the services that agents talk to. The
catalog of the evaluated version lists 184 endpoint entries (exact hosts, host
suffixes and address ranges) in nine categories (\cref{tab:endpoints}). The
long tail matters: beyond the well-known model providers, agents reach models
through aggregators that resell many models behind a single API, inference
platforms that host open-weight models, hosted gateways, and relays that
re-expose a provider's API under another domain.

\begin{table}[t]
  \centering
  \caption{Endpoint categories of the catalog (version 2026.11.09), with
    representative providers.}
  \label{tab:endpoints}
  \small
  \begin{tabularx}{\linewidth}{@{}l r X@{}}
    \toprule
    Category & Entries & Representative providers \\
    \midrule
    Model provider     & 50 & OpenAI, Anthropic, Google Gemini and Vertex AI, Azure OpenAI and Azure AI Foundry, Mistral, xAI, DeepSeek, Alibaba DashScope, Moonshot \\
    Inference platform & 29 & Together AI, Fireworks AI, Groq, Cerebras, SambaNova, DeepInfra, Replicate, Hugging Face inference, NVIDIA NIM \\
    Web assistant      & 28 & ChatGPT, Claude.ai, Gemini, Microsoft Copilot, Perplexity, Le Chat, Grok, Meta AI \\
    Aggregator         & 25 & OpenRouter, Requesty, Poe API, OpenCode Zen, AI/ML API, Eden AI, Unify, Glama, Martian \\
    Developer tool     & 24 & GitHub Copilot (personal, Business, Enterprise), Cursor, Kiro and Amazon Q, Windsurf, Augment, Amp, Zed, Trae, Tabnine \\
    Agent infrastructure & 11 & Browserbase, Browser Use Cloud, Steel, Hyperbrowser, E2B sandboxes, Composio, Smithery, hosted MCP servers \\
    Gateway            & 8  & Portkey, Cloudflare AI Gateway, Vercel AI Gateway, Helicone \\
    Relay              & 7  & Publicly known resellers of OpenAI-compatible APIs \\
    Model download     & 2  & Ollama model registry, Hugging Face (shared with non-AI use) \\
    \midrule
    Total              & 184 & \\
    \bottomrule
  \end{tabularx}
\end{table}

Two attributes of an entry drive the confidence of a network finding. An
entry is \emph{shared} when the host also serves non-AI traffic (eight
entries, such as \code{huggingface.co}); a connection to it alone never makes
a finding of medium confidence. An entry can also be an \emph{address range}
dedicated to a service: the evaluated catalog contains the IPv4 and IPv6
ranges that Anthropic documents for its API~\cite{anthropic2026ipaddresses}.
In the evaluation, the IPv4 range gives every agent that uses this API a
finding of medium confidence on its own, whether or not the provider's name
is still in the DNS cache (\cref{sec:evaluation}). An earlier version of the catalog lacked
the IPv6 range and still flagged the IPv4 range as unverified after it had
been checked; we report this as it is, because it illustrates the main
maintenance risk of any catalog-driven detector: drift between the catalog
and a fast-moving ecosystem (\cref{sec:discussion}).

Finally, the catalog holds 39 tool signatures. By category, they cover
command-line agents (8), IDE agents (8), local LLM runtimes (7), AI-native IDEs
(4), IDE assistants (4), agent frameworks and SDKs (3), and one entry each for
desktop agents, desktop assistants, agentic browsers, AI browser extensions
and LLM gateways. \Cref{app:catalog} lists them.

\section{Landscape of Detection Approaches}\label{sec:landscape}

\begin{figure}[t]
  \centering
\begin{tikzpicture}[x=0.85cm,
    zone/.style={draw=vizAxis, rounded corners=4pt, fill=vizSurface, inner sep=6pt},
    itembox/.style={viznode, minimum width=3.3cm, minimum height=7mm},
    faded/.style={viznode, minimum width=3.3cm, minimum height=7mm, dashed, text=vizInkTwo},
    zonetitle/.style={font=\small\bfseries, text=vizInk},
    who/.style={vizlabel, text=vizInk, text width=3.6cm, align=flush center}]
  \node[itembox] (ide) at (0, 1.2) {IDE with a coding agent};
  \node[itembox] (cli) at (0, 0.3) {Command-line agent};
  \node[itembox] (scr) at (0, -0.6) {Home-made script};
  \node[itembox] (loc) at (0, -1.5) {Local LLM (loopback)};
  \node[faded] (vm) at (0, -2.4) {Agent in WSL, VM, container};
  \begin{scope}[on background layer]
    \node[zone, fit=(ide)(vm), label={[zonetitle]above:Laptop (endpoint)}] (ep) {};
  \end{scope}
  \node[itembox, minimum width=3.0cm] (px) at (5.6, 0.3) {Proxy, CASB,\\TLS inspection};
  \begin{scope}[on background layer]
    \node[zone, fit=(px), label={[zonetitle]above:Corporate network}] (net) {};
  \end{scope}
  \node[itembox] (api) at (11.0, 1.0) {LLM APIs};
  \node[itembox] (agg) at (11.0, 0.1) {Aggregators, gateways};
  \node[itembox] (rel) at (11.0, -0.8) {Relays on any domain};
  \begin{scope}[on background layer]
    \node[zone, fit=(api)(rel), label={[zonetitle]above:Internet services}] (inet) {};
  \end{scope}
  \node[faded] (cloud) at (11.0, -2.4) {Cloud agents};
  \draw[vizarrow] (ep.east |- px) -- node[above, vizlabel] {on site} (net.west);
  \draw[vizarrow] (net.east) -- (inet.west |- px);
  \draw[vizarrow] (ep.east |- 0,-1.2) -- node[below, vizlabel] {off site: home, travel} (inet.west |- 0,-1.2);
  \node[who, anchor=north] at (0, -3.25) {\textbf{Endpoint:} \AD{}, EDR inventory, vendor local-agent discovery. Sees processes and users; not inside VMs.};
  \node[who, anchor=north] at (5.6, -3.25) {\textbf{Network:} proxy and CASB discovery, traffic analysis. Sees hosts on site; not processes, not local models.};
  \node[who, anchor=north] at (11.0, -3.25) {\textbf{Cloud:} app catalogs, data posture. Sees services and data; not the endpoint.};
\end{tikzpicture}
  \caption{Vantage points. Each approach sees part of the path from an agent
    to its model. Dashed boxes are out of reach of the endpoint: agents inside
    virtual machines, containers or WSL, and agents that run entirely in the
    cloud.}
  \label{fig:vantage}
\end{figure}
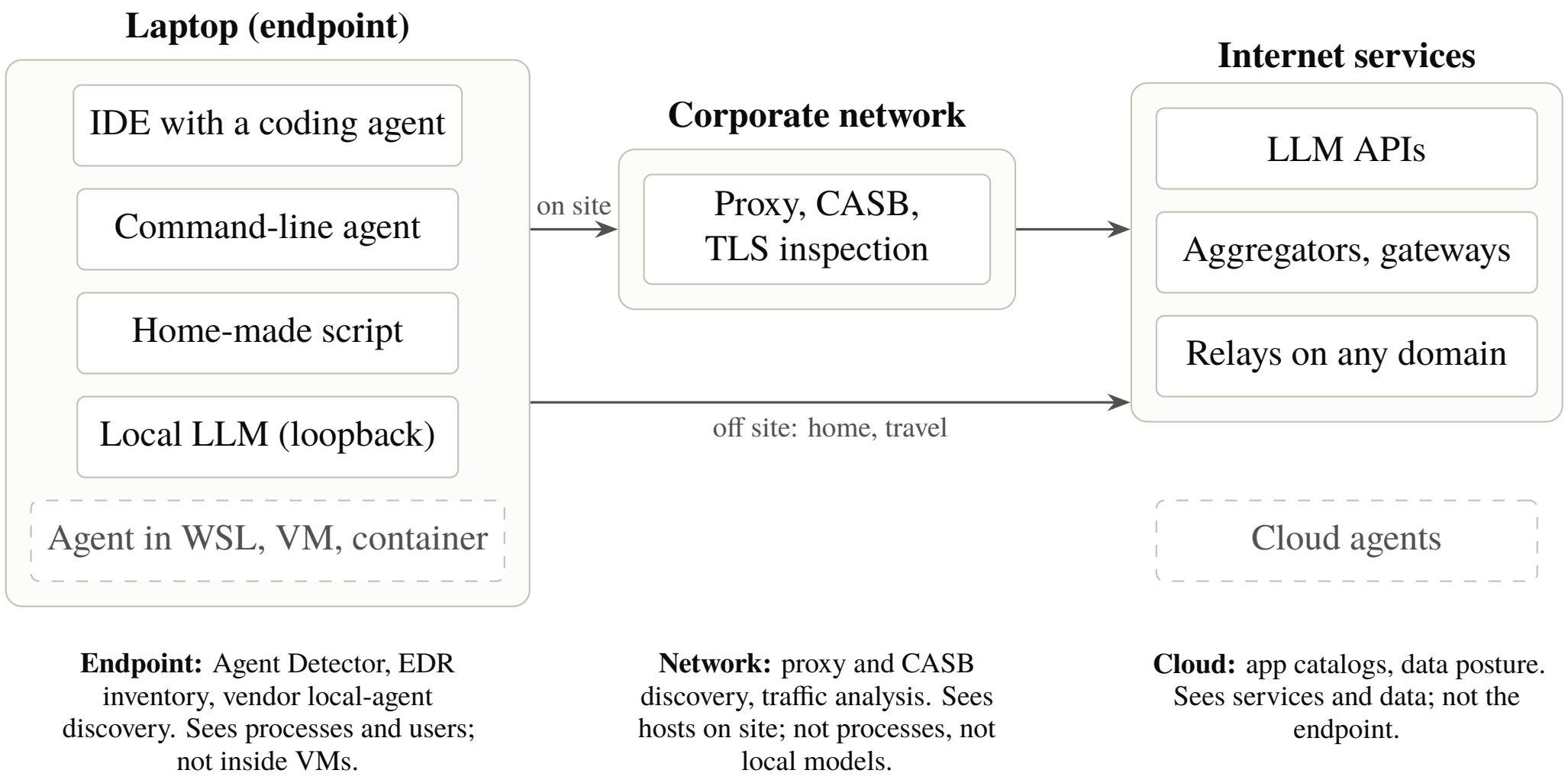

Organisations already have several instruments that could, in principle, see
AI agents. This section reviews them by vantage point (network, traffic
analysis, data, endpoint inventory, vendor agent discovery) and asks of each
the questions of \cref{sec:threat}: does it see product-less agents, local
models, laptops outside the corporate network, and the user behind the agent;
and does it inform that user? \Cref{fig:vantage} shows where each approach looks,
and \cref{tab:landscape} summarises the answers.

\subsection{Network discovery: proxies and CASB}

Secure web gateways and cloud access security brokers discover cloud
applications from traffic logs and match them against a catalog. Microsoft
Defender for Cloud Apps, for example, lists more than 31{,}000 discoverable
applications, with categories for generative AI, AI model providers and MCP
servers; by design, it ``can't discover apps that aren't in the
catalog''~\cite{microsoft2026dcacatalog}. Network discovery is the natural
place to govern \emph{web} use of AI services, and \AD{} deliberately leaves
browser traffic to it.

For agents, the network vantage point has four structural limits. It sees a
connection to an API, not the process behind it, so an unapproved script and
an approved application that call the same provider are indistinguishable.
Laptops spend much of their time outside the corporate network, where only an
endpoint-based proxy client sees their traffic. Local runtimes generate no
external traffic. And encryption keeps tightening: with TLS Encrypted Client
Hello, now a published standard, the server name no longer travels in clear,
even though the RFC notes that ECH alone ``is not in itself sufficient to
protect the identity of the server'' when DNS queries or server addresses
remain visible~\cite{rescorla2026ech}.

\subsection{Encrypted traffic analysis}

A line of research shows that LLM traffic remains highly informative even when
encrypted. The token-length side channel of streaming assistants allowed
Weiss et~al.\ to reconstruct 27\% of responses and infer the topic of 53\% of
them~\cite{weiss2024prompt}; Whisper Leak infers prompt topics from packet
sizes and timing across 28 models~\cite{mcdonald2025whisperleak}; AgentPrint
identifies which \emph{agent} produced a traffic trace with an F1-score of
0.866~\cite{zhang2025agentprint}; and a large-scale benchmark reports model
fingerprinting at 97.7\% balanced accuracy, with task fingerprints detectable
``even from a single agent's traffic''~\cite{pouryousef2026llmtraffic}. At the
transport layer, TLS client fingerprints such as JA3 and JA4 characterise the
client library that opened a connection~\cite{althouse2017ja3,althouse2023ja4}.

These results establish that agent behaviour is, in principle, learnable from
the network. They also show the cost of that route: a capture point on the
path, trained models per agent and per provider, a probabilistic answer, no
attribution to a process or a user, and an analysis that, by inferring topics,
comes uncomfortably close to the content that a governance tool should not
look at. This is the trade-off discussed in \cref{sec:behavioral}.

Continuous endpoint telemetry is a natural alternative that this paper does
not evaluate. Sysmon logs process creation (event 1), network connections
(event 3) and, when enabled, DNS queries with the querying process (event
22)~\cite{microsoft2026sysmonevents}; EDR platforms expose similar tables for
hunting queries~\cite{microsoft2026devicenetworkevents}. A rule such as
``a process resolved an LLM host and then started a shell'' implements the
principle of \cref{sec:behavioral} continuously: it needs no access to the DNS
cache and would catch short-lived agents that a periodic scan misses. It
requires the telemetry pipeline and its volume, and it sees neither
configuration nor installation; the two approaches are complementary.
Scanners of MCP configurations look at a different layer: they read the MCP
servers that clients declare and analyse their tool descriptions for
poisoning~\cite{invariant2025mcpscan}.

\subsection{Data-centric controls}

Data security posture management takes the opposite view: rather than
``focusing on infrastructure or endpoints'', it ``centers on the data
itself''~\cite{microsoft2026dspm}. Such controls answer which sensitive data
reaches which AI service, and complement agent discovery; they do not answer
which agents run on a laptop.

\subsection{Endpoint software inventory}

EDR and vulnerability-management platforms maintain a software inventory of
every device, built from registry and disk evidence and refreshed every few
hours~\cite{microsoft2026mdvminventory}. An inventory sees installed products
(layers L1 and L4 of \cref{sec:taxonomy}), including AI tools, but it sees
neither the agent that has no installer nor whether an installed tool is used.
Forensic work on the ChatGPT Windows application shows how much an AI client
leaves behind, including traces recoverable after
deletion~\cite{kankanamge2025chatgpt}; this supports using history and residue
as evidence, but on its own it describes the past, not current use.

\subsection{Vendor discovery of local agents}

The closest commercial analogue to this work appeared while it was under
way. Microsoft Defender for Endpoint now discovers local AI agents: ``for
process-based discovery, Defender identifies an agent when it observes the
agent's process running'', and it defines an agent as ``a combination of a
user, a device, and an agent type''~\cite{microsoft2026localagents}, which is
the same unit as our per-tool, per-user findings. The Microsoft~365 admin
center surfaces the result as \emph{shadow AI}, in public preview, for a list of
named agents that includes ChatGPT Desktop, Ollama, OpenCode and Claude
Desktop, with blocking available for one of them; it requires Defender for
Endpoint and Intune enrolment~\cite{microsoft2026shadowai}. Microsoft's
deployment guidance states that the scans ``run silently on managed devices;
end users are not notified or affected''~\cite{microsoft2026zta176}.

This confirms both the demand and the endpoint vantage point. It differs from
our approach in three ways that the rest of this paper examines: discovery is
tied to a list of products, so product-less agents fall outside it; it is
silent, so it produces visibility without the informed acknowledgement that
the governance goal of \cref{sec:threat} requires; and it depends on a
specific security suite.

\begin{table}[t]
  \centering
  \caption{Detection approaches against the requirements of
    \cref{sec:threat}. \cmark: yes; \ding{109}: partially or under
    conditions; \xmark: no.}
  \label{tab:landscape}
  \small
  \begin{tabularx}{\linewidth}{@{}>{\raggedright\arraybackslash}X c c c c c c@{}}
    \toprule
    Approach & \makecell{Known\\products} & \makecell{Product-less\\agents} & \makecell{Local\\models} & \makecell{Off\\network} & \makecell{Per\\user} & \makecell{Informs\\user} \\
    \midrule
    Proxy and CASB discovery~\cite{microsoft2026dcacatalog}      & \ding{109} & \ding{109} & \xmark     & \ding{109} & \ding{109} & \xmark \\
    Encrypted traffic analysis~\cite{zhang2025agentprint,pouryousef2026llmtraffic} & \ding{109} & \ding{109} & \xmark & \xmark & \xmark & \xmark \\
    Endpoint software inventory~\cite{microsoft2026mdvminventory} & \cmark     & \xmark     & \ding{109} & \cmark     & \ding{109} & \xmark \\
    Vendor local-agent discovery~\cite{microsoft2026localagents}  & \cmark     & \xmark     & \ding{109} & \cmark     & \cmark     & \xmark \\
    Continuous telemetry rules~\cite{microsoft2026sysmonevents}   & \ding{109} & \ding{109} & \ding{109} & \cmark     & \cmark     & \xmark \\
    \AD{} (this work)                                             & \cmark     & \ding{109} & \ding{109} & \cmark     & \cmark     & \cmark \\
    \bottomrule
  \end{tabularx}
\end{table}

The table is a summary of design properties, not of measured performance.
For \AD{}, the partial marks anticipate what \cref{sec:evaluation} measures:
product-less agents are caught only when their remote service can be
identified and while they run, and local models only on their default ports
or through known module names. Informing the user is a design choice that any
endpoint approach could adopt. The ``product-less'' column is the one this
paper is about; \cref{sec:behavioral} explained why the obvious way to fill
it, behavioural detection alone, is weaker than it looks, and the next
section presents our design.

\section{Design of Agent Detector}\label{sec:design}

This section presents \AD{}, the open-source endpoint detector we built to
test the claim of this paper: that shadow AI agents become observable when
several weak, individually ambiguous signals are correlated around the
identity of the software and the identity of the remote service. The source
code, the catalog and the evaluation bench are public~\cite{canale2026repo}.
\Cref{sec:implementation} describes the implementation and its deployment with
Microsoft Intune; this section focuses on the detection logic.

\subsection{Design goals}

The requirements of \cref{sec:threat} translate into six design goals.

\begin{description}[leftmargin=1.2em,style=nextline]
  \item[G1 -- Product-less coverage.] Detect agents that ship as no product at
    all, such as a short script that calls a model API, and not only
    the well-known tools.
  \item[G2 -- Silence on the network.] Never send a packet. Everything is
    read from local state that Windows already maintains: process table,
    TCP table, DNS client cache, registry and file system.
  \item[G3 -- Explainability.] Every finding carries the evidence that
    produced it, so that an analyst, and the user, can understand why a tool
    was reported.
  \item[G4 -- Privacy by design.] Never read prompts, conversations or file
    contents; never record or transmit secret values: record the presence of
    an API key, never its value, and only the host part of a base URL.
  \item[G5 -- Low cost.] One scan must take well under two seconds and a few
    tens of megabytes, so that it can run every hour on every laptop.
  \item[G6 -- Inform, do not block.] The output is a finding, an event and,
    when relevant, a notice that the user acknowledges. The detector never
    kills, quarantines or uninstalls anything.
\end{description}

\subsection{Architecture}

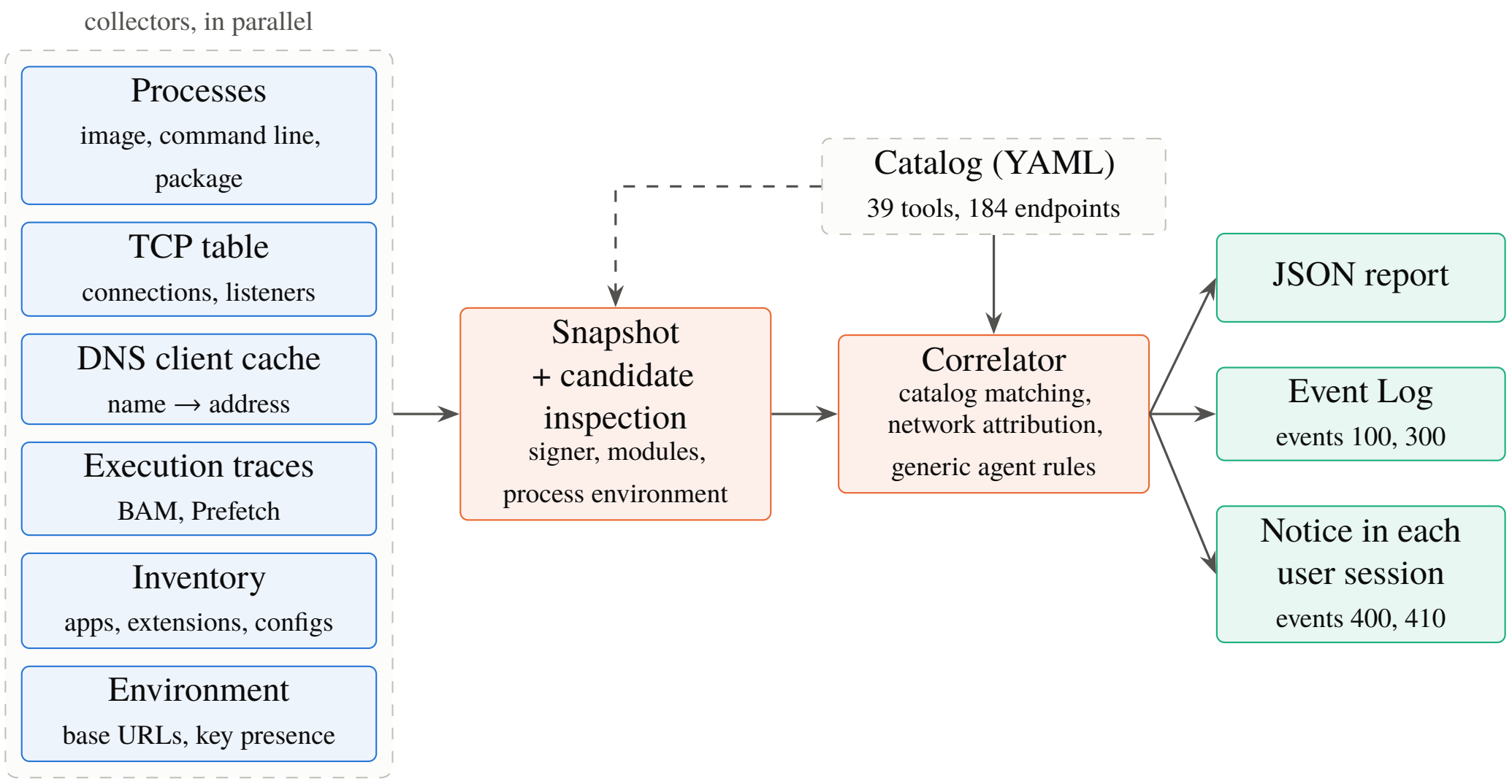
\begin{figure}[t]
  \centering
  \begin{tikzpicture}[
      font=\small,
      box/.style={draw=vizAxis, rounded corners=2pt, align=flush center, minimum height=8mm, inner sep=3pt, fill=white, text=vizInk},
      colbox/.style={box, text width=30mm, fill=vizBlue!8, draw=vizBlue},
      stagebox/.style={box, text width=26mm, fill=vizOrange!10, draw=vizOrange},
      outbox/.style={box, text width=24mm, fill=vizAqua!10, draw=vizAqua},
      arr/.style={-{Stealth[length=2mm]}, semithick, draw=vizInkTwo}]
    \node[colbox] (p) {Processes\\\scriptsize image, command line, package};
    \node[colbox, below=1.5mm of p] (n) {TCP table\\\scriptsize connections, listeners};
    \node[colbox, below=1.5mm of n] (d) {DNS client cache\\\scriptsize name $\to$ address};
    \node[colbox, below=1.5mm of d] (t) {Execution traces\\\scriptsize BAM, Prefetch};
    \node[colbox, below=1.5mm of t] (i) {Inventory\\\scriptsize apps, extensions, configs};
    \node[colbox, below=1.5mm of i] (e) {Environment\\\scriptsize base URLs, key presence};
    \begin{scope}[on background layer]
      \node[draw=vizAxis, dashed, rounded corners=3pt, fill=vizSurface, fit=(p)(e), inner sep=4pt,
        label={[font=\scriptsize, text=vizInkTwo]above:collectors, in parallel}] (coll) {};
    \end{scope}
    \node[stagebox, right=6mm of coll] (s) {Snapshot\\+ candidate\\inspection\\\scriptsize signer, modules,\\process environment};
    \node[stagebox, right=6mm of s] (c) {Correlator\\\scriptsize catalog matching,\\network attribution,\\generic agent rules};
    \node[outbox, right=6mm of c] (ev) {Event Log\\\scriptsize events 100, 300};
    \node[outbox, above=4mm of ev] (j) {JSON report};
    \node[outbox, below=4mm of ev] (dl) {Notice in each\\user session\\\scriptsize events 400, 410};
    \node[box, dashed, above=9mm of c, text width=29mm, fill=vizSurface] (cat) {Catalog (YAML)\\\scriptsize 39 tools, 184 endpoints};
    \draw[arr] (coll.east |- s) -- (s.west);
    \draw[arr] (s) -- (c);
    \draw[arr] (cat) -- (c);
    \draw[arr, dashed] (cat.west) -| (s.north);
    \draw[arr] (c.east) -- (j.west);
    \draw[arr] (c.east) -- (ev.west);
    \draw[arr] (c.east) -- (dl.west);
  \end{tikzpicture}
  \caption{Architecture of \AD{}. The collectors only read local state; the
    catalog drives both candidate inspection and correlation.}
  \label{fig:architecture}
\end{figure}

\Cref{fig:architecture} shows the pipeline. A scan is a one-shot process,
started by a scheduled task. Six collectors run in parallel and fill a
\emph{snapshot}; a candidate-inspection stage enriches the processes that
matter; a correlator turns the snapshot into \emph{findings}; the findings
feed three outputs: a JSON report, Windows Event Log entries for the SIEM, and
the detect-and-inform loop described in \cref{sec:disclaimer-loop}.

\subsection{Collectors}

\Cref{tab:collectors} lists the collectors, the Windows source each one reads
and the privilege it needs. Two properties matter. First, each collector reads
state that Windows maintains anyway: the scan creates no network traffic and
leaves no trace beyond its own report. Second, coverage depends on privilege.
A standard user sees only their own processes and profile; the Background
Activity Moderator (BAM) keys and the Prefetch folder require administrator
or SYSTEM rights. In production the scan runs as SYSTEM;
\cref{sec:evaluation} quantifies what is lost without these rights.

\begin{table}[t]
  \centering
  \caption{Collectors of \AD{}. ``Admin'' means that the source is readable
    only with administrator or SYSTEM rights on the Windows build we tested
    (\cref{sec:evaluation}).}
  \label{tab:collectors}
  \small
  \begin{tabularx}{\linewidth}{@{}l X X l@{}}
    \toprule
    Collector & Windows source & Yields & Privilege \\
    \midrule
    Processes & Toolhelp snapshot, \code{QueryFullProcessImageName},
      \code{NtQueryInformationProcess}, \code{GetPackageFullName}
      & Image, path, command line (masked), parent, owner, session, MSIX package & User (own processes) \\
    TCP table & \code{GetExtendedTcpTable} (IPv4, IPv6)
      & Established connections and listening ports, with owning process & User \\
    DNS cache & \code{DnsGetCacheDataTable}, \code{DnsQuery} without wire query
      & Catalog host names currently cached, and their addresses & User \\
    Traces & BAM registry keys, Prefetch file names and times
      & Programs executed recently, per user (BAM) & Admin \\
    Inventory & AppX and Uninstall registry keys, services, extension and
      package folders, configuration and history folders
      & Installed tools, IDE and browser extensions, global npm and pip packages,
      MCP configurations, local models & User (own profile) \\
    Environment & User and machine environment variables in the registry
      & \code{*\_BASE\_URL} hosts, presence of \code{*\_API\_KEY} variables & User \\
    \bottomrule
  \end{tabularx}
\end{table}

\paragraph{Candidate inspection.}
Reading the Authenticode signer of an executable, its loaded modules or the
environment block of a running process is comparatively expensive. \AD{}
therefore inspects only \emph{candidates}: processes whose image, path,
command line or package matches the catalog, processes connected to a catalog
endpoint, processes listening on a port used by a local LLM runtime, and, for
modules and environment, interpreters such as \code{python.exe},
\code{node.exe} or \code{powershell.exe}. The environment block is read from
the process memory (\code{ReadProcessMemory} on the process environment
block) and filtered on the fly: only LLM base-URL variables are kept, with
their host, and API-key variables are kept by name only.

\subsection{Network attribution}\label{sec:attribution}

The TCP table gives, for each connection, the owning process and the remote
address; it does not give the host name. Since almost all LLM services sit
behind TLS, and many behind content-delivery networks shared with unrelated
services, an address alone proves little. \AD{} attributes a connection to a
catalog endpoint in two ways:

\begin{enumerate}
  \item \textbf{By the DNS client cache.} The cache maps the names recently
    resolved by the machine to their addresses. Inverting it gives, for a
    remote address, the catalog host names that resolved to it a few minutes
    earlier. This attribution is strong when the address is dedicated to the
    service and weaker when it belongs to a shared CDN range.
  \item \textbf{By dedicated address ranges.} A few providers publish address
    ranges reserved for their API; the catalog records them, for example the
    range announced for the Anthropic API. A connection to such a range is
    attributed without any DNS information, which also covers software that
    resolves names without the system resolver.
\end{enumerate}

Reading the cache is less simple than it looks. The detector lists the cached
names with \code{DnsGetCacheDataTable}, which Microsoft does not document, and
then looks up the addresses of each catalog name with \code{DnsQuery} in
cache-only mode, which never sends a query on the network. On the Windows
build we tested, the list was empty with or without administrator rights, so
only the exact host names of the catalog could be looked up; host suffixes
such as Azure OpenAI's per-tenant names could not be attributed through DNS.
The lookups themselves are subtle: a cache-only lookup found an entry only
when it used the same record type and query options as the query that had
created it. The resolver of Python, Node.js and .NET (\code{getaddrinfo}
with an unspecified address family) stores a single IPv6 (AAAA) entry,
queried with options that return the IPv4 answers as IPv4-mapped IPv6
addresses. Up to version 0.3.1, \AD{} looked up only plain IPv4 and IPv6
entries and therefore missed the resolutions of most agents, at every
privilege level; we had first attributed the loss to missing privileges. The
evaluated version looks each name up under the three keys that resolvers
use.

Connections from browsers are deliberately ignored: a browser talking to
\code{chatgpt.com} is web use of an AI service, which the web proxy already
governs, and attributing it on the endpoint would flood the results.

\subsection{Correlation model}\label{sec:correlation}

The correlator produces \emph{one finding per tool and per user}. A finding
groups all the evidence about a tool for a user, whatever the collector that
produced it, and carries three attributes.

\begin{description}[leftmargin=1.2em,style=nextline]
  \item[Level] How present the tool is: \emph{active} (running or connected
    now), \emph{recent} (executed within the last 30 days according to the
    execution traces or the history folders; the BAM keys may keep a shorter
    history than Prefetch and the history folders), \emph{installed}, or
    \emph{residue} (leftover data of a removed tool). The level of a finding
    is the highest level of its evidence.
  \item[Confidence] How sure the attribution is: \emph{high} when it rests on a
    package identity, an install path, a command line or a vendor signature;
    \emph{medium} for an executable name, a behaviour or a configuration;
    \emph{low} for circumstantial evidence, such as a connection to a shared
    CDN address.
  \item[Approval] Whether the tool or the endpoint is in the
    \code{sanctioned} list of the catalog.
\end{description}

A finding is \emph{actionable} when it is not approved, at least
\emph{installed}, and of medium or high confidence. Actionable findings raise
warnings in the Event Log and set the exit code. The notice uses a related
but broader trigger (\cref{sec:disclaimer-loop}).

\paragraph{Generic agent rules.}
Catalog matching covers known tools. Product-less agents are caught by
\emph{generic} findings, built from the catalog's generic patterns rather than
from tool signatures:

\begin{itemize}
  \item \code{llm-api}: a non-browser process connected to an LLM endpoint;
    medium confidence on a dedicated address, low otherwise.
  \item \code{probable-agent}: a process that talks to an LLM \emph{and} has a
    shell or interpreter among its children or grandchildren. This rule
    encodes the definition of an agent used in this paper: a model in a
    loop with the ability to act.
  \item \code{relay-candidate}: an LLM base-URL variable, in the environment
    of a process or in the user or machine environment, that points to a host
    absent from the catalog.
  \item \code{local-llm}: a process listening on a port used by local LLM
    runtimes, a client of such a port, or a process that loads an inference
    library such as \code{ggml} or \code{llama}.
  \item \code{mcp-server}: a process whose command line matches the
    conventions of Model Context Protocol servers; it is attached to the
    finding of the agent that started it, when that agent is identified.
  \item \code{browser-automation}: a browser started with automation flags,
    attached to the agent that drives it, or reported with low confidence
    otherwise.
\end{itemize}

Ownership is resolved through the process tree: an MCP server or an automated
browser is attached to the nearest ancestor that already holds an LLM finding,
with a guard against reused process identifiers.

\Cref{alg:correlate} summarises the correlation. Its structure reflects the
principle of \cref{sec:behavioral}: behaviour (line~\ref{alg:shells}) is only
evaluated for processes whose remote service, configuration or local port has
already identified them as talking to a model
(lines~\ref{alg:net}--\ref{alg:local}).

\begin{algorithm}[t]
  \caption{Correlation of a snapshot into findings (simplified).}
  \label{alg:correlate}
  \small
  \begin{algorithmic}[1]
    \Require snapshot $S$ (processes, connections, DNS map, traces, inventory, environment), catalog $K$
    \Ensure findings, one per (tool, user), each with its evidence
    \For{each process $p$ of $S$}
      \For{each tool $t$ of $K$ whose image, path, command line, package or signer matches $p$}
        \State add evidence (\textit{process}, active) to finding $(t, \mathrm{user}(p))$
      \EndFor
    \EndFor
    \For{each established connection $c$ owned by process $p$}\label{alg:net}
      \State $e \gets$ endpoint of $K$ matching the names that resolved to $c.\mathit{remote}$ in the DNS map, else the dedicated range containing it
      \If{$e$ is defined \textbf{and} $p$ is not a browser}
        \If{$p$ matches a tool $t$} \State add evidence (\textit{network}, active) to finding $(t, \mathrm{user}(p))$
        \Else \State add generic finding \code{llm-api}$(e, p)$: medium if $e$ is a dedicated, non-shared entry, else low
        \EndIf
        \State mark $p$ as \emph{LLM-connected}
      \EndIf
    \EndFor
    \For{each base-URL variable $v$ (in a process environment or in the registry)}
      \State \textbf{if} $\mathrm{host}(v)$ is a catalog endpoint \textbf{then} add \code{llm-config} \textbf{else} add \code{relay-candidate} (medium)
      \State mark the process holding $v$, if any, as \emph{LLM-connected}
    \EndFor
    \For{each listener on a local LLM port, and each client of such a port}\label{alg:local}
      \State add \code{local-llm} evidence; mark the client as \emph{LLM-connected}
    \EndFor
    \For{each process $p$ marked \emph{LLM-connected}}
      \If{a shell or interpreter is among the children or grandchildren of $p$}\label{alg:shells}
        \State add evidence (\textit{behaviour}, active) to the finding of $p$, or generic \code{probable-agent}$(p)$ (medium)
      \EndIf
    \EndFor
    \For{each MCP server or automated browser $q$}
      \State attach it to the nearest LLM-connected ancestor of $q$, else add a generic finding
    \EndFor
    \For{each trace, inventory item, configuration or history folder matching a tool $t$}
      \State add evidence to finding $(t, \mathit{user})$, with level recent, installed or residue
    \EndFor
    \For{each finding $f$}
      \State level$(f) \gets$ highest level of its evidence; confidence$(f) \gets$ highest confidence
      \State approved$(f) \gets$ tool or endpoint of $f$ is in the \code{sanctioned} list of $K$
    \EndFor
  \end{algorithmic}
\end{algorithm}

\subsection{The detect-and-inform loop}\label{sec:disclaimer-loop}

\begin{figure}[t]
  \centering
\begin{tikzpicture}[x=0.875cm,
    lane/.style={draw=vizGrid, fill=vizSurface, minimum width=13.8cm, minimum height=2.05cm, anchor=west},
    stepbox/.style={viznode, minimum height=11mm, text width=2.05cm},
    ev/.style={viznode, minimum height=9mm, fill=vizBlue!8, draw=vizBlue, text width=2.05cm},
    lanelabel/.style={font=\scriptsize\bfseries, text=vizInkTwo, rotate=90}]
  \node[lane] at (0, 2.3) {};
  \node[lane] at (0, 0) {};
  \node[lane] at (0, -2.3) {};
  \node[lanelabel] at (0.25, 2.3) {SYSTEM};
  \node[lanelabel] at (0.25, 0) {User};
  \node[lanelabel] at (0.25, -2.3) {SIEM};
  \node[stepbox] (task) at (1.85, 2.3) {Scheduled task\\{\scriptsize at logon + 2 min,\\then hourly}};
  \node[stepbox] (scan) at (4.95, 2.3) {Scan and correlate\\{\scriptsize findings per user}};
  \node[stepbox] (launch) at (8.05, 2.3) {Start the notice\\{\scriptsize in each session with a triggering finding}};
  \node[stepbox, fill=vizOrange!10, draw=vizOrange] (win) at (11.15, 0) {Notice\\{\scriptsize tools, policy, contact}};
  \node[stepbox] (ack) at (14.25, 0) {User ticks the\\box and accepts};
  \node[ev] (e100) at (4.95, -2.3) {Events 100, 300\\{\scriptsize summary, findings}};
  \node[ev] (e400) at (8.05, -2.3) {Event 400\\{\scriptsize notice shown}};
  \node[ev] (e410) at (14.25, -2.3) {Event 410\\{\scriptsize accepted, user SID}};
  \draw[vizarrow] (task) -- (scan);
  \draw[vizarrow] (scan) -- (launch);
  \draw[vizarrow] (launch.east) -| (win.north);
  \draw[vizarrow] (win) -- (ack);
  \draw[vizarrow] (scan) -- (e100);
  \draw[vizarrow] (launch) -- (e400);
  \draw[vizarrow] (ack) -- (e410);
  \draw[vizarrow, densely dotted] (ack.north) -- (14.25, 3.6) -- node[above, vizlabel]
    {next scan: the notice returns as long as the agent is detected} (1.85, 3.6) -- (task.north);
\end{tikzpicture}
  \caption{The detect-and-inform loop. The scan runs as SYSTEM in session~0;
    for each user session with a triggering finding it starts the notice in
    that session with the user's token. Every step leaves an event for the
    SIEM; the acknowledgement is written under the user's own identity.}
  \label{fig:loop}
\end{figure}
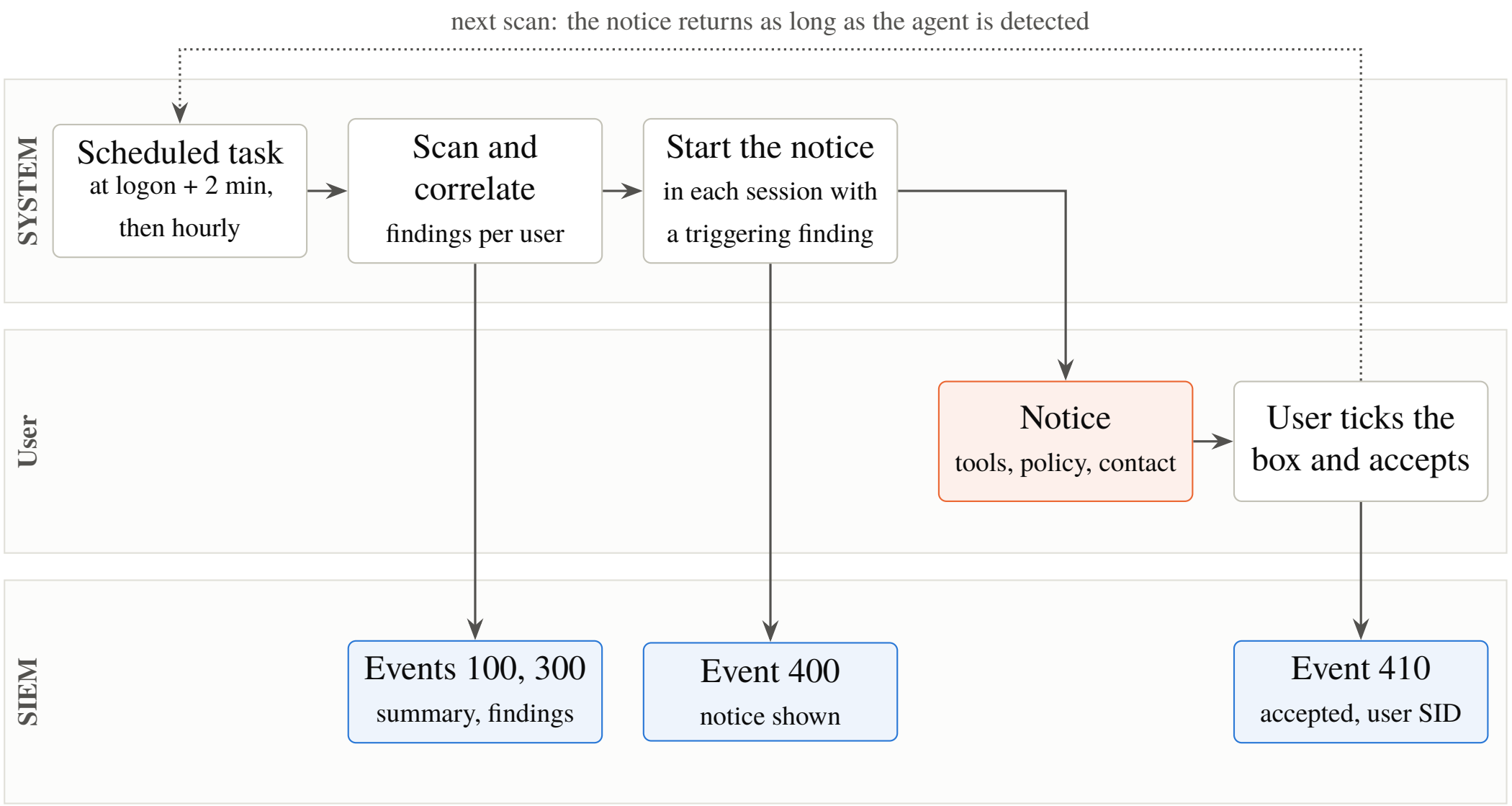

\begin{figure}[t]
  \centering
  \includegraphics[width=0.62\linewidth]{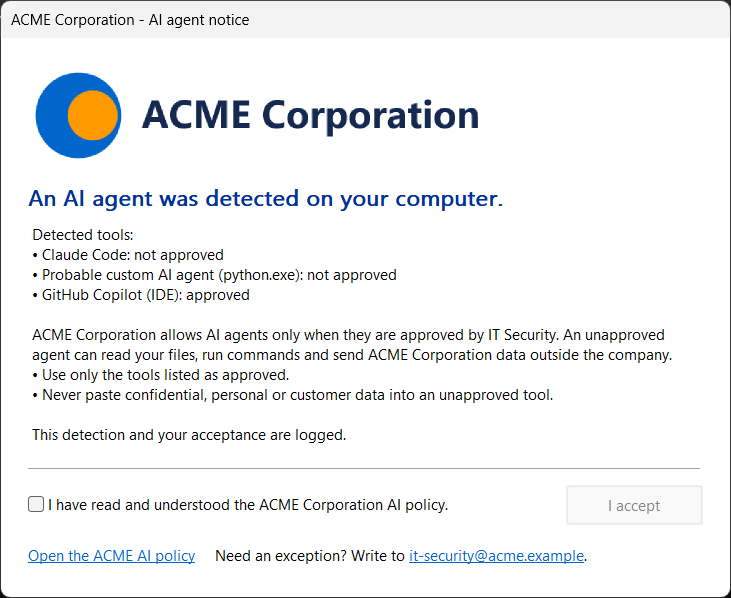}
  \caption{The notice shown to a user when an agent is detected, here with
    the branding of a fictitious company. The user must tick the box before
    the only button is enabled; the window has no close button.}
  \label{fig:disclaimer}
\end{figure}

Detection alone leaves the governance question open: the user does not know
that the tool is unapproved, and the organisation cannot show that the user
was informed. \AD{} closes the loop with a notice that the user must
acknowledge (\cref{fig:loop,fig:disclaimer}).

\begin{enumerate}
  \item \textbf{Trigger.} A finding opens the notice when it is
    \emph{active}, \emph{recent} or \emph{installed} and of medium or high
    confidence and at least one tool of the user is not approved; approved
    tools are then listed alongside. Residues and low-confidence findings
    never open it, and a user of approved tools only is never interrupted.
    (Version 0.3.0 also opened the notice for approved tools alone, which
    would have interrupted every user of a sanctioned tool every hour.)
  \item \textbf{Session launch.} The scan runs as SYSTEM in session~0, which
    cannot draw on a user's desktop. For each active session that holds
    triggering findings, it obtains the user token
    (\code{WTSQueryUserToken}), builds the user environment, and starts its
    own executable in the user's session (\code{CreateProcessAsUser}) with
    the list of tools. It does not wait for the user.
  \item \textbf{Notice.} The window lists the detected tools, each marked
    approved or not approved, followed by the organisation's text, a link to
    its AI policy and a contact address. The texts, the logo and the links
    come from a branding file read at run time.
  \item \textbf{Acknowledgement.} When the user accepts, the window process,
    which runs under the user's identity, writes an event that carries the
    user's security identifier, the version of the text and the tools. The
    organisation can therefore record which text each user acknowledged.
    The record is not proof: any process running as the user can write an
    event under the same source, so event~410 shows that code ran under the
    user's identity, not that the person read the text, and a SIEM should
    accept events 100 and 300 only when they carry the SYSTEM identifier.
    Returning the acknowledgement to the SYSTEM component over an
    authenticated channel would make the record stronger.
  \item \textbf{Repetition.} The notice comes back at every scan for as long
    as the agent is still detected. Acknowledgement is a record, not an
    exemption.
\end{enumerate}

\subsection{Privacy by design}

The detector observes behaviour on employees' machines, so its data
minimisation is part of its design rather than a configuration option.
Prompts, conversations and file contents are never read: history folders are
characterised only by their modification time. Command lines and messages
pass through a masking stage that removes API keys of known formats, bearer
tokens, credentials embedded in URLs and the values of variables whose names
denote secrets. Of a base URL only the host is kept, and of an API-key
variable only its name. The environment block of a candidate process, which
may hold API keys, is read in memory and filtered before anything is
recorded; secrets are read transiently but never stored or transmitted.
Reading another process's memory can itself raise EDR alerts, which an
organisation must expect or exclude. \Cref{sec:discussion} discusses the
legal framework.

\section{Implementation and Deployment}\label{sec:implementation}

\subsection{Implementation}

\AD{} is written in Go and compiled without cgo into a single static Windows
executable of about 3.9\,MiB. It depends only on the Go standard library,
the \code{golang.org/x/sys/windows} package for system calls, and a YAML
parser. The production code amounts to about 5{,}500 non-blank,
non-comment lines in 27 files, and 40 unit tests cover catalog matching,
correlation on synthetic snapshots, the notice trigger rules, branding
validation, logo rendering, parsing helpers, DNS record decoding and secret
masking.

\paragraph{Operating system interfaces.}
Processes are enumerated with a Toolhelp snapshot and inspected with
\code{QueryFullProcessImageName}, \code{NtQueryInformationProcess} for the
command line and the process environment block, \code{GetPackageFullName} for
packaged (MSIX) applications and a module snapshot for loaded libraries.
Executables are identified by their version resource
(\code{GetFileVersionInfo}) and, for candidates, by the subject of their
embedded Authenticode certificate (\code{CryptQueryObject}). A signer counts
only if the signature verifies (\code{WinVerifyTrust}), offline: the file
hash and the certificate chain are checked against the local trust store,
without revocation checks or URL retrieval, so a scan never touches the
network. Without this check, anyone could sign a binary with a self-issued
certificate bearing a vendor's name and obtain a finding of high confidence;
version 0.3.0 parsed signatures without verifying them. Verification hashes
the whole binary, which is costly for large agents, so it runs only when no
package, install path or command line already identifies the tool. Binaries
signed through a catalog file rather than an embedded signature are
identified by other evidence, and packaged applications by their package
family name. The TCP table comes from
\code{GetExtendedTcpTable}, the DNS cache from \code{DnsGetCacheDataTable}
complemented by cache-only \code{DnsQuery} lookups, and execution traces from
the BAM registry keys and the names and timestamps of Prefetch
files~\cite{velociraptor2026bam,neyaz2022prefetch}. \code{DnsGetCacheDataTable}
is not documented by Microsoft, and the detector mirrors a structure that a
Windows release may change. This risk has materialised: on the build we
tested, the function returned no table at all (\cref{sec:attribution}).
The cache-only lookups run in four parallel workers, three per catalog host.

\paragraph{Catalog.}
Everything that names a product or a service lives in a YAML catalog embedded
in the binary (771 lines in the evaluated version). A tool entry lists its
identifiers by kind: process images, install paths, command-line patterns,
package family names, vendor signer and product names, services, IDE and
browser extension identifiers, npm and pip packages, configuration and history
folders, listening ports and API endpoints. The catalog also holds the generic
patterns used by the rules of \cref{sec:correlation}: the names of shells and
interpreters, the command-line conventions of MCP servers, the module names of
inference runtimes, and the patterns of base-URL and API-key variables. An
organisation lists its approved tools, hosts and host suffixes in a
\code{sanctioned} section. The catalog can be replaced at run time for
testing, and a version string identifies it in every report.

\paragraph{Robustness of the scan.}
Each collector runs in its own goroutine and hands back a function that stores
its result; only the main goroutine applies these functions, so a collector
that misses the global deadline is simply ignored and the report states that
the results are partial. A collector that panics is reported as an error
without stopping the others. The report lists every coverage gap, such as the
absence of administrator rights, so that a missing finding is never mistaken
for an absence of agents.

\paragraph{Outputs.}
A scan writes a JSON report (findings with their evidence, coverage notes,
summary counters and, on request, cost metrics), Windows Event Log entries
whose message is a JSON document, and an exit code: 0 when nothing actionable
was found, 1 when at least one unapproved tool was found, 2 on error.
\Cref{tab:events} lists the events, which are designed for SIEM collection.

\begin{table}[t]
  \centering
  \caption{Windows Event Log entries written by \AD{} (Application log, source \code{DetectorAgent}).}
  \label{tab:events}
  \small
  \begin{tabularx}{\linewidth}{@{}l l X@{}}
    \toprule
    Id & Level & Content \\
    \midrule
    100 & Information or Warning & Scan summary and compact list of all findings; one per scan and per machine, so its absence reveals a removed or broken agent \\
    300 & Information or Warning & One finding with its evidence (installed or above, medium or high confidence) \\
    400 & Information & Notice shown to a user, with the tools listed \\
    410 & Information & Notice accepted; written under the user's identity, with the version of the text \\
    900 & Error & Scanner or notice error \\
    \bottomrule
  \end{tabularx}
\end{table}

\subsection{Deployment with Microsoft Intune}

\begin{figure}[t]
  \centering
\begin{tikzpicture}[x=0.86cm,
    stepbox/.style={viznode, minimum height=12mm, text width=2.7cm},
    dev/.style={stepbox, fill=vizBlue!8, draw=vizBlue},
    result/.style={viznode, minimum height=8mm, text width=2.7cm, fill=vizAqua!10, draw=vizAqua},
    optional/.style={stepbox, dashed, text=vizInkTwo},
    zone/.style={draw=vizAxis, rounded corners=4pt, fill=vizSurface, inner sep=2mm},
    zonetitle/.style={font=\small\bfseries, text=vizInk}]
  \node[dev] (build) at (0, 0) {\textbf{Build}\\{\scriptsize tests, build, sign the binary and the scripts}};
  \node[dev] (pkg) at (3.9, 0) {\textbf{Package}\\{\scriptsize binary, scripts and branding as \texttt{.intunewin}}};
  \node[dev] (app) at (8.3, 0) {\textbf{Win32 app}\\{\scriptsize required, all devices, registry detection rule}};
  \node[optional] (rem) at (12.0, 0) {\textbf{Remediations}\\{\scriptsize optional: report per device in the console}};
  \node[stepbox] (inst) at (0, -3.0) {\textbf{Install as SYSTEM}\\{\scriptsize 64-bit, ACLs, event source, task, marker}};
  \node[stepbox] (task) at (3.9, -3.0) {\textbf{Scheduled task}\\{\scriptsize SYSTEM, at logon + 2 min, then hourly}};
  \node[result] (siem) at (8.3, -2.5) {Event Log $\to$ SIEM};
  \node[result] (note) at (8.3, -3.5) {Notice in the user session};
  \node[optional] (det) at (12.0, -3.0) {\textbf{Detection script}\\{\scriptsize scan on the Remediations schedule; no notice, no event}};
  \begin{scope}[on background layer]
    \node[zone, fit=(build)(pkg), label={[zonetitle]above:Administrator}] {};
    \node[zone, fit=(app)(rem), label={[zonetitle]above:Microsoft Intune}] {};
    \node[zone, fit=(inst)(siem)(note)(det), label={[zonetitle]below:Every managed laptop}] {};
  \end{scope}
  \draw[vizarrow] (build) -- (pkg);
  \draw[vizarrow] (pkg) -- (app);
  \draw[vizarrow] (app.south) -- ++(0, -0.75) -| node[pos=0.25, above, vizlabel] {at its next check-in} (inst.north);
  \draw[vizarrow] (inst) -- (task);
  \draw[vizarrow] (task.east) -- (siem.west);
  \draw[vizarrow] (task.east) -- (note.west);
  \draw[vizarrow, densely dotted] (det) -- node[right, vizlabel] {result} (rem);
\end{tikzpicture}
  \caption{Deployment with Microsoft Intune. The administrator builds, signs
    and packages once; every managed laptop installs the package as SYSTEM at
    its next check-in and then scans on schedule.}
  \label{fig:deploy}
\end{figure}
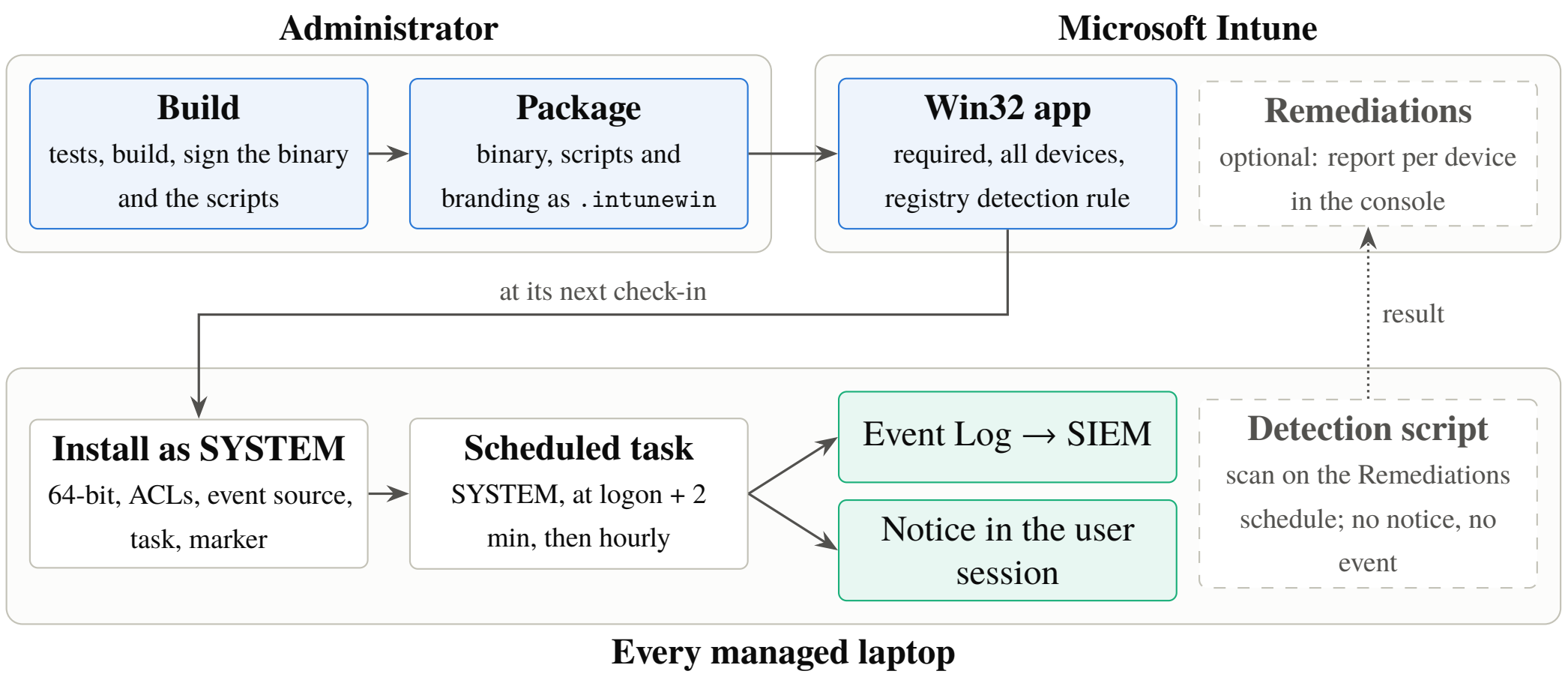

The target environment is a fleet of Windows laptops managed by Microsoft
Intune, the mobile device management service of Microsoft~365
(\cref{fig:deploy}). \AD{} is
packaged as an Intune Win32 application~\cite{microsoft2026win32}: the
executable, an installation script, an uninstallation script and the branding
folder are wrapped into an \code{.intunewin} package and assigned as
\emph{required} to device groups. The installation script, run as SYSTEM:

\begin{enumerate}
  \item restarts itself in 64-bit PowerShell when Intune starts it as a 32-bit
    process~\cite{microsoft2026addwin32}, so that files, registry keys and the scheduled task land in their
    64-bit locations;
  \item copies the executable and the branding folder to \code{Program Files},
    where only administrators can modify them, and refuses an invalid branding
    file;
  \item recreates the report folder under \code{ProgramData} with an access
    control list that only SYSTEM and administrators can write, because
    standard users can pre-create folders there;
  \item registers the Event Log source;
  \item creates a scheduled task that runs the scan as SYSTEM two minutes after
    every logon and then every hour, one instance at a time, with a ten-minute
    limit, on battery too, and that runs a missed scan as soon as the device
    wakes up;
  \item writes, last, an installation marker in the registry. Its value
    combines the version of the executable, the version of the notice
    text and the mode (with or without notice). The Intune detection rule
    compares this value, so that Intune sees the application as installed only
    when every step succeeded, and reinstalls it whenever any of the three
    components changes.
\end{enumerate}

A \emph{silent} mode installs the same task without the notice. It lets
an organisation observe what the detector finds on a pilot population before
any user sees a notice, then switch the population to the notice by changing
the install command and the detection value. An optional detection-only
Intune Remediations script runs a scan without notice or event and reports, per
device, the unapproved tools found, which gives administrators a fleet view in
the Intune console without a SIEM; Remediations require Windows Enterprise or
Education licences, or an equivalent, for the users of the
devices~\cite{microsoft2025remediations}.

\paragraph{Signing.}
Where the PowerShell execution policy \code{AllSigned} is enforced by group
policy, it overrides the \code{-ExecutionPolicy Bypass} argument of the
install command, and unsigned scripts do not run. The build script therefore
signs the executable and the three scripts with the organisation's
code-signing certificate, and the certificate must also be deployed to the
devices' Trusted Publishers store, which Intune can do through a custom
configuration profile. These details matter in practice: a detector that
cannot be installed silently on locked-down machines does not cover them.

\section{Evaluation}\label{sec:evaluation}

We evaluate \AD{} against six research questions:

\begin{description}[leftmargin=1.2em,style=nextline]
  \item[RQ1 -- Coverage.] Which classes of agents does it detect, and at which
    level, including product-less agents, relays, local models, MCP servers and
    automated browsers? How does it fare on a real workstation?
  \item[RQ2 -- Signals.] Which layers of evidence support each detection?
  \item[RQ3 -- False positives.] Does it flag ordinary developer activity?
  \item[RQ4 -- Cost.] How long does a scan take, and how much CPU and memory
    does it use?
  \item[RQ5 -- Robustness.] What happens with an undeclared relay, an agent
    that runs for a few seconds, an agent inside WSL~2, an agent that is
    only intermittently visible, and DNS cache entries that expire?
  \item[RQ6 -- Behavioural baselines.] How do signature-only and
    behaviour-only detectors compare, in particular on behavioural twins?
\end{description}

\subsection{Setup}

\paragraph{Machine and configuration.}
All measurements were taken on a laptop with an 8-core AMD Ryzen~7 4800H
processor and 64\,GB of memory, running Windows~11 \StdOsEdition{}, version
\StdOsVersion{} (build \StdOsBuild{}), with \AD{} version \StdToolVersion{} and
catalog \StdCatalogVersion{}. This is a consumer edition on a recent build: it
is not enrolled in Intune and cannot enforce group policy, so the deployment
of \cref{sec:implementation} was not exercised on it, and its privilege
behaviour may differ from the Pro and Enterprise editions of enterprise
fleets. We ran the whole bench in two configurations: with \emph{standard
user} rights, and from an elevated \emph{administrator} window. On this
build, the BAM keys, the Prefetch folder and the processes of other users
could be read only with elevation, and the reports of the standard-user
scans state these gaps. The elevated configuration approaches the production
scheduled task, which runs as SYSTEM, without being identical to it
(\cref{sec:eval-admin}). Unless stated otherwise, the results below are for
the standard user, the configuration that sees less. The machine is the first author's
development workstation, with real AI tools installed and running, which we
use for an in-situ cross-check.

\paragraph{The evaluated version, and how the bench shaped it.}
All results are for version \StdToolVersion{}, and two of its rules were
changed because of this bench. The first version we evaluated (0.3.0) flagged
benign episode B09 as an agent, because it treated a process carrying a
base-URL variable as connected to an LLM; version 0.3.1 requires an actual
connection and ignores values inherited from the user's persistent
environment. Version 0.3.1 then detected none of the agents whose provider
sits behind a shared content-delivery network (CDN), in either
configuration. We had first attributed this gap to privilege, since the
standard-user reports said so; the administrator run, which showed the same
gap, is what revealed the real cause, the key under which the DNS client
stores its entries (\cref{sec:attribution}), and version 0.3.2 fixed it.
Because these corrections were made after observing the bench, the episodes
that revealed them (B09 and the four CDN-hosted agents) are no longer
held-out tests: they show that the corrections work, not how the detector
fares on cases nobody has looked at.

\paragraph{The bench.}
The evaluation bench is part of the published artifact~\cite{canale2026repo}.
It runs \emph{episodes}: each episode starts one or more processes, waits
8\,s for them to settle, observes them for 24\,s, and stops them. During the
observation window, the baseline detectors sample the process table and the
TCP table about once per second, and \AD{} scans three times. A detection is
attributed to an episode when it concerns a process of the episode's tree: its
root processes and their descendants created after the episode started. For
\AD{}, an episode counts as detected when at least one finding of medium or
high confidence carries evidence from that tree; all findings of the
episodes are at the \emph{active} level, so this is also the threshold that
opens the notice. The sampling is asymmetric: the baselines look about once
per second, \AD{} three times per window, and once per hour in production,
which favours the baselines on transient patterns. The bench also measures a
favourable case for network attribution: every scan follows the start of
the agent by seconds, while the agent's own name resolution is still in the
DNS cache; \cref{sec:eval-rq5} measures what happens later.

We ran the whole bench \StdRuns{} times in each configuration (standard user:
runs \StdRunFirst{} to \StdRunLast{}; administrator: \AdmRunFirst{} to
\AdmRunLast{}). For each episode and detector, \cref{tab:episodes} reports
whether the detection occurred in every run, in none, or in some, and the
metrics use the majority outcome. Every run records its provenance (Windows
build, source commit, SHA-256 of the executable, beginning with
\code{\StdBinarySha{}} in both configurations, and catalog version), and the
scoring script refuses to combine runs that differ in any of them or that
miss an episode.

\paragraph{Synthetic agents.}
To exercise every class without accounts, costs or real model calls, the
bench uses synthetic agents written in Python, Node.js and PowerShell. Each one
reproduces the footprint that matters on the endpoint: it keeps a TLS
connection to a real API endpoint, polled every five seconds with an
\emph{invalid key} (the service answers HTTP 401; no prompt or data is sent,
nothing is billed), and it keeps a persistent shell to which it sends a
harmless command at every poll. This makes the agents easy to see: both
signals are present all the time. Real agents are often intermittent: some
start a process per command instead of keeping a shell, HTTP clients close
idle connections after a few seconds, and users pause between requests.
Episode A12 reproduces this pattern: a call every ten seconds, a connection
closed five seconds after each call, and one short-lived shell per call.
Options add an MCP server (the reference filesystem server, started with
\code{npx}), a headless browser with a remote-debugging port, a local
LLM-style HTTP server on Ollama's port, or a module named like an inference
runtime.

We separate two kinds of positives. An \emph{agent} (A01--A12) matches the
definition of \cref{sec:agents}: it reaches a model and acts through tools.
An \emph{AI component} (C01--C03) is AI software that is not an agent by
itself: a process loading an inference runtime, an MCP server alone, and a
client that calls a model API without any tool. Components are shadow AI,
and \AD{} reports them, but a detector of agents need not flag them, so
precision and recall are computed on agents only and components are
reported apart.

\paragraph{Benign workloads and behavioural twins.}
The benign episodes are of two kinds. Three \emph{behavioural twins} run the
same agent scripts against non-LLM APIs (the PyPI JSON API, the npm registry,
Microsoft Graph): identical process trees, connections and timing, a different
remote service. Five developer workloads repeat ordinary tasks in a loop:
\code{npm install} of a package with an install script, a shallow
\code{git clone} over HTTPS, a \code{pip} download, a .NET restore and
build, and a Go build with module download, each with an empty cache so that
it really uses the network. A ninth benign episode (B09) targets a
false-positive mode of \AD{} itself: a PowerShell loop running \code{git}
that inherited an \code{OPENAI\_BASE\_URL} variable, as every process does
when a developer sets such a variable at user level to reach a corporate
LLM gateway.

\paragraph{Baselines.}
Three baselines run on the same samples, without any code from \AD{}.
\textbf{D0}, a signature detector, flags a process whose image name is one of
the executable names of the catalog, used alone. \textbf{D1}, the behavioural
rule of \cref{sec:behavioral}, flags an interpreter or runtime host (Python,
Node.js, PowerShell, .NET, Java and similar) that holds an established
connection to a remote port 443, taken as TLS, and has a shell among its
children or grandchildren. \textbf{D2}
drops the interpreter condition and flags any process with these properties.

\paragraph{An experimenter effect, and how we removed it.}
The bench was itself driven by an AI coding agent. In a first run, an MCP
server started by an episode was attributed to that coding agent, its
genuine ancestor, because \AD{} attaches MCP servers to their nearest
LLM-connected ancestor. The attribution was correct for the machine but
wrong for the experiment, and a production scan never has the bench as an
ancestor. The bench therefore starts every episode through a short-lived
launcher, so that episode processes have no living ancestor, and all results
below come from that configuration.

\subsection{RQ1: Coverage}

\Cref{tab:episodes} lists all episodes and the detectors that caught them.
With standard-user rights, \AD{} detected \StdADTp{} of the \StdAgents{}
agent episodes in the majority of runs, and \StdADComponents{} of the
\StdComponents{} AI components in every run. The detected agents cover every
class of the bench: agents that reach a provider with a dedicated address
range or one behind a shared CDN, written in Python, Node.js and PowerShell
or run by a renamed copy of the Node.js runtime; an agent configured through
\code{OPENAI\_BASE\_URL} to use a relay unknown to the catalog; a local
LLM-style server and its client; agents with an MCP server or an automated
browser; and the intermittent agent. The components were found through a
module name, an MCP command line and a connection to the Anthropic API.

The evidence differs by provider. Anthropic publishes a dedicated address
range, so a connection to it yields an \code{llm-api} finding of medium
confidence on its own. The four CDN-hosted providers do not: a connection
attributed through the DNS cache yields an \code{llm-api} finding of low
confidence, below the detection threshold, and these agents count as
detected only because they also run a shell (the \code{probable-agent}
rule).

Two agents were detected in only three runs out of five, and each
instability has a cause that matters beyond the bench. The OpenAI agent
(A01) was missed twice. OpenAI's answers live at most 60\,s, and when the
agent receives one near the end of its life, the cache entry expires before
the first scan, 8.6\,s after the start; this is the expiry effect of
\cref{sec:eval-rq5}, visible even in the favourable case. The intermittent
agent (A12) was caught in three runs as standard user and in none as
administrator; \cref{sec:eval-rq5} shows that this depends on the phase
between two regular schedules, not on privilege. The baselines were stable
in the standard-user runs.

\begin{table}[p]
  \centering
  \caption{Episodes and detections with standard-user rights over
    \StdRuns{} runs: \cmark{} detected in every run, \xmark{} in none, $k/n$
    in $k$ of $n$ runs. A: agents; C: AI components; B: benign; R:
    robustness. D0: signature; D1: interpreter + port 443 + shell; D2: any
    process + port 443 + shell; AD: \AD{} (finding of medium or high
    confidence). The last column lists the kinds of findings \AD{}
    produced.}
  \label{tab:episodes}
  \scriptsize
  \setlength{\tabcolsep}{3pt}
  \begin{tabular}{@{}l >{\raggedright\arraybackslash}p{2.4cm} >{\raggedright\arraybackslash}p{4.4cm} c c c c >{\raggedright\arraybackslash}p{3.0cm}@{}}
\toprule
Id & Class & Episode & D0 & D1 & D2 & AD & Agent Detector findings \\
\midrule
A01 & Model provider (CDN-hosted) & Python agent, OpenAI API, persistent cmd & \xmark & \cmark & \cmark & {\scriptsize 3/5} & probable-agent \\
A02 & Model provider (dedicated range) & Python agent, Anthropic API, persistent cmd & \xmark & \cmark & \cmark & \cmark & llm-api, probable-agent \\
A03 & Model provider (dedicated range) & Node.js agent, Anthropic API, persistent PowerShell & \xmark & \cmark & \cmark & \cmark & llm-api, probable-agent \\
A04 & Model provider (CDN-hosted) & PowerShell agent, Mistral API, persistent cmd & \xmark & \cmark & \cmark & \cmark & probable-agent \\
A05 & Aggregator & Python agent, OpenRouter, persistent cmd & \xmark & \cmark & \cmark & \cmark & probable-agent \\
A06 & Inference platform & Python agent, Groq API, persistent cmd & \xmark & \cmark & \cmark & \cmark & probable-agent \\
A07 & Unknown relay (declared) & Python agent through a relay set in OPENAI\_BASE\_URL & \xmark & \cmark & \cmark & \cmark & probable-agent, relay-candidate \\
A08 & Local LLM & Local LLM-style server on port 11434 and an agent using it & \xmark & \xmark & \xmark & \cmark & local-llm-server, probable-agent \\
A09 & Agent with MCP server & Python agent, Anthropic API, reference MCP filesystem server & \xmark & \cmark & \cmark & \cmark & llm-api, probable-agent \\
A10 & Agent driving a browser & Python agent, Anthropic API, headless Chrome with remote debugging, no shell & \xmark & \xmark & \xmark & \cmark & llm-api \\
A11 & Renamed runtime & Node.js agent run by a copy of node.exe renamed updater.exe & \xmark & \xmark & \cmark & \cmark & llm-api, probable-agent \\
A12 & Intermittent agent & Python agent, Anthropic API, a call every 10 s, connection closed after 5 s, one short shell per call & \xmark & \cmark & \cmark & {\scriptsize 3/5} & llm-api \\
\midrule
C01 & Inference runtime & Process loading a module named like an inference runtime (ggml), no network & \xmark & \xmark & \xmark & \cmark & local-llm-runtime \\
C02 & MCP server & Reference MCP filesystem server alone, no LLM connection & \xmark & \cmark & \cmark & \cmark & mcp-server \\
C03 & LLM client without tools & Python script polling the Anthropic API, no shell, no tools & \xmark & \xmark & \xmark & \cmark & llm-api \\
\midrule
B01 & Behavioural twin & Twin of A02: Python script, PyPI JSON API, persistent cmd & \xmark & \cmark & \cmark & \xmark & -- \\
B02 & Behavioural twin & Twin of A03: Node.js script, npm registry, persistent PowerShell & \xmark & \cmark & \cmark & \xmark & -- \\
B03 & Behavioural twin & Twin of A04: PowerShell script, Microsoft Graph, persistent cmd & \xmark & \cmark & \cmark & \xmark & -- \\
B04 & Developer workload & npm install with install script (cmd) & \xmark & \cmark & \cmark & \xmark & -- \\
B05 & Developer workload & git clone over HTTPS & \xmark & \xmark & \xmark & \xmark & -- \\
B06 & Developer workload & pip download of a large wheel & \xmark & \xmark & \xmark & \xmark & -- \\
B07 & Developer workload & dotnet restore and build & \xmark & \xmark & \xmark & \xmark & -- \\
B08 & Developer workload & go build with module download & \xmark & \xmark & \xmark & \xmark & -- \\
B09 & Inherited base URL & git loop in a PowerShell that inherited OPENAI\_BASE\_URL from the user environment & \xmark & \xmark & \xmark & \xmark & -- \\
\midrule
R01 & Undeclared relay & Python agent through a relay hard-coded in the script & \xmark & \cmark & \cmark & \xmark & -- \\
R02 & Ephemeral agent & Python agent running 5 s, scanned after it exited & \xmark & \xmark & \xmark & \xmark & -- \\
R03 & Agent inside WSL 2 & Python agent running in the Ubuntu WSL 2 distribution & \xmark & \xmark & \xmark & \xmark & -- \\
\bottomrule
\end{tabular}

\end{table}

\paragraph{In-situ cross-check.}
On the workstation itself, with the tools actually installed,
\StdInsituCorrect{} of the \StdInsituTotal{} expected tools were reported at
the expected level (\cref{tab:insitu}). The expectation comes from a
separate script that lists installed packages, extension folders, global
npm packages and running processes. That script does not share code with
\AD{}, but it reads the same kinds of sources, so the check is a
cross-check of the detector's logic, not an independent ground truth; it
covers only the tools we expected, and it cannot verify recent use. Three tools
were active (Claude Code, GitHub Copilot in VS Code, the Codex desktop
application), Claude Desktop was installed and had been used within the trace
window according to its log folder, OpenCode was installed, and three tools
had left residues after being uninstalled. An earlier catalog disagreed on
one tool, and the disagreement is instructive: \AD{} reported a residue of
the GitHub Copilot command-line agent, which is not installed, because its
configuration folder \code{\textasciitilde/.copilot} is now also used by the
Copilot runtime built into VS Code. We removed that folder from the evidence
of the command-line agent, and kept the row in the table as a check. A
vendor that moves or shares its state folder silently changes the meaning of
a catalog entry; here the consequence was limited, because residues never
open the notice.

\begin{table}[t]
  \centering
  \caption{In-situ cross-check: findings on the first author's workstation
    (standard user) against the expectation of a separate inventory script.}
  \label{tab:insitu}
  \scriptsize
  \begin{tabular}{@{}>{\raggedright\arraybackslash}p{3.2cm} >{\raggedright\arraybackslash}p{5.4cm} l l c@{}}
\toprule
Tool & Evidence found by the inventory script & Expected & Reported & \\
\midrule
Claude Code & claude.exe running; VS Code extension anthropic.claude-code; global npm package @anthropic-ai/claude-code & active & active, high & \cmark \\
GitHub Copilot (IDE) & copilot-runtime.exe running in VS Code & active & active, high & \cmark \\
OpenAI Codex / ChatGPT app & MSIX package OpenAI.Codex installed; codex.exe and ChatGPT.exe running & active & active, high & \cmark \\
Claude Desktop (incl. Cowork) & MSIX package Claude installed (use within 30 days not independently verifiable) & installed & recent, high & \cmark \\
OpenCode & installed program OpenCode 1.18.3 & installed & installed, high & \cmark \\
Cursor & \textasciitilde{}/.cursor folder present; Cursor not installed & residue & residue, medium & \cmark \\
Ollama & \textasciitilde{}/.ollama folder present; Ollama not installed & residue & residue, medium & \cmark \\
Perplexity Comet (agentic browser) & empty \%LOCALAPPDATA\%\textbackslash{}Perplexity folder; Comet not installed & residue & residue, medium & \cmark \\
github-copilot-cli & Copilot CLI not installed; \textasciitilde{}/.copilot is used by the Copilot runtime of VS Code & absent & -- & \cmark \\
\bottomrule
\end{tabular}

\end{table}

\subsection{RQ2: Signals}

\Cref{tab:signals} shows which kinds of evidence supported the detections.
Two observations stand out. First, the generic rules carried every detection
of a synthetic agent: none of them matched a product signature, by
construction, yet each was found through network attribution, behaviour,
configuration, a local port, a module, an MCP command line or a browser flag.
Second, no single layer covers all classes: the network layer is useless for
local models and for relays it does not know; the configuration layer catches
the declared relay but not the hard-coded one; the module and port signals
catch local inference but nothing else. Detection coverage comes from the
union of layers, and confidence from their conjunction.

\begin{table}[t]
  \centering
  \caption{Evidence kinds behind the detected agent and component episodes
    (findings of medium or high confidence; an episode can have several).}
  \label{tab:signals}
  \small
  \begin{tabular}{@{}l l r >{\raggedright\arraybackslash}p{5.4cm}@{}}
\toprule
Evidence kind & Layer & Episodes & Which \\
\midrule
LLM endpoint & L3 network & 7 & A02, A03, A09, A10, A11, A12, C03 \\
LLM + shell & L2 runtime & 10 & A01, A02, A03, A04, A05, A06, A07, A08, A09, A11 \\
base URL & L2/L5 configuration & 1 & A07 \\
local LLM port & L3 network & 1 & A08 \\
inference module & L2 runtime & 1 & C01 \\
MCP server & L2 runtime & 2 & A09, C02 \\
automated browser & L2 runtime & 1 & A10 \\
\bottomrule
\end{tabular}

\end{table}

\subsection{RQ3: False positives}

\AD{} flagged \StdADFp{} of the \StdBenign{} benign episodes. It flagged none
of the three behavioural twins, whose behaviour is identical to that of
agents, and none of the five developer workloads, including \code{npm
install}, which runs shells while downloading over TLS. Nor did it flag B09,
the \code{git} loop whose PowerShell inherited an \code{OPENAI\_BASE\_URL}
variable, but this episode is no longer a fair test. Version 0.3.0 flagged it
as a probable agent with a relay candidate, because it treated a process
that carries a base-URL variable as connected to an LLM, and \code{git}
counts among the shells it looks for. This was a real false-positive mode,
not an artefact of the bench: a developer who sets such a variable at user
level, to reach a corporate gateway, makes every interpreter that runs a
command look like an agent. The rule now requires an outbound connection and
ignores values inherited from the persistent environment. The change has a
cost: a relay declared in the persistent environment is still reported, at
the \emph{installed} level, but the processes that inherit the variable are
no longer marked as connected to an LLM, so an agent that relies on such a
relay is found only if its connection is identified otherwise. On the
workstation, all findings concerned AI tools
(\cref{tab:insitu}). The baselines tell a different story, discussed under
RQ6.

\subsection{RQ4: Cost}

\Cref{tab:overhead} breaks down the cost of \StdOverheadRuns{} consecutive
scans on the workstation, which ran a median of \StdProcessesMedian{}
processes with \StdConnectionsMedian{} TCP connections. A scan took
\StdDurationmsMedian{}\,ms at the median (\StdDurationmsPNineFive{}\,ms at the
95th percentile, \StdWallmsMedian{}\,ms including process start as seen from
outside), consumed \StdCputimemsMedian{}\,ms of CPU time across its parallel
collectors, and peaked at \StdPeakworkingsetmbMedian{}\,MB of working set. At
one scan per hour, this amounts to less than ten seconds of CPU per laptop
and per day. These figures are for a standard user with warm caches: BAM
and Prefetch were unreadable and the processes of other users could not be
inspected. With these sources read, as administrator, a scan took
\AdmDurationmsMedian{}\,ms (\cref{sec:eval-admin}). The most expensive
stages are candidate inspection, the cache-only DNS lookups, inventory,
correlation and process enumeration, each of the order of a few tens of
milliseconds.

\begin{table}[t]
  \centering
  \caption{Cost of a scan (standard user, \StdOverheadRuns{} consecutive
    scans). Collectors run in parallel, so their times do not add up to the
    total.}
  \label{tab:overhead}
  \small
  \begin{tabular}{@{}l r r@{}}
\toprule
Stage & Median (ms) & p95 (ms) \\
\midrule
Processes & 23.2 & 27.5 \\
TCP table & 2.2 & 4.6 \\
DNS cache & 30.0 & 32.2 \\
Traces & 0.0 & 0.6 \\
Inventory & 29.0 & 31.4 \\
Environment & 0.5 & 1.2 \\
Candidate inspection & 53.8 & 58.7 \\
Correlation & 27.2 & 28.3 \\
\midrule
Scan duration (elapsed) & 154 & 166 \\
CPU time (all threads) & 406 & 531 \\
\bottomrule
\end{tabular}

\end{table}

\begin{figure}[t]
  \centering
  \begin{tikzpicture}
    \begin{axis}[vizaxis,
        xbar, bar width=8pt, width=0.62\linewidth, height=5.8cm,
        xlabel={Time per scan (ms)}, xmin=0, xmax=65,
        ytick=data, yticklabels from table={eval/results/tex/overhead-stages-standard.dat}{label},
        y tick label style={font=\small, text=vizInk},
        enlarge y limits={abs=0.6},
        axis y line*=left, axis x line*=bottom, ymajorgrids=false, xmajorgrids=true]
      \addplot[fill=vizBlue, draw=none,
          error bars/.cd, x dir=plus, x explicit,
          error bar style={draw=vizInkTwo, line width=0.8pt}, error mark options={rotate=90, mark size=2.5pt}]
        table[x=median, y expr=\coordindex, x error expr=\thisrow{p95}-\thisrow{median}]
        {eval/results/tex/overhead-stages-standard.dat};
    \end{axis}
  \end{tikzpicture}
  \caption{Time of each stage of a scan (standard user, \StdOverheadRuns{}
    scans): the bar is the median, the whisker reaches the 95th percentile.
    The collectors run in parallel; candidate inspection and correlation
    follow them. Values are in \cref{tab:overhead}.}
  \label{fig:stages}
\end{figure}
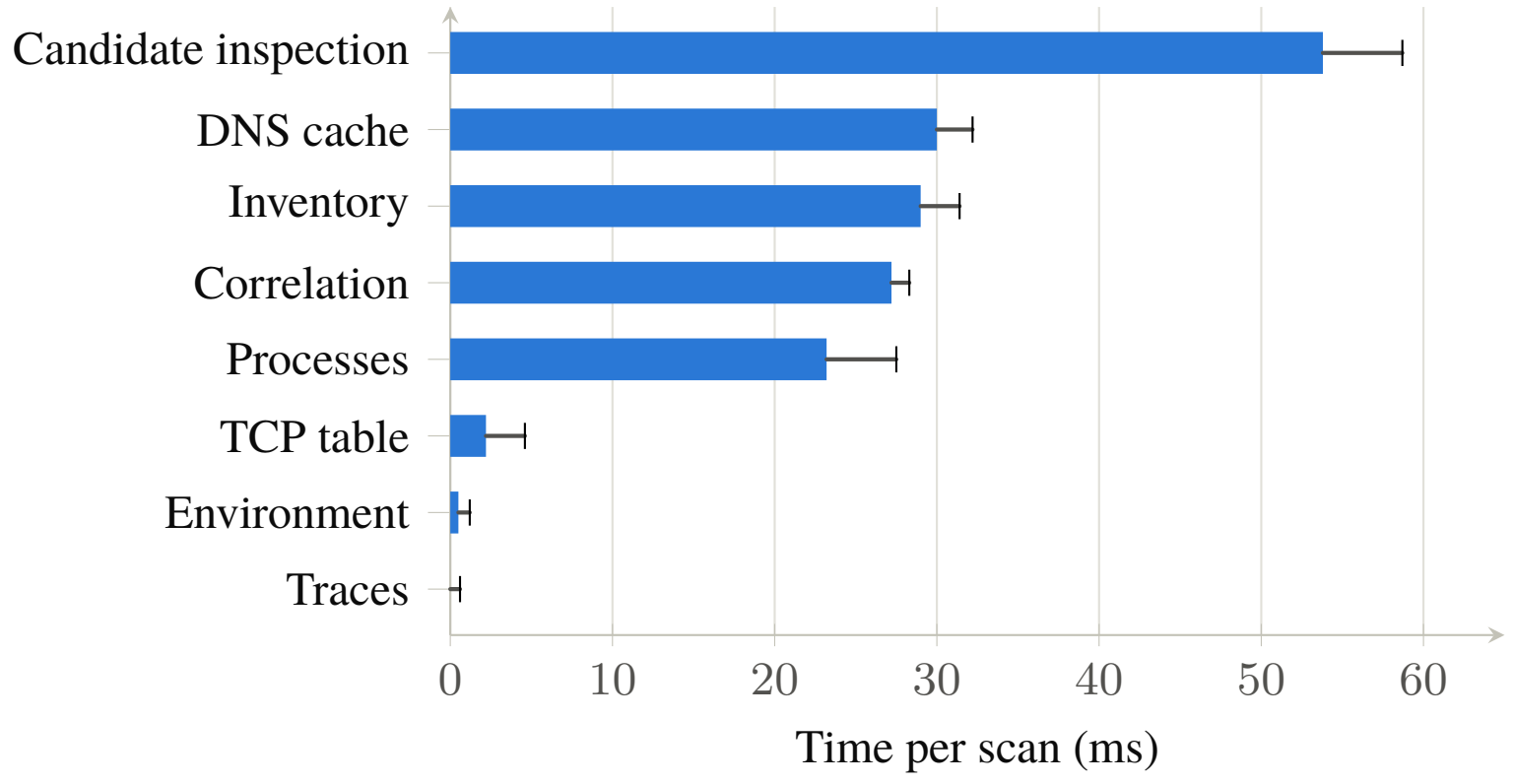

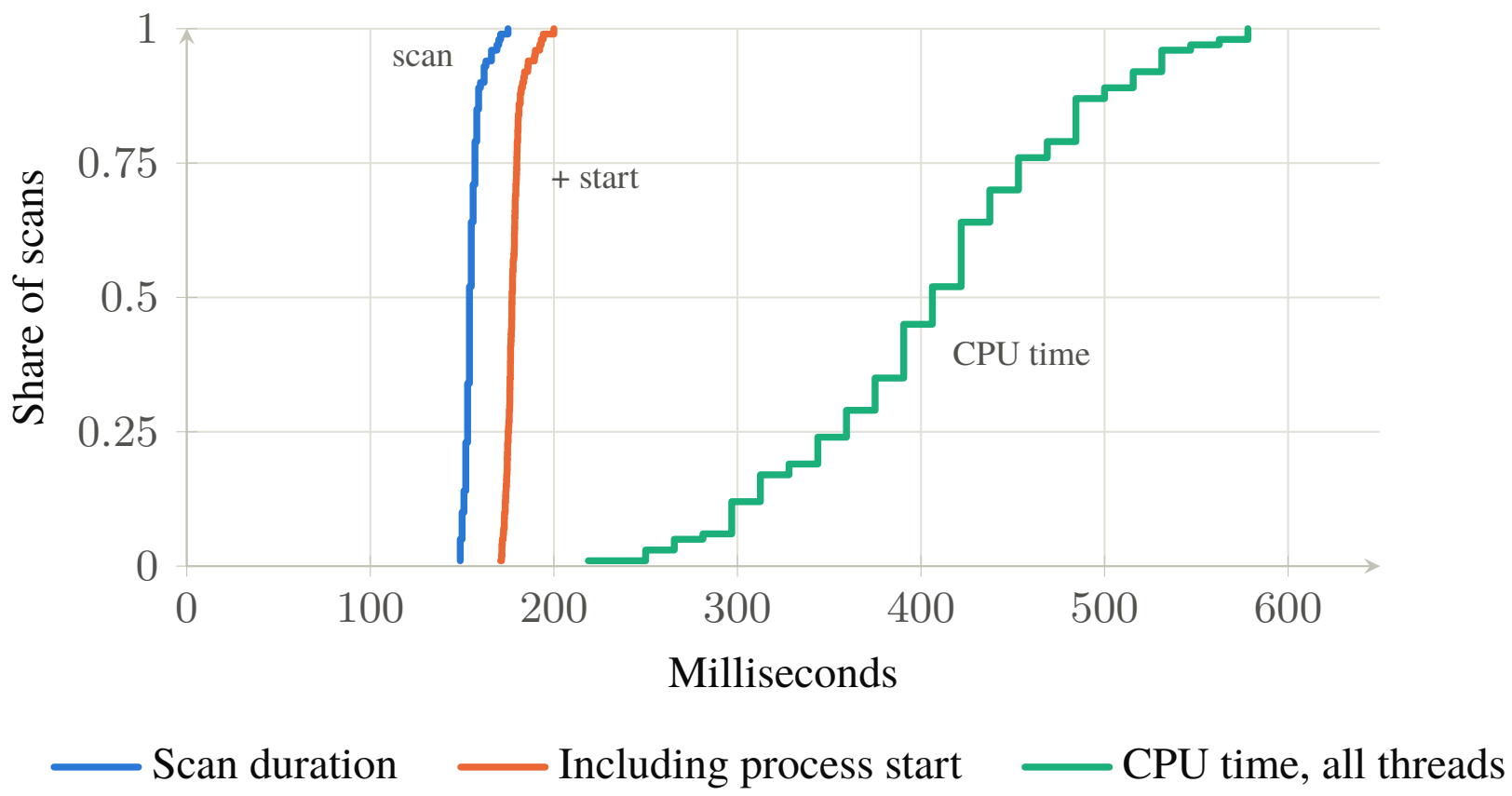
\begin{figure}[t]
  \centering
  \begin{tikzpicture}
    \begin{axis}[vizaxis,
        width=0.72\linewidth, height=5.4cm,
        xlabel={Milliseconds}, ylabel={Share of scans},
        xmin=0, xmax=650, ymin=0, ymax=1.0, ytick={0,0.25,0.5,0.75,1},
        legend columns=3,
        legend style={at={(0.5,-0.3)}, anchor=north,
          /tikz/every even column/.append style={column sep=10pt}}]
      \addplot[vizBlue, const plot] table[x=duration, y=frac] {eval/results/tex/overhead-standard.dat}
        node[pos=0.6, anchor=east, font=\scriptsize, text=vizInkTwo, xshift=-3pt] {scan};
      \addlegendentry{Scan duration}
      \addplot[vizOrange, const plot] table[x=wall, y=frac] {eval/results/tex/overhead-standard.dat}
        node[pos=0.3, anchor=west, font=\scriptsize, text=vizInkTwo, xshift=3pt] {+ start};
      \addlegendentry{Including process start}
      \addplot[vizAqua, const plot] table[x=cpu, y=frac] {eval/results/tex/overhead-standard.dat}
        node[pos=0.5, anchor=north west, font=\scriptsize, text=vizInkTwo, xshift=3pt] {CPU time};
      \addlegendentry{CPU time, all threads}
    \end{axis}
  \end{tikzpicture}
  \caption{Cumulative distribution of the cost of \StdOverheadRuns{}
    consecutive scans (standard user, warm caches). The scan duration is
    reported by the scanner; the second curve adds the start of the process,
    measured from outside. Windows counts CPU time in clock ticks of
    15.6\,ms, hence the steps of the third curve.}
  \label{fig:overhead}
\end{figure}

\Cref{fig:stages} ranks the stages by their median time; \cref{fig:overhead}
shows the distribution of the totals: scan times are tightly grouped,
with no long tail, because every collector reads bounded, local state and the
candidate inspection is limited to the processes that matter.

\subsection{RQ5: Robustness}\label{sec:eval-rq5}

The three robustness episodes probe known limits (lower part of
\cref{tab:episodes}), and two separate experiments measure the effect of
time: an agent that is visible only part of the time, and DNS cache entries
that expire.

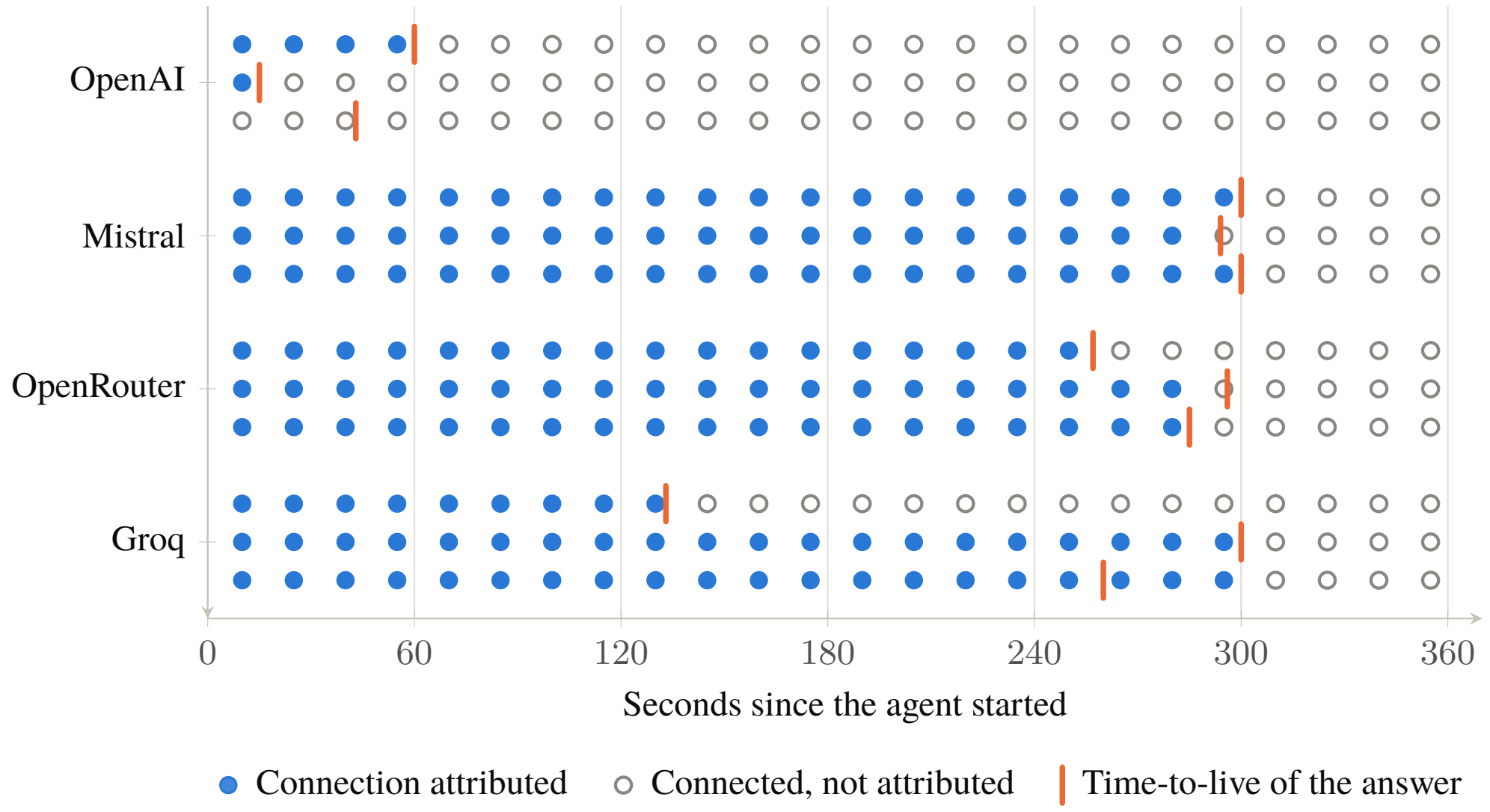
\begin{figure}[t]
  \centering
  \begin{tikzpicture}
    \begin{axis}[vizaxis,
        width=0.92\linewidth, height=7cm,
        xlabel={Seconds since the agent started}, xmin=0, xmax=370,
        xtick={0,60,120,180,240,300,360},
        ymin=0, ymax=16, y dir=reverse,
        ytick={2,6,10,14}, yticklabels={OpenAI, Mistral, OpenRouter, Groq},
        y tick label style={font=\small, text=vizInk},
        axis y line*=left, axis x line*=bottom, ymajorgrids=false, xmajorgrids=true,
        unbounded coords=discard,
        legend columns=3,
        legend style={at={(0.5,-0.22)}, anchor=north,
          /tikz/every odd column/.append style={column sep=3pt},
          /tikz/every even column/.append style={column sep=10pt}}]
      \addplot[only marks, mark=*, mark size=2pt, color=vizBlue, line width=0.6pt]
        table[x=t, y=ya] {eval/results/tex/dns-ttl-strip-standard.dat};
      \addlegendentry{Connection attributed}
      \addplot[only marks, mark=o, mark size=2pt, color=vizMuted, line width=0.8pt]
        table[x=t, y=yc] {eval/results/tex/dns-ttl-strip-standard.dat};
      \addlegendentry{Connected, not attributed}
      \addplot[only marks, mark=|, mark size=4.5pt, line width=1.4pt, color=vizOrange]
        table[x=ttl, y=y] {eval/results/tex/dns-ttl-marks-standard.dat};
      \addlegendentry{Time-to-live of the answer}
    \end{axis}
  \end{tikzpicture}
  \caption{Attribution of a long-lived connection through the DNS cache
    (standard user). For each CDN-hosted provider, a synthetic agent keeps
    one connection open and \AD{} scans every 15\,s; each row is one of three
    repetitions. A filled dot means that the scan attributed the agent's
    connection to the provider's host name, an open dot that the agent was
    connected but the connection was not attributed. The tick marks the
    time-to-live of an answer obtained just before the agent started; the
    agent's own answer, obtained a second later, may come from another cache
    and live a little longer or shorter. Attribution ends when the cache
    entry expires, while the connection lives on.}
  \label{fig:dnsttl}
\end{figure}

\begin{itemize}
  \item \textbf{Undeclared relay.} An agent that reaches its model through a
    relay hard-coded in its source, on a domain absent from the catalog, with
    no base-URL variable, is invisible to \AD{}: nothing identifies the remote
    service as an LLM. The behavioural rules flag it, but, as RQ6 shows, they
    flag its benign twin just as readily.
  \item \textbf{Ephemeral agent.} An agent that runs for five seconds and is
    gone at scan time is missed by every periodic detector. This is not a
    corner case: it is how many home-made scripts run. With administrator
    rights, the execution traces record that \code{python.exe} ran, which
    says nothing about an agent; continuous telemetry would see it
    (\cref{sec:landscape}).
  \item \textbf{Agent in WSL~2.} An agent running inside the Ubuntu
    distribution of WSL~2 is invisible to all detectors: its processes and
    sockets live in the WSL virtual machine, and the host's tables show
    neither.
  \item \textbf{Intermittent agent.} Episode A12 was caught in three of the
    five standard-user runs and in none of the administrator runs, and the
    cause lies in the bench, not in privilege. Its scans fall at nearly the
    same moments of every episode, and the agent's calls are just as
    regular, so every run samples almost the same phase. Across the fifteen
    runs of versions 0.3.1 and 0.3.2 in both configurations, A12 was
    detected exactly when the third scan started less than about 26.0\,s
    after the episode began, and missed when it started up to a second
    later: the agent closes its connection at about that moment. To measure
    detection rather than phase, we ran the same agent for about seven
    minutes and scanned it \StdIntScans{} times at random intervals of 4 to
    11\,s. The agent held a connection at \StdIntConnected{} of the scans,
    and \AD{} reported it at \StdIntActionable{} of them
    (\StdIntActionableRatio{}, 95\% CI [\StdIntActionableLo{},
    \StdIntActionableHi{}]), always through its connection to the Anthropic
    API; the \code{probable-agent} finding, which needs the connection and
    one of the short-lived shells at the same instant, appeared at only
    \StdIntProbableagent{}. A scan at a random moment thus sees this agent
    about as often as its connection is open, here a little more than one
    scan in three while it works. Behind a CDN, where a connection alone
    gives only a low-confidence finding, the same agent would be detected
    only at the rare scans that also find one of its shells.
  \item \textbf{Expiring DNS entries.} A connection to a CDN-hosted provider
    is attributable only while the name it was opened with is in the DNS
    cache, and an agent that keeps its connection open never resolves the
    name again. We ran one such agent for each of the four CDN-hosted
    providers, scanned every 15\,s for six minutes, and repeated the
    experiment three times (\cref{fig:dnsttl}). All twelve agents stayed
    connected to the end, but no connection was attributed after
    \StdTtlLastOverall{}\,s: attribution ended with the cache entry, after
    two to five minutes for Mistral, OpenRouter and Groq, whose answers live
    up to \StdTtlMistral{}\,s. OpenAI's answers live at most
    \StdTtlOpenai{}\,s, and its connection was attributed for
    \StdTtlLastMaxOpenai{}\,s in the first repetition, ten seconds in the
    second, and not at all in the third, although an answer obtained just
    before the agent started still had 43\,s to live: the agent's own answer,
    obtained a second later, can come from another cache, closer to its
    expiry. Over the six minutes, \StdTtlAttributedAll{} of the
    \StdTtlScansAll{} scans attributed the connection. In a second phase, the
    agents' connections were closed two seconds after each call, as pooled
    HTTP clients do when idle, so that each new connection resolved the name
    again: \StdTtlReconnectAll{} of the \StdTtlReconnectOfAll{} scans that
    found an agent connected attributed its connection, including
    connections opened after the first entry had expired. The misses include
    connections opened while that entry was still valid, which points to
    connections that closed between our check and the scan rather than to
    the cache. For a periodic scan, a CDN-hosted agent is attributable only
    if some process resolved the provider's name within the last few
    minutes: an agent that holds one connection for an hour is attributable
    during its first minutes only, and an hourly scan will usually miss it.
    The episodes, scanned seconds after the agents start, measure the
    favourable case; the dedicated address ranges, which do not expire, are
    the robust signal, for the providers that publish them.
\end{itemize}

The renamed-runtime episode (A11) belongs to the main set but is also a
robustness test: a copy of the Node.js runtime renamed \code{updater.exe}
defeats D1, whose interpreter list is name-based, and the signature detector
D0. D2 and \AD{} still catch it, \AD{} through its connection and its shell,
which do not depend on the name of the executable. Its inspection of modules
and environment, by contrast, covers known interpreter names and processes
already connected to a known LLM service, so a renamed runtime that reaches
an unknown relay declared in a base-URL variable would escape it.

\subsection{RQ6: Behavioural baselines}\label{sec:eval-rq6}

\begin{figure}[t]
  \centering
  \begin{tikzpicture}
    \begin{axis}[vizaxis,
        width=0.62\linewidth, height=0.62\linewidth,
        xlabel={Recall}, ylabel={Precision},
        xmin=0, xmax=1.0, ymin=0, ymax=1.0,
        xtick={0,0.2,0.4,0.6,0.8,1.0}, ytick={0,0.2,0.4,0.6,0.8,1.0},
        legend columns=1,
        legend style={at={(0.5,-0.22)}, anchor=north}]
      \addplot[forget plot, vizMuted, densely dashed, line width=0.5pt, domain=0.3334:1, samples=60]
        {0.5*x/(2*x-0.5)} node[pos=1, anchor=west, font=\scriptsize, text=vizMuted] {$F_1=0.5$};
      \addplot[forget plot, vizMuted, densely dashed, line width=0.5pt, domain=0.5385:1, samples=60]
        {0.7*x/(2*x-0.7)} node[pos=1, anchor=west, font=\scriptsize, text=vizMuted] {$F_1=0.7$};
      \addplot[forget plot, vizMuted, densely dashed, line width=0.5pt, domain=0.8182:1, samples=60]
        {0.9*x/(2*x-0.9)} node[pos=1, anchor=west, font=\scriptsize, text=vizMuted] {$F_1=0.9$};
      \addplot[only marks, mark=*, mark size=3.2pt, color=vizBlue, mark options={draw=white, line width=1pt},
          error bars/.cd, x dir=both, y dir=both, x explicit, y explicit,
          error bar style={vizBlue, line width=0.7pt}, error mark=none]
        table[x=recall, y=precision, x error plus expr=\thisrow{rhi}-\thisrow{recall},
          x error minus expr=\thisrow{recall}-\thisrow{rlo}, y error plus expr=\thisrow{phi}-\thisrow{precision},
          y error minus expr=\thisrow{precision}-\thisrow{plo}] {eval/results/tex/pr-standard-D1.dat}
        node[pos=0, anchor=north west, font=\scriptsize, text=vizInk] {D1};
      \addlegendentry{D1: interpreter + port 443 + shell}
      \addplot[only marks, mark=*, mark size=3.2pt, color=vizOrange, mark options={draw=white, line width=1pt},
          error bars/.cd, x dir=both, y dir=both, x explicit, y explicit,
          error bar style={vizOrange, line width=0.7pt}, error mark=none]
        table[x=recall, y=precision, x error plus expr=\thisrow{rhi}-\thisrow{recall},
          x error minus expr=\thisrow{recall}-\thisrow{rlo}, y error plus expr=\thisrow{phi}-\thisrow{precision},
          y error minus expr=\thisrow{precision}-\thisrow{plo}] {eval/results/tex/pr-standard-D2.dat}
        node[pos=0, anchor=south west, font=\scriptsize, text=vizInk] {D2};
      \addlegendentry{D2: any process + port 443 + shell}
      \addplot[only marks, mark=*, mark size=3.2pt, color=vizAqua, mark options={draw=white, line width=1pt},
          error bars/.cd, x dir=both, y dir=both, x explicit, y explicit,
          error bar style={vizAqua, line width=0.7pt}, error mark=none]
        table[x=recall, y=precision, x error plus expr=\thisrow{rhi}-\thisrow{recall},
          x error minus expr=\thisrow{recall}-\thisrow{rlo}, y error plus expr=\thisrow{phi}-\thisrow{precision},
          y error minus expr=\thisrow{precision}-\thisrow{plo}] {eval/results/tex/pr-standard-AD.dat}
        node[pos=0, anchor=south east, font=\scriptsize, text=vizInk] {AD};
      \addlegendentry{\AD{} (multi-signal)}
    \end{axis}
  \end{tikzpicture}
  \caption{Precision and recall of the detectors on the agent episodes
    (standard user, majority outcome over \StdRuns{} runs), with exact 95\%
    confidence intervals (Clopper--Pearson). Dashed curves join the points of
    equal $F_1$. The intervals overlap: with \StdAgents{} agents and
    \StdBenign{} benign workloads, the bench shows where each detector fails
    rather than ranking them. The signature detector D0 flagged no episode,
    so its precision is undefined and it is not shown. Values are in
    \cref{tab:metrics}.}
  \label{fig:pr}
\end{figure}
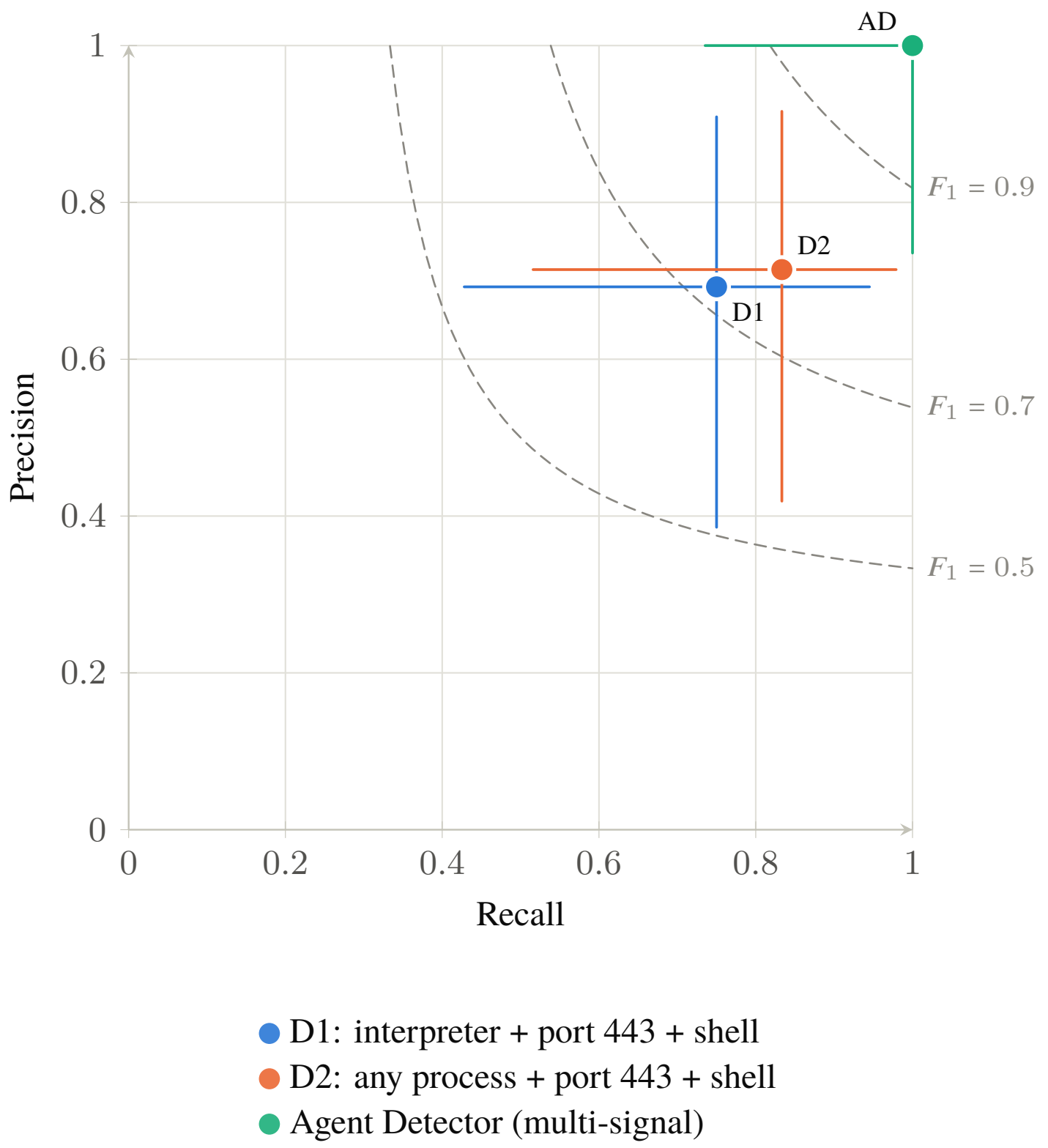

\Cref{tab:metrics} summarises the four detectors on the main episodes, and
\cref{fig:pr} places them in the precision--recall plane.

\begin{table}[t]
  \centering
  \caption{Detectors on the \StdAgents{} agent episodes and the
    \StdBenign{} benign episodes of the main set (standard user, majority
    outcome over \StdRuns{} runs), with exact 95\% confidence intervals.
    The last column counts the AI components (inference runtime, MCP server,
    LLM client without tools) that each detector flagged; they are not
    agents and are not counted in precision and recall.}
  \label{tab:metrics}
  \small\setlength{\tabcolsep}{4pt}
  \begin{tabular}{@{}l r r r r l l r c@{}}
\toprule
Detector & TP & FN & FP & TN & Precision [95\% CI] & Recall [95\% CI] & F1 & Components \\
\midrule
D0 signature & 0 & 12 & 0 & 9 & n/a & 0.00 {\scriptsize[0.00, 0.26]} & n/a & 0/3 \\
D1 interpreter rule & 9 & 3 & 4 & 5 & 0.69 {\scriptsize[0.39, 0.91]} & 0.75 {\scriptsize[0.43, 0.95]} & 0.72 & 1/3 \\
D2 any-process rule & 10 & 2 & 4 & 5 & 0.71 {\scriptsize[0.42, 0.92]} & 0.83 {\scriptsize[0.52, 0.98]} & 0.77 & 1/3 \\
Agent Detector & 12 & 0 & 0 & 9 & 1.00 {\scriptsize[0.74, 1.00]} & 1.00 {\scriptsize[0.74, 1.00]} & 1.00 & 3/3 \\
\bottomrule
\end{tabular}

\end{table}

The behavioural rules behave as \cref{sec:behavioral} predicts. They flag
every behavioural twin and \code{npm install}: \StdDOneFp{} of the
\StdBenign{} benign episodes for D1 and \StdDTwoFp{} for D2, for a precision
of \StdDOnePrecision{} and \StdDTwoPrecision{}. They also miss the agents that
do not fit the pattern: the local-model agent (no connection on port 443) and
the agent that drives a browser rather than a shell; among the components,
they flag only the MCP server, because \code{npx} downloads it over TLS and
starts it through \code{cmd.exe}. On the workstation, D2 also flagged
\code{code.exe}, the VS Code editor, whose integrated terminal runs shells
while the editor keeps TLS connections, \code{explorer.exe}, the Windows
shell, which also starts command prompts and holds TLS connections, and
\code{claude.exe}, a real coding agent: the
rule is right about the agent and wrong about the editor and the shell, for
the same reason. The signature detector D0, by construction, catches none of the
product-less agents, although it does find the real tools running on the
workstation.

\AD{} detected every agent in the majority of runs without flagging any
benign workload (precision \StdADPrecision{}, recall \StdADRecall{}, F1
\StdADFOne{}, against \StdDOneFOne{} and \StdDTwoFOne{} for the behavioural
rules). The twins differ from the agents only in the identity of the remote
service, which the behavioural rules do not see and \AD{} does. These
figures flatter \AD{} more than any other in this paper, for three reasons:
two of its detections held in only three runs out of five; the episodes
that once produced its errors (B09 and the CDN-hosted agents) were used to
correct it; and every episode is scanned in the favourable seconds after
the agent starts. The confidence intervals overlap widely, and precision
depends on how many twins the bench contains, so these numbers describe the
bench, not a population. What the bench does establish is qualitative: each
detector fails on identifiable classes of episodes, and the classes differ.
The structural rules cannot separate an agent from its twin; \AD{} cannot
see an agent whose service it cannot identify, such as the undeclared relay
of R01, and it sees a CDN-hosted agent only while the provider's name is in
the DNS cache.

\subsection{Administrator configuration}\label{sec:eval-admin}

An elevated administrator token reads what a standard user cannot on this
build: the BAM keys, the Prefetch folder, the processes of other accounts,
including services, and the profiles of every user. It is the closest
configuration to the production SYSTEM task that we could measure without
enrolling the machine, but not the same one: the scan ran in the
administrator's interactive session rather than in session 0, under another
account. We ran the whole bench \AdmRuns{} times in this configuration
(\cref{tab:admin}).

On the episodes, elevation changed little, as expected: the episodes run
under the user's own account and are visible at either level, and the
sources that elevation opens describe past executions and other accounts,
which the episodes do not exercise. \AD{} detected \AdmADTp{} of the
\AdmAgents{} agents and \AdmADComponents{} of the \AdmComponents{}
components, and flagged \AdmADFp{} benign workloads (precision
\AdmADPrecision{}, recall \AdmADRecall{}). The two differences with the
standard user are the timing effects of RQ1: the intermittent agent was
missed in every run, and the OpenAI agent was caught in every run. The
ephemeral agent R02 was still missed: Prefetch and BAM record that
\code{python.exe} ran, which identifies no agent.

Elevation changed the in-situ scan in two ways. First, the execution traces
confirmed what the log folder suggested: Claude Desktop had last run on
22~October, eighteen days before the scan, so its \emph{recent} level now
rests on a BAM entry. Second, the scan found two more findings at the
\emph{active} level, attributed to the SYSTEM account: the virtual-machine
service of Claude Desktop's Cowork mode and the sandbox service of the Codex
application, two Windows services that run whether or not anyone uses the
application. They open no notice, since the notice lists only the findings
of the session's user, but they show that \emph{active} means that a
process of the tool runs, not that a person uses it; a fleet report should
count service findings apart from user findings.

Elevation also costs little. A scan took \AdmDurationmsMedian{}\,ms at the
median (\AdmDurationmsPNineFive{}\,ms at the 95th percentile), against
\StdDurationmsMedian{}\,ms for the standard user, used
\AdmCputimemsMedian{}\,ms of CPU time (\StdCputimemsMedian{}\,ms) and peaked
at \AdmPeakworkingsetmbMedian{}\,MB (\StdPeakworkingsetmbMedian{}\,MB).

\begin{table}[t]
  \centering
  \caption{Detectors on the main episodes from an elevated administrator
    window (majority outcome over \AdmRuns{} runs), with exact 95\%
    confidence intervals; compare with \cref{tab:metrics}.}
  \label{tab:admin}
  \small\setlength{\tabcolsep}{4pt}
  \begin{tabular}{@{}l r r r r l l r c@{}}
\toprule
Detector & TP & FN & FP & TN & Precision [95\% CI] & Recall [95\% CI] & F1 & Components \\
\midrule
D0 signature & 0 & 12 & 0 & 9 & n/a & 0.00 {\scriptsize[0.00, 0.26]} & n/a & 0/3 \\
D1 interpreter rule & 9 & 3 & 4 & 5 & 0.69 {\scriptsize[0.39, 0.91]} & 0.75 {\scriptsize[0.43, 0.95]} & 0.72 & 1/3 \\
D2 any-process rule & 10 & 2 & 4 & 5 & 0.71 {\scriptsize[0.42, 0.92]} & 0.83 {\scriptsize[0.52, 0.98]} & 0.77 & 1/3 \\
Agent Detector & 11 & 1 & 0 & 9 & 1.00 {\scriptsize[0.72, 1.00]} & 0.92 {\scriptsize[0.62, 1.00]} & 0.96 & 3/3 \\
\bottomrule
\end{tabular}

\end{table}

\subsection{Threats to validity}\label{sec:eval-threats}

\paragraph{Internal validity.}
Synthetic agents reproduce the footprint that the detector observes, not the
full behaviour of real agents; they poll with invalid keys instead of
streaming completions, and all but one keep both of their signals present
all the time. Real agents are intermittent: \cref{sec:eval-rq5} measures,
for one synthetic pattern, how often a scan at a random moment sees such an
agent, but the share of time during which real agents hold an identifiable
connection, and a shell, remains to be measured on real sessions. The
baselines sample about once
per second and \AD{} three times per window, an asymmetry that favours the
baselines on transient patterns in the bench and would reverse in
production. The behavioural twins are constructed by us; they are a worst
case for structural rules, but a realistic one, as the \code{npm install}
and VS Code cases show. Each relay episode uses \code{httpbin.org} as a
stand-in relay. The in-situ cross-check reads the same kinds of sources as
the detector.

\paragraph{Privilege, time and production configuration.}
The elevated runs approach the SYSTEM task without reproducing it
(\cref{sec:eval-admin}). In both configurations, a connection attributed
through the DNS cache is never treated as dedicated: the \code{llm-api}
finding stays at low confidence, and detection rests on the
\code{probable-agent} rule, so an agent behind a CDN without a shell child
is reported below the notice threshold. The inversion of the DNS cache is
machine-wide, so a process that talks to an address of a shared CDN that
also served an LLM host is marked as LLM-connected; no benign workload of
the bench was affected, but few of them talk to the same CDNs, so this
precision is not established for a population. Above all, the episodes are
scanned seconds after the agents start, while their name resolutions are
still cached and their signals present: they measure the best case of a
periodic scan. The two experiments of \cref{sec:eval-rq5} measure how fast
this best case decays, on one machine and with synthetic agents.

\paragraph{External validity.}
One machine, one consumer edition of Windows on a recent build, one user,
and a small number of episodes per class: the results establish what each
layer can and cannot see, not rates that would hold across a fleet. The
confidence intervals of \cref{tab:metrics} overlap for all detectors.
Benign workloads on other machines (build farms, data-science notebooks,
administrators' automation, LLM pipelines that use several processes) will
produce other patterns.

\paragraph{Construct validity.}
Our operational definition of an agent and our detection threshold (medium
confidence) are choices, and \AD{}'s \code{probable-agent} rule (talks to a
model and has a shell descendant) is a narrower proxy of the definition.
Counting a finding of AI use, such as \code{llm-api}, as the detection of an
agent is generous to \AD{}: such a finding says that a process talks to a
model, not that it acts. \Cref{tab:episodes} lists the kinds of findings
behind each detection so that a reader can apply a stricter criterion.

\paragraph{Next step: a field study.}
The natural complement is a silent field study: \AD{} deployed in its silent
mode on a pilot population for a few weeks, reporting only aggregated counts
per tool, level and confidence, after the information and consultation steps
discussed in \cref{sec:discussion}. Such a study would measure what this
paper cannot: the prevalence of shadow agents, the share of product-less
agents among them, and the false-positive rate on a real population.

\section{Discussion}\label{sec:discussion}

\subsection{Inform rather than block}

\AD{} never blocks. This is a design decision, not a missing feature, and it
rests on three arguments. First, visibility must come before control: an
organisation that blocks what it cannot see pushes usage towards what it
cannot see, such as personal devices or web interfaces, and loses even the
partial view it had. Second, the agents of this paper are mostly used in good
faith (actor A1 of \cref{sec:threat}); for them, the most effective control is
often to learn that a policy exists, which approved alternatives it lists, and
whom to ask. Third, the tools that should block already exist: application
control can refuse to run an executable, and the network can refuse a domain.
A detector that feeds those controls with evidence is more useful than one
that duplicates them with less context.

The notice is the part of the system that turns detection into governance. It
informs the user at the moment of use, it records which version of the policy
text each user acknowledged, and it repeats as long as the agent remains, so
that acknowledgement documents awareness without granting an exemption. This
contrasts with silent discovery, which Microsoft's deployment guidance
describes in so many words: ``end users are not notified or
affected''~\cite{microsoft2026zta176}. Silent discovery gives the organisation
a list; it does not give the employee the information that would let them
comply, and it leaves the organisation unable to show that the employee was
informed.

The repetition has a cost. A user whose approved tool is installed sees the
notice, with the tool marked approved, at every scan; organisations will
likely want approved tools to be listed without re-opening the notice, or a
lower frequency once the user has acknowledged a given version of the text.
We kept the strictest behaviour in the evaluated version and report it as a
usability question for field studies.

\subsection{Legal and ethical framework}

A detector that observes employees' machines is a monitoring device, and it
must be designed and deployed as one. We discuss the European and French
frameworks, which are among the strictest; other jurisdictions differ.

\paragraph{Data protection.}
Under the GDPR, the processing needs a legal basis, typically the legitimate
interest of the employer in securing its information system
(Article~6(1)(f)), and it must respect data minimisation: data ``adequate,
relevant and limited to what is necessary in relation to the
purposes''~\cite[Art.~5(1)(c)]{eu2016gdpr}. The design choices of
\cref{sec:design} serve this principle directly: no content, no secret values,
the host of a base URL rather than the URL, tool names rather than activity
logs. Minimisation has a limit that we must state: an hourly finding of an
\emph{active} tool, per user, accumulates into a timeline of when that user
works with an agent. The retention period of the events, and who may query
them, are therefore part of the design of a deployment, not details.
Employees must be informed when the data are collected
(Article~13), and a data protection impact assessment is required where the
processing is likely to result in a high risk to their rights, with ``an
assessment of the necessity and proportionality''
(Article~35)~\cite{eu2016gdpr}; for a sensor that runs on every laptop every
hour, an organisation should assume that an assessment is needed. The legal
basis is not consent: between an employer and its employees, consent is
rarely free, and the acknowledgement that the notice collects is information,
not consent. Legitimate interest is, moreover, not available to public
authorities in the performance of their tasks, which need another basis. The
notice contributes to the information duty, since it tells users that
detections and acknowledgements are logged, but it does not replace the
general information of the workforce.

\paragraph{French labour law.}
In France, ``no information personally concerning an employee may be
collected by a device that has not been brought to their attention
beforehand''~\cite{france2008l12224}, and, in undertakings of at least fifty
employees, the works council (CSE) must be informed and consulted before the
introduction of ``means or techniques that allow the monitoring of employees'
activity''~\cite{france2018l231238}. More generally, any restriction of
employees' rights must be justified by the nature of the task and
proportionate to its aim (article L1121-1 of the Labour Code). The
French data protection authority adds that monitoring must not be excessive,
that permanent surveillance is generally considered an excessive
infringement, and that the employer must check that no less intrusive means
exists~\cite{cnil2026controle}. In our reading, these requirements favour the
design choices we made (collect identities of tools, not activity; inform at
the time of use; never record content) and they make the deployment process,
not the code, the place where compliance is achieved: policy update, worker
information and consultation, record of processing, impact assessment.

\paragraph{AI literacy.}
The AI Act initially required providers and deployers of AI systems to ``take
measures to ensure, to their best extent, a sufficient level of AI literacy of
their staff'', from 2 February 2025~\cite{eu2024aiact}. The
Digital Omnibus on AI of July 2026 replaced this with an obligation to ``take
measures to support the development of AI literacy'' of
staff~\cite{eu2026aiomnibus}. Either way, informing an employee, at the moment
they use an agent, of the organisation's AI policy and of the approved tools
is a measure of that kind. We do not claim that a notice satisfies the
obligation on its own, nor that an employer is the deployer of a tool that an
employee uses without authorisation; where it is not, the notice is good
practice rather than compliance.

\paragraph{Fleet reporting.}
Intune's documentation asks administrators not to use detection scripts ``to
collect personal data from devices''~\cite{microsoft2025remediations}. The
optional fleet report of \AD{} outputs tool names, not user names; but a
laptop usually belongs to one person, so the report should be treated as
personal data and included in the impact assessment.

\subsection{Operational lessons}

Building the detector was the smaller part of the work; packaging it for
Intune, and checking each step against the platform's documentation, taught
lessons that we believe generalise to any endpoint sensor for agents. They
come from building and testing the package, not from a measured fleet
deployment.

\begin{enumerate}
  \item \textbf{Privilege is coverage, but test the cause of each gap.} On the
    Windows build we tested, the BAM keys, the Prefetch folder and the
    processes of other users were not readable without administrator or
    SYSTEM rights; we have not checked whether this holds on other builds and
    editions. We first believed the same of the DNS client cache, because a
    standard-user scan found almost nothing in it. The cause was a lookup key
    (\cref{sec:attribution}), and the elevated run, which found nothing
    either, is what revealed it. The sensor should run as SYSTEM, and every
    report must state what it could not see, so that an absence of findings
    is never read as an absence of agents; but a coverage note is a
    hypothesis about its cause until it has been tested.
  \item \textbf{Session 0 cannot speak to users.} A SYSTEM process runs in an
    isolated session and cannot show a window. Informing the user requires
    starting a process in each interactive session with the user's token,
    which has a useful side effect: the acknowledgement is then recorded
    under the user's own identity. It records who acknowledged, but it does
    not prove it (\cref{sec:disclaimer-loop}).
  \item \textbf{The deployment agent may run in 32 bits.} The Intune
    management extension can start installation commands as 32-bit
    processes~\cite{microsoft2026addwin32}; an installer that does not relaunch
    itself in 64 bits writes its files and registry keys to redirected
    locations, and the application is never detected as installed.
  \item \textbf{Detect the installation, not the file.} Writing an
    installation marker last, whose value combines the binary version, the
    notice-text version and the mode, makes the MDM's detection rule reliable
    and turns every change of text or mode into an automatic reinstallation.
  \item \textbf{Signing is not enough under strict policies.} When group
    policy enforces the \code{AllSigned} PowerShell policy, scripts must be
    signed \emph{and} their publisher trusted, or PowerShell waits for a
    confirmation that a silent installation can never give.
  \item \textbf{Never trust a pre-created folder.} Standard users can create
    folders under \code{ProgramData}; a sensor that writes its reports there
    must recreate the folder with its own access control list.
  \item \textbf{Start silently.} A silent first phase, in which the sensor
    reports but shows nothing, measures the false positives of a real
    population before any user is interrupted, and lets the organisation
    complete its approved-tools list first.
  \item \textbf{Mind notice fatigue.} Repeating the notice at every scan keeps
    awareness high but also reminds users of tools that are already approved;
    the right frequency is a policy decision that deserves measurement.
  \item \textbf{Expect the catalog to drift.} Vendors rename products, move or
    share configuration folders, and add address ranges; new agents appear
    between two catalog releases. Personal agents such as OpenClaw, which
    security vendors describe as able to ``run shell commands, read and write
    files, and execute scripts''~\cite{chang2026openclaw} and which Microsoft
    lists among detected local agents~\cite{microsoft2026shadowai}, appeared
    after our catalog was designed; until a signature is added, only the
    generic rules can see them.
\end{enumerate}

\subsection{Limitations}

\paragraph{Privilege and periodicity.}
Coverage depends on privilege (\cref{sec:evaluation}): without administrator
rights, BAM and Prefetch are unreadable and the processes of other users
cannot be inspected. In production the scan runs as SYSTEM, but a
standard-user run is a realistic picture of what a non-privileged sensor
sees. Periodic scanning also
misses agents that start and stop between two scans; known tools leave
history, but a short script run by \code{python.exe} leaves only the trace of
\code{python.exe}. Continuous telemetry, for example from Event Tracing for
Windows, would close this gap at a higher cost.

\paragraph{The DNS cache.}
Attribution through the DNS client cache works only while the entries live,
at most one to five minutes for the CDN-hosted providers we measured
(\cref{fig:dnsttl}), and only for names resolved through the system resolver.
An agent that keeps one connection open for an hour is therefore attributable
during its first minutes only, and an hourly scan will usually miss it;
lengthening the time-to-live is not ours to decide, but a sensor could keep
its own record of recent resolutions, at the cost of running continuously.
Software that resolves names itself, or uses DNS over HTTPS internally,
escapes it. It also rests on two behaviours that Microsoft does not
document: the listing of the cache, which returned nothing on the build we
tested, and the keys under which resolvers store their entries, which we
inferred from experiments (\cref{sec:attribution}). Either may change with a
Windows release; a sensor that relies on them needs a regression test per
build, such as the bench of \cref{app:bench}.

\paragraph{Out-of-host execution.}
Agents in virtual machines, containers and WSL~2 run behind a virtual network
interface; their processes and sockets are not in the host's tables. Cloud
agents never touch the endpoint. These cases need hypervisor-, network- or
cloud-side controls.

\paragraph{Catalog drift.}
The generic rules reduce the dependence on the catalog, but network
attribution still needs to know which hosts are LLM services. While writing
this paper we found that an earlier version of the catalog lacked the IPv6
range that Anthropic documents for its API~\cite{anthropic2026ipaddresses},
and that the configuration folder used as residue evidence for the GitHub
Copilot command-line agent had become shared with the Copilot runtime of VS
Code.
Endpoint lists, configuration paths~\cite{microsoft2026vscodemcp} and product
names change monthly. A catalog-driven detector must be maintained like an
antivirus signature base, from vendor documentation, analyst catalogs and the
unknown relays reported by the fleet itself.

\paragraph{Evaluation scope.}
Our evaluation runs on one workstation, with synthetic agents that poll real
endpoints with invalid keys rather than performing real tasks, and with a
small set of benign workloads. Synthetic agents reproduce the footprint that
the detector observes, which is what is being measured, but not the full
diversity of real agents; real deployments will show other false positives.
\Cref{sec:eval-threats} describes the field study that would complement it.

\subsection{Evasion}

A deliberate evader (actor A4) can defeat endpoint detection: run the agent in
a virtual machine or container, reach the model through a relay on an
innocuous domain without any base-URL variable, rename the interpreter, run
for seconds at a time, or simply disable the scheduled task with local
administrator rights. We measured some of these cases (\cref{sec:evaluation})
and describe them only at the level of classes, which are well known. The
operational answers are outside the endpoint detector: remove local
administrator rights, alert on missing scan summaries (the absence of event
100 from a machine for more than a day is itself a signal), and combine
endpoint findings with network controls for relays and with application
control for virtualisation software.

\subsection{Beyond Windows}

The taxonomy and the correlation model do not depend on Windows. On macOS and
Linux, the same layers exist: process tables with command lines and
environments, socket tables, resolver caches where a local caching resolver
runs, package managers, application bundles and configuration files in the
user's home directory. Execution history is weaker on Linux and different on
macOS. Porting the collectors is engineering; the hypothesis that identity
must anchor behaviour carries over unchanged.

\section{Related Work}\label{sec:related}

\paragraph{Shadow IT and shadow AI.}
Information-systems research has studied shadow IT for over a decade: its
definition~\cite{silic2014shadow,haag2017shadow}, its governance
approaches~\cite{gyory2012exploring}, its nature and management in case
studies~\cite{zimmermann2017multiple}, its causes and outcomes in a systematic
review~\cite{klotz2019causing}, and what happens to shadow systems once they
are identified~\cite{furstenau2021leaving}. Recent work extends the concept to
generative AI~\cite{silic2025shadowai}, studies it in critical
infrastructure~\cite{baruwalchhetri2026frontier}, and models the governance of
``bring your own AI''~\cite{bello2026governing}. This literature is mostly
organisational and survey-based; it documents that shadow systems are
under-estimated and that identification is the precondition of governance, but
it does not provide the instrument that identifies them. Our work supplies
such an instrument for the agentic case, and our acceleration argument
(\cref{sec:acceleration}) connects the two literatures through the retirement
rate of shadow artifacts.

\paragraph{LLM agents and their ecosystem.}
The agent paradigm builds on reasoning-and-acting loops~\cite{yao2023react}
and tool use~\cite{schick2023toolformer}, surveyed in~\cite{wang2024survey}.
Software engineering benchmarks and agents~\cite{jimenez2024swebench,yang2024sweagent,wang2025openhands}
made coding agents practical, and empirical studies now measure them in
public repositories: their pull requests~\cite{li2025aiteammates,watanabe2025agenticprs},
their adoption~\cite{robbes2026agentic}, the configuration mechanisms
developers use to steer them~\cite{galster2026harness}, and the MCP tools they
call~\cite{stein2026mcptools}. These studies observe agents through the traces
they leave in code repositories. We observe them through the traces they leave
on the machines where they run, which is where an organisation's governance
applies and where agents that never commit code are visible.

\paragraph{Security of agents.}
A large body of work studies how agents can be attacked or misused: indirect
prompt injection~\cite{greshake2023indirect}, injection benchmarks for
tool-integrated agents~\cite{zhan2024injecagent,debenedetti2024agentdojo},
MCP-specific threats~\cite{hou2025mcplandscape,radosevich2025mcpsafety} and
tool poisoning~\cite{beurerkellner2025toolpoisoning,wang2026mcptox}.
Frameworks catalogue these risks for practitioners~\cite{owasp2024llmtop10,owasp2026llmtop10,owasp2025agentictop10,mitre2026atlas}
and for risk management~\cite{tabassi2023airmf,autio2024genaiprofile}. This
work secures agents that an organisation knows about. Ours addresses the prior
question of which agents are present at all; the two meet in the
configuration files that both attackers and defenders target.

\paragraph{Network-side observation of LLM traffic.}
Side-channel and traffic-analysis studies show that encrypted LLM traffic
reveals response contents~\cite{weiss2024prompt}, prompt
topics~\cite{mcdonald2025whisperleak}, the agent at
work~\cite{zhang2025agentprint}, and the serving model and
task~\cite{pouryousef2026llmtraffic}; TLS fingerprints characterise client
libraries~\cite{althouse2017ja3,althouse2023ja4}, while Encrypted Client Hello
reduces what the network sees~\cite{rescorla2026ech}. These works take an
attacker's or a network observer's vantage point and answer probabilistically.
We take the endpoint's vantage point, which attributes connections to
processes and users deterministically but sees only the machine it runs on.
The two are complementary, and \cref{sec:behavioral} explains why the
discriminating signal, in both cases, is the identity of the remote service.

\paragraph{Endpoint forensics and inventory.}
Digital forensics has long used Windows execution artifacts such as BAM and
Prefetch as evidence of program
execution~\cite{velociraptor2026bam,neyaz2022prefetch}, and recent work
documents the artifacts left by the ChatGPT Windows
application~\cite{kankanamge2025chatgpt}. EDR platforms maintain software
inventories~\cite{microsoft2026mdvminventory}. We reuse forensic artifacts as
one layer among six, and differ from inventories by covering software that is
not installed.

\paragraph{Vendor discovery of shadow AI.}
Cloud discovery catalogs classify generative-AI
services~\cite{microsoft2026dcacatalog}, data-centric posture management
tracks sensitive data flowing to AI applications~\cite{microsoft2026dspm}, and,
most recently, endpoint security now discovers named local agents and reports
them as shadow AI~\cite{microsoft2026localagents,microsoft2026shadowai}. To our
knowledge, the vendor documentation does not describe a mechanism for agents
without a product, and the discovery is designed to be silent for
users~\cite{microsoft2026zta176}. Continuous telemetry (Sysmon, EDR hunting
tables) and scanners of MCP configurations cover parts of the same ground
(\cref{sec:landscape})~\cite{microsoft2026sysmonevents,invariant2025mcpscan}.
Forensic studies have characterised the artifacts that individual AI
applications leave on a machine~\cite{kankanamge2025chatgpt}. We are not aware
of academic work that characterises the endpoint footprint of AI agents
across classes, compares behavioural and multi-signal detection on the same
bench, or couples detection with an acknowledgement loop; this paper is a
first step in that direction.

\paragraph{Visibility into agents.}
A line of work on AI governance asks how the deployers and users of agents
can see what agents do: agent identifiers, real-time monitoring and activity
logs~\cite{chan2024visibility}, agent infrastructure for attribution,
interaction and response~\cite{chan2025infrastructure}, authenticated
delegation of authority from a user to an agent~\cite{south2025authenticated},
and practices such as legibility, attributability and automatic
monitoring~\cite{shavit2023practices}. These proposals assume that agents
cooperate, by carrying an identifier or by running on infrastructure that
logs them. Shadow agents are, by definition, the ones that do not: they run
on a laptop, outside the deployer's infrastructure, often written by their
user. Endpoint detection is the fallback for this population, and the
footprint we measure is what an agent reveals without cooperating. The two
directions meet: an agent that declares itself, for example in its user
agent or in the client information of the Model Context Protocol, would make
the identity layer of our taxonomy reliable.

\section{Conclusion}\label{sec:conclusion}

Can unsanctioned agents be detected on enterprise endpoints? Our answer has
three parts.

\emph{Not by their products alone.} Signature and inventory approaches see the
agents that ship as products, and the products are many: our catalog lists 39
of them, and it was already incomplete by the time we wrote this paper. But
the agents that matter most for the acceleration of shadow IT are the ones
that ship as no product at all: scripts that call a model through an
aggregator, automations written by other agents, MCP servers started on the
fly. By construction, signatures miss them.

\emph{Not by simple behaviour.} An agent behaves like a developer, inside
processes that are legitimate. Its behavioural twin, the same program talking
to a package index instead of a model, is indistinguishable to any detector
that looks only at the structure of processes and connections. Our bench
illustrates the argument: two structural rules flagged every twin and an
ordinary \code{npm install}, and still missed agents that use local models
or drive browsers. Richer behavioural features remain to be tested.

\emph{Partly, when behaviour is anchored to identity.} Correlating what a
process does with the identity of the service it talks to, and completing
the picture with the configuration, installation and history that agents leave
behind, separates agents from their twins. It is not enough on its own, and
its limits are mostly about time. On our bench, \AD{} saw an agent behind a
shared content-delivery network only while the DNS cache held the
provider's name, a few minutes at most; it saw an intermittent agent at
about one scan in three; and it missed agents active too briefly for a
periodic scan, agents behind an undeclared relay and agents inside WSL~2.
The one false positive of an earlier version came from an inherited
configuration variable, and the bench showed how to remove it. A scan costs
about \StdDurationmsMedian{}\,ms and \StdPeakworkingsetmbMedian{}\,MB.
The bench is small, synthetic and run on one workstation; it maps what each
kind of evidence can and cannot see, and it does not estimate rates for a
fleet.

Detection is only the first step of governance. We have argued that it is a
precondition for the others: an organisation that cannot see agentic
artifacts cannot deliberately retire them, and if they reproduce faster than
they are retired, shadow IT can grow exponentially. Detection alone does not
bound that growth; sanctioned alternatives, least privilege and policies on
automations are needed too. Coupling detection with a notice that users
acknowledge informs users rather than observing them silently, and gives
organisations a record that their policies were communicated.

\paragraph{Future work.}
Five directions follow. Running real agents on a fixed set of tasks, and
recording their endpoint footprint over time, would replace our synthetic
agents and measure how often a periodic scan catches them. A field study in
silent mode would measure the prevalence of shadow agents and the share of
product-less agents on a real population. Continuous telemetry, for instance
from Event Tracing for Windows or Sysmon, would catch short-lived agents that
periodic scans miss, and keep the name resolutions that the DNS cache
forgets. Collectors
for macOS and Linux would test the claim that the taxonomy and the
correlation model are platform-independent. And standard ways for agents to
declare themselves, such as agent identities and registries, would turn part
of the detection problem into a disclosure protocol, leaving detection for the
agents that do not declare.

\paragraph{Availability.}
\AD{}, its catalog, the evaluation bench and the scripts that produce every
table and number of this paper are open source~\cite{canale2026repo}.

\bibliographystyle{plainnat}
\bibliography{refs}

\appendix
\section{The Evaluated Catalog}\label{app:catalog}

This appendix lists the content of the catalog evaluated in this paper
(version 2026.11.09). The tables are generated from the catalog embedded in
the evaluated binary, through the same loader as the detector, so that they
cannot drift from it.

\subsection{Tool signatures}

Each tool is recognised by one or more kinds of identifiers: process image
names, version-resource product names, Authenticode signer names, install
paths, command-line patterns, MSIX package families, services, uninstall
entries, IDE or browser extension identifiers, npm or pip packages, installed
files, history folders, residue folders, MCP configuration files, model
folders, listening ports and environment variables.

{\small
\begin{longtable}{@{}>{\raggedright\arraybackslash}p{2.8cm} >{\raggedright\arraybackslash}p{3.8cm} >{\raggedright\arraybackslash}p{2.0cm} >{\raggedright\arraybackslash}p{4.0cm}@{}}
\caption{Tool signatures of the evaluated catalog (version 2026.11.09).}\label{tab:app-tools}\\
\toprule Id & Tool & Category & Identifiers \\ \midrule
\endfirsthead
\toprule Id & Tool & Category & Identifiers \\ \midrule
\endhead
\bottomrule
\endfoot
\code{agent-\allowbreak{}frameworks} & AI agent framework (Python) & agent-\allowbreak{}framework & pip \\
\code{agent-\allowbreak{}sdks-\allowbreak{}node} & AI agent SDK (Node.js, global) & agent-\allowbreak{}framework & npm \\
\code{browser-\allowbreak{}use} & Browser Use & agent-\allowbreak{}framework & pip \\
\code{perplexity-\allowbreak{}comet} & Perplexity Comet (agentic browser) & agentic-\allowbreak{}browser & image, product, signer, path, uninstall, residue \\
\code{cursor} & Cursor & ai-\allowbreak{}ide & image, product, signer, path, cmdline, uninstall, files, history, residue, MCP config \\
\code{kiro} & Kiro & ai-\allowbreak{}ide & image, product, path, uninstall, files, residue, MCP config \\
\code{trae} & Trae & ai-\allowbreak{}ide & image, product, uninstall, residue \\
\code{windsurf} & Windsurf & ai-\allowbreak{}ide & image, product, path, uninstall, extension, files, history, residue, MCP config \\
\code{browser-\allowbreak{}ai-\allowbreak{}extension} & AI browser extension & browser-\allowbreak{}extension & extension \\
\code{aider} & Aider & cli-\allowbreak{}agent & image, cmdline, pip, residue \\
\code{claude-\allowbreak{}code} & Claude Code & cli-\allowbreak{}agent & image, product, signer, path, cmdline, extension, npm, files, history, residue, MCP config, env \\
\code{gemini-\allowbreak{}cli} & Gemini CLI & cli-\allowbreak{}agent & cmdline, npm, history, residue, MCP config \\
\code{github-\allowbreak{}copilot-\allowbreak{}cli} & GitHub Copilot CLI & cli-\allowbreak{}agent & cmdline, npm \\
\code{goose} & Goose & cli-\allowbreak{}agent & image, residue, env \\
\code{open-\allowbreak{}interpreter} & Open Interpreter & cli-\allowbreak{}agent & cmdline, pip \\
\code{openai-\allowbreak{}codex} & OpenAI Codex / ChatGPT app & cli-\allowbreak{}agent & image, product, signer, path, cmdline, MSIX, service, extension, npm, files, history, residue, env \\
\code{opencode} & OpenCode & cli-\allowbreak{}agent & image, product, signer, path, cmdline, uninstall, npm, files, history, residue \\
\code{claude-\allowbreak{}desktop} & Claude Desktop (incl. Cowork) & desktop-\allowbreak{}agent & path, MSIX, service, files, history, residue, MCP config \\
\code{chatgpt-\allowbreak{}desktop} & ChatGPT Classic desktop & desktop-\allowbreak{}assistant & product, signer, path, MSIX \\
\code{litellm} & LiteLLM proxy & gateway & cmdline, pip, port \\
\code{augment} & Augment Code & ide-\allowbreak{}agent & extension, residue \\
\code{cline} & Cline & ide-\allowbreak{}agent & extension \\
\code{continue} & Continue & ide-\allowbreak{}agent & extension, residue \\
\code{jetbrains-\allowbreak{}ai} & JetBrains AI Assistant / Junie & ide-\allowbreak{}agent & extension \\
\code{kilo-\allowbreak{}code} & Kilo Code & ide-\allowbreak{}agent & extension \\
\code{qodo} & Qodo Gen & ide-\allowbreak{}agent & extension \\
\code{roo-\allowbreak{}code} & Roo Code & ide-\allowbreak{}agent & extension \\
\code{vscode-\allowbreak{}agent-\allowbreak{}mcp} & VS Code agent mode (MCP servers configured) & ide-\allowbreak{}agent & MCP config \\
\code{amazon-\allowbreak{}q} & Amazon Q Developer & ide-\allowbreak{}assistant & extension \\
\code{gemini-\allowbreak{}code-\allowbreak{}assist} & Gemini Code Assist & ide-\allowbreak{}assistant & extension \\
\code{github-\allowbreak{}copilot} & GitHub Copilot (IDE) & ide-\allowbreak{}assistant & image, product, signer, extension \\
\code{tabnine} & Tabnine & ide-\allowbreak{}assistant & extension \\
\code{anythingllm} & AnythingLLM & local-\allowbreak{}llm & image, product, uninstall, residue, port \\
\code{gpt4all} & GPT4All & local-\allowbreak{}llm & path, uninstall, residue, port \\
\code{huggingface-\allowbreak{}models} & Hugging Face local models & local-\allowbreak{}llm & models \\
\code{jan} & Jan & local-\allowbreak{}llm & image, product, path, uninstall, residue, port \\
\code{llama-\allowbreak{}cpp} & llama.cpp / llamafile & local-\allowbreak{}llm & image, port \\
\code{lm-\allowbreak{}studio} & LM Studio & local-\allowbreak{}llm & image, product, path, uninstall, residue, models, port \\
\code{ollama} & Ollama & local-\allowbreak{}llm & image, path, uninstall, files, residue, models, port, env \\
\end{longtable}

}

\subsection{Generic patterns}

The generic rules of \cref{sec:correlation} rely on the following patterns of
the catalog.

{\small
{\sloppy
\begin{description}[leftmargin=1.2em,style=nextline]
  \item[Shells and interpreters counted as agent actions] \code{cmd.\allowbreak{}exe}, \code{powershell.\allowbreak{}exe}, \code{pwsh.\allowbreak{}exe}, \code{bash.\allowbreak{}exe}, \code{sh.\allowbreak{}exe}, \code{wsl.\allowbreak{}exe}, \code{git.\allowbreak{}exe}, \code{python.\allowbreak{}exe}, \code{python3.\allowbreak{}exe}, \code{node.\allowbreak{}exe}
  \item[Interpreters whose modules and environment are inspected] \code{python.\allowbreak{}exe}, \code{python3.\allowbreak{}exe}, \code{pythonw.\allowbreak{}exe}, \code{py.\allowbreak{}exe}, \code{node.\allowbreak{}exe}, \code{bun.\allowbreak{}exe}, \code{deno.\allowbreak{}exe}, \code{pwsh.\allowbreak{}exe}, \code{powershell.\allowbreak{}exe}, \code{java.\allowbreak{}exe}
  \item[Browsers (excluded from network attribution)] \code{chrome.\allowbreak{}exe}, \code{msedge.\allowbreak{}exe}, \code{brave.\allowbreak{}exe}, \code{firefox.\allowbreak{}exe}, \code{chromium.\allowbreak{}exe}, \code{chrome-\allowbreak{}headless-\allowbreak{}shell.\allowbreak{}exe}, \code{opera.\allowbreak{}exe}, \code{vivaldi.\allowbreak{}exe}
  \item[Browser automation flags] \code{-\allowbreak{}-\allowbreak{}remote-\allowbreak{}debugging-\allowbreak{}port}, \code{-\allowbreak{}-\allowbreak{}remote-\allowbreak{}debugging-\allowbreak{}pipe}, \code{-\allowbreak{}-\allowbreak{}enable-\allowbreak{}automation}, \code{-\allowbreak{}-\allowbreak{}headless}, \code{ms-\allowbreak{}playwright}, \code{puppeteer}, \code{-\allowbreak{}juggler}
  \item[MCP server command-line patterns] \code{\allowbreak{}(?i)@modelcontextprotocol[\textbackslash{}\allowbreak{}\textbackslash{}\allowbreak{}/\allowbreak{}]\allowbreak{}}, \code{\allowbreak{}(?i)\textbackslash{}\allowbreak{}bmcp-\allowbreak{}server-\allowbreak{}[a-\allowbreak{}z0-\allowbreak{}9-\allowbreak{}]\allowbreak{}+}, \code{\allowbreak{}(?i)\textbackslash{}\allowbreak{}bmcp-\allowbreak{}remote\textbackslash{}\allowbreak{}b}, \code{\allowbreak{}(?i)@playwright[\textbackslash{}\allowbreak{}\textbackslash{}\allowbreak{}/\allowbreak{}]\allowbreak{}mcp}, \code{\allowbreak{}(?i)chrome-\allowbreak{}devtools-\allowbreak{}mcp}, \code{\allowbreak{}(?i)\textbackslash{}\allowbreak{}buvx\allowbreak{}(\textbackslash{}\allowbreak{}.\allowbreak{}exe)?"?\textbackslash{}\allowbreak{}s+mcp-\allowbreak{}[a-\allowbreak{}z0-\allowbreak{}9-\allowbreak{}]\allowbreak{}+}
  \item[Inference module patterns] \code{\allowbreak{}(?i)\textasciicircum{}ggml[a-\allowbreak{}z0-\allowbreak{}9\_\allowbreak{}-\allowbreak{}]\allowbreak{}*\allowbreak{}\textbackslash{}\allowbreak{}.\allowbreak{}dll\$}, \code{\allowbreak{}(?i)\textasciicircum{}llama[a-\allowbreak{}z0-\allowbreak{}9\_\allowbreak{}-\allowbreak{}]\allowbreak{}*\allowbreak{}\textbackslash{}\allowbreak{}.\allowbreak{}dll\$}, \code{\allowbreak{}(?i)\textasciicircum{}onnxruntime-\allowbreak{}genai\textbackslash{}\allowbreak{}.\allowbreak{}dll\$}, \code{\allowbreak{}(?i)\textasciicircum{}mlc\_\allowbreak{}llm[a-\allowbreak{}z0-\allowbreak{}9\_\allowbreak{}-\allowbreak{}]\allowbreak{}*\allowbreak{}\textbackslash{}\allowbreak{}.\allowbreak{}dll\$}
  \item[Tokenizer module patterns] \code{\allowbreak{}(?i)\textasciicircum{}\_\allowbreak{}tiktoken[\textasciicircum{}\textbackslash{}\allowbreak{}\textbackslash{}\allowbreak{}]\allowbreak{}*\allowbreak{}\textbackslash{}\allowbreak{}.\allowbreak{}pyd\$}, \code{\allowbreak{}(?i)\textasciicircum{}tokenizers[\textasciicircum{}\textbackslash{}\allowbreak{}\textbackslash{}\allowbreak{}]\allowbreak{}*\allowbreak{}\textbackslash{}\allowbreak{}.\allowbreak{}pyd\$}
  \item[Base-URL variable patterns] \code{\allowbreak{}(?i)\textasciicircum{}\allowbreak{}(OPENAI|\allowbreak{}ANTHROPIC|\allowbreak{}AZURE\_\allowbreak{}OPENAI|\allowbreak{}GEMINI|\allowbreak{}GOOGLE\_\allowbreak{}GEMINI|\allowbreak{}GOOGLE\_\allowbreak{}GENAI|\allowbreak{}MISTRAL|\allowbreak{}GROQ|\allowbreak{}OPENROUTER|\allowbreak{}DEEPSEEK|\allowbreak{}XAI|\allowbreak{}TOGETHER|\allowbreak{}FIREWORKS|\allowbreak{}COHERE|\allowbreak{}PERPLEXITY|\allowbreak{}LITELLM|\allowbreak{}LLM|\allowbreak{}HF|\allowbreak{}HUGGINGFACE|\allowbreak{}AI\_\allowbreak{}GATEWAY|\allowbreak{}CLAUDE|\allowbreak{}CODEX)\_\allowbreak{}[A-\allowbreak{}Z0-\allowbreak{}9\_\allowbreak{}]\allowbreak{}*\allowbreak{}\allowbreak{}(BASE\_\allowbreak{}URL|\allowbreak{}API\_\allowbreak{}BASE|\allowbreak{}API\_\allowbreak{}URL|\allowbreak{}ENDPOINT)\$}
  \item[API-key variable patterns] \code{\allowbreak{}(?i)\textasciicircum{}\allowbreak{}(OPENAI|\allowbreak{}ANTHROPIC|\allowbreak{}AZURE\_\allowbreak{}OPENAI|\allowbreak{}GEMINI|\allowbreak{}GOOGLE\_\allowbreak{}GENAI|\allowbreak{}GOOGLE\_\allowbreak{}AI|\allowbreak{}MISTRAL|\allowbreak{}GROQ|\allowbreak{}OPENROUTER|\allowbreak{}DEEPSEEK|\allowbreak{}XAI|\allowbreak{}TOGETHER|\allowbreak{}FIREWORKS|\allowbreak{}COHERE|\allowbreak{}PERPLEXITY|\allowbreak{}HUGGINGFACE|\allowbreak{}HF|\allowbreak{}REPLICATE|\allowbreak{}CEREBRAS|\allowbreak{}POE|\allowbreak{}REQUESTY|\allowbreak{}PORTKEY|\allowbreak{}HELICONE|\allowbreak{}VOYAGE|\allowbreak{}AI\_\allowbreak{}GATEWAY|\allowbreak{}MOONSHOT|\allowbreak{}DASHSCOPE|\allowbreak{}ZHIPU|\allowbreak{}SILICONFLOW)\_\allowbreak{}\allowbreak{}(API\_\allowbreak{})?\allowbreak{}(KEY|\allowbreak{}TOKEN)\$}, \code{\allowbreak{}(?i)\textasciicircum{}CLAUDE\_\allowbreak{}CODE\_\allowbreak{}OAUTH\_\allowbreak{}TOKEN\$}
\end{description}
}

}

\subsection{Endpoints}

{\footnotesize
\begin{longtable}{@{}>{\raggedright\arraybackslash}p{2.4cm} >{\raggedright\arraybackslash}p{5.6cm} >{\raggedright\arraybackslash}p{5.0cm}@{}}
\caption{Endpoint entries of the evaluated catalog (version 2026.11.09). A suffix entry matches every subdomain; ``shared'' marks hosts that also serve non-AI traffic; ``unverified'' marks entries not yet confirmed on an official source when the catalog was frozen.}\label{tab:app-endpoints}\\
\toprule Category & Entry & Provider \\ \midrule
\endfirsthead
\toprule Category & Entry & Provider \\ \midrule
\endhead
\bottomrule
\endfoot
agent-infra & \code{*\allowbreak{}.\allowbreak{}e2b.\allowbreak{}app} & E2B sandboxes \\
agent-infra & \code{pattern \textasciicircum{}mcp\textbackslash{}\allowbreak{}.\allowbreak{}[a-\allowbreak{}z0-\allowbreak{}9.\allowbreak{}-\allowbreak{}]\allowbreak{}+\$} & Remote MCP server \\
agent-infra & \code{api.\allowbreak{}browser-\allowbreak{}use.\allowbreak{}com} & Browser Use Cloud \\
agent-infra & \code{api.\allowbreak{}browserbase.\allowbreak{}com} & Browserbase \\
agent-infra & \code{api.\allowbreak{}e2b.\allowbreak{}dev} & E2B sandboxes \\
agent-infra & \code{api.\allowbreak{}hyperbrowser.\allowbreak{}ai} & Hyperbrowser \\
agent-infra & \code{api.\allowbreak{}steel.\allowbreak{}dev} & Steel \\
agent-infra & \code{backend.\allowbreak{}composio.\allowbreak{}dev} & Composio \\
agent-infra & \code{connect.\allowbreak{}browserbase.\allowbreak{}com} & Browserbase \\
agent-infra & \code{mcp.\allowbreak{}zapier.\allowbreak{}com} & Zapier MCP \\
agent-infra & \code{server.\allowbreak{}smithery.\allowbreak{}ai} & Smithery hosted MCP \\
aggregator & \code{*\allowbreak{}.\allowbreak{}crazyrouter.\allowbreak{}com} & Crazyrouter \\
aggregator & \code{ai-\allowbreak{}gateway.\allowbreak{}kilo.\allowbreak{}ai} & Kilo gateway \\
aggregator & \code{aihubmix.\allowbreak{}com} & AIHubMix \\
aggregator & \code{api.\allowbreak{}302.\allowbreak{}ai} & 302.AI \\
aggregator & \code{api.\allowbreak{}aimlapi.\allowbreak{}com} & AI/ML API \\
aggregator & \code{api.\allowbreak{}aisa.\allowbreak{}one} & AIsa \\
aggregator & \code{api.\allowbreak{}cline.\allowbreak{}bot} & Cline provider \\
aggregator & \code{api.\allowbreak{}edenai.\allowbreak{}run} & Eden AI \\
aggregator & \code{api.\allowbreak{}kilo.\allowbreak{}ai} & Kilo gateway \\
aggregator & \code{api.\allowbreak{}kilocode.\allowbreak{}ai} & Kilo gateway (legacy name) \\
aggregator & \code{api.\allowbreak{}llmgateway.\allowbreak{}io} & LLM Gateway \\
aggregator & \code{api.\allowbreak{}notdiamond.\allowbreak{}ai} & Not Diamond \\
aggregator & \code{api.\allowbreak{}poe.\allowbreak{}com} & Poe API \\
aggregator & \code{api.\allowbreak{}puter.\allowbreak{}com} & Puter AI \\
aggregator & \code{api.\allowbreak{}unify.\allowbreak{}ai} & Unify \\
aggregator & \code{api.\allowbreak{}venice.\allowbreak{}ai} & Venice AI \\
aggregator & \code{api.\allowbreak{}withmartian.\allowbreak{}com} & Martian \\
aggregator & \code{gen.\allowbreak{}pollinations.\allowbreak{}ai} & Pollinations \\
aggregator & \code{glama.\allowbreak{}ai} & Glama \\
aggregator & \code{nano-\allowbreak{}gpt.\allowbreak{}com} & NanoGPT \\
aggregator & \code{opencode.\allowbreak{}ai} & OpenCode Zen \emph{(shared)} \\
aggregator & \code{openrouter.\allowbreak{}ai} & OpenRouter \\
aggregator & \code{router.\allowbreak{}requesty.\allowbreak{}ai} & Requesty \\
aggregator & \code{text.\allowbreak{}pollinations.\allowbreak{}ai} & Pollinations (legacy name) \\
aggregator & \code{withmartian.\allowbreak{}com} & Martian (legacy API base and website) \emph{(shared)} \\
dev-tool & \code{*\allowbreak{}.\allowbreak{}api.\allowbreak{}augmentcode.\allowbreak{}com} & Augment Code \\
dev-tool & \code{*\allowbreak{}.\allowbreak{}business.\allowbreak{}githubcopilot.\allowbreak{}com} & GitHub Copilot Business \\
dev-tool & \code{*\allowbreak{}.\allowbreak{}codeium.\allowbreak{}com} & Windsurf / Codeium \\
dev-tool & \code{*\allowbreak{}.\allowbreak{}codeiumdata.\allowbreak{}com} & Windsurf / Codeium \\
dev-tool & \code{*\allowbreak{}.\allowbreak{}cursor.\allowbreak{}sh} & Cursor \\
dev-tool & \code{*\allowbreak{}.\allowbreak{}devin.\allowbreak{}ai} & Devin / Windsurf \\
dev-tool & \code{*\allowbreak{}.\allowbreak{}enterprise.\allowbreak{}githubcopilot.\allowbreak{}com} & GitHub Copilot Enterprise \\
dev-tool & \code{*\allowbreak{}.\allowbreak{}factory.\allowbreak{}ai} & Factory \\
dev-tool & \code{*\allowbreak{}.\allowbreak{}githubcopilot.\allowbreak{}com} & GitHub Copilot \\
dev-tool & \code{*\allowbreak{}.\allowbreak{}individual.\allowbreak{}githubcopilot.\allowbreak{}com} & GitHub Copilot (personal plan) \\
dev-tool & \code{*\allowbreak{}.\allowbreak{}kiro.\allowbreak{}aws.\allowbreak{}dev} & Kiro \\
dev-tool & \code{*\allowbreak{}.\allowbreak{}kiro.\allowbreak{}dev} & Kiro \\
dev-tool & \code{*\allowbreak{}.\allowbreak{}mchost.\allowbreak{}guru} & Trae \\
dev-tool & \code{*\allowbreak{}.\allowbreak{}tabnine.\allowbreak{}com} & Tabnine \\
dev-tool & \code{*\allowbreak{}.\allowbreak{}trae.\allowbreak{}ai} & Trae \\
dev-tool & \code{*\allowbreak{}.\allowbreak{}traeapi.\allowbreak{}us} & Trae \\
dev-tool & \code{*\allowbreak{}.\allowbreak{}windsurf.\allowbreak{}com} & Windsurf \\
dev-tool & \code{*\allowbreak{}.\allowbreak{}zed.\allowbreak{}dev} & Zed AI \emph{(shared)} \\
dev-tool & \code{pattern \textasciicircum{}q\textbackslash{}\allowbreak{}.\allowbreak{}[a-\allowbreak{}z0-\allowbreak{}9-\allowbreak{}]\allowbreak{}+\textbackslash{}\allowbreak{}.\allowbreak{}amazonaws\textbackslash{}\allowbreak{}.\allowbreak{}com\$} & Amazon Q / Kiro \\
dev-tool & \code{ampcode.\allowbreak{}com} & Amp \\
dev-tool & \code{api.\allowbreak{}v0.\allowbreak{}dev} & v0 API \\
dev-tool & \code{copilot-\allowbreak{}proxy.\allowbreak{}githubusercontent.\allowbreak{}com} & GitHub Copilot \\
dev-tool & \code{cursor-\allowbreak{}cdn.\allowbreak{}com} & Cursor \\
dev-tool & \code{marketplace.\allowbreak{}cursorapi.\allowbreak{}com} & Cursor \\
gateway & \code{ai-\allowbreak{}gateway.\allowbreak{}helicone.\allowbreak{}ai} & Helicone AI Gateway \\
gateway & \code{ai-\allowbreak{}gateway.\allowbreak{}vercel.\allowbreak{}sh} & Vercel AI Gateway \\
gateway & \code{anthropic.\allowbreak{}helicone.\allowbreak{}ai} & Helicone (Anthropic proxy) \\
gateway & \code{api.\allowbreak{}keywordsai.\allowbreak{}co} & Keywords AI (legacy name) \\
gateway & \code{api.\allowbreak{}portkey.\allowbreak{}ai} & Portkey \\
gateway & \code{api.\allowbreak{}respan.\allowbreak{}ai} & Respan (formerly Keywords AI) \\
gateway & \code{gateway.\allowbreak{}ai.\allowbreak{}cloudflare.\allowbreak{}com} & Cloudflare AI Gateway \\
gateway & \code{oai.\allowbreak{}helicone.\allowbreak{}ai} & Helicone (OpenAI proxy) \\
inference-platform & \code{*\allowbreak{}.\allowbreak{}endpoints.\allowbreak{}huggingface.\allowbreak{}cloud} & Hugging Face Inference Endpoints \\
inference-platform & \code{*\allowbreak{}.\allowbreak{}lepton.\allowbreak{}run} & Lepton (NVIDIA DGX Cloud) \\
inference-platform & \code{*\allowbreak{}.\allowbreak{}modal.\allowbreak{}run} & Modal endpoints \emph{(shared)} \\
inference-platform & \code{api-\allowbreak{}inference.\allowbreak{}huggingface.\allowbreak{}co} & Hugging Face Inference API \\
inference-platform & \code{api.\allowbreak{}arliai.\allowbreak{}com} & ArliAI \\
inference-platform & \code{api.\allowbreak{}cerebras.\allowbreak{}ai} & Cerebras \\
inference-platform & \code{api.\allowbreak{}deepinfra.\allowbreak{}com} & DeepInfra \\
inference-platform & \code{api.\allowbreak{}featherless.\allowbreak{}ai} & Featherless \\
inference-platform & \code{api.\allowbreak{}fireworks.\allowbreak{}ai} & Fireworks AI \\
inference-platform & \code{api.\allowbreak{}friendli.\allowbreak{}ai} & FriendliAI \\
inference-platform & \code{api.\allowbreak{}groq.\allowbreak{}com} & Groq \\
inference-platform & \code{api.\allowbreak{}hyperbolic.\allowbreak{}xyz} & Hyperbolic \\
inference-platform & \code{api.\allowbreak{}inference.\allowbreak{}net} & Inference.net \\
inference-platform & \code{api.\allowbreak{}kluster.\allowbreak{}ai} & kluster.ai \\
inference-platform & \code{api.\allowbreak{}novita.\allowbreak{}ai} & Novita AI \\
inference-platform & \code{api.\allowbreak{}parasail.\allowbreak{}io} & Parasail \\
inference-platform & \code{api.\allowbreak{}replicate.\allowbreak{}com} & Replicate \\
inference-platform & \code{api.\allowbreak{}runpod.\allowbreak{}ai} & RunPod Serverless \\
inference-platform & \code{api.\allowbreak{}sambanova.\allowbreak{}ai} & SambaNova \\
inference-platform & \code{api.\allowbreak{}siliconflow.\allowbreak{}cn} & SiliconFlow \\
inference-platform & \code{api.\allowbreak{}siliconflow.\allowbreak{}com} & SiliconFlow \\
inference-platform & \code{api.\allowbreak{}studio.\allowbreak{}nebius.\allowbreak{}com} & Nebius AI Studio (legacy name) \\
inference-platform & \code{api.\allowbreak{}together.\allowbreak{}ai} & Together AI \\
inference-platform & \code{api.\allowbreak{}together.\allowbreak{}xyz} & Together AI \\
inference-platform & \code{api.\allowbreak{}tokenfactory.\allowbreak{}nebius.\allowbreak{}com} & Nebius Token Factory \\
inference-platform & \code{inference.\allowbreak{}baseten.\allowbreak{}co} & Baseten \\
inference-platform & \code{integrate.\allowbreak{}api.\allowbreak{}nvidia.\allowbreak{}com} & NVIDIA NIM API \\
inference-platform & \code{llm.\allowbreak{}chutes.\allowbreak{}ai} & Chutes \\
inference-platform & \code{router.\allowbreak{}huggingface.\allowbreak{}co} & Hugging Face Inference Providers \\
model-download & \code{huggingface.\allowbreak{}co} & Hugging Face \emph{(shared)} \\
model-download & \code{registry.\allowbreak{}ollama.\allowbreak{}ai} & Ollama model registry \\
model-provider & \code{*\allowbreak{}.\allowbreak{}cognitiveservices.\allowbreak{}azure.\allowbreak{}com} & Azure AI services \emph{(shared)} \\
model-provider & \code{*\allowbreak{}.\allowbreak{}openai.\allowbreak{}azure.\allowbreak{}com} & Azure OpenAI \\
model-provider & \code{*\allowbreak{}.\allowbreak{}services.\allowbreak{}ai.\allowbreak{}azure.\allowbreak{}com} & Azure AI Foundry \\
model-provider & \code{160.\allowbreak{}79.\allowbreak{}104.\allowbreak{}0/\allowbreak{}23} & Anthropic API (dedicated range) \\
model-provider & \code{2607:6bc0::/\allowbreak{}48} & Anthropic API (dedicated range) \\
model-provider & \code{pattern \textasciicircum{}[a-\allowbreak{}z0-\allowbreak{}9-\allowbreak{}]\allowbreak{}+-\allowbreak{}aiplatform\textbackslash{}\allowbreak{}.\allowbreak{}googleapis\textbackslash{}\allowbreak{}.\allowbreak{}com\$} & Google Vertex AI (regional) \emph{(shared)} \\
model-provider & \code{pattern \textasciicircum{}bedrock-\allowbreak{}\allowbreak{}(agent-\allowbreak{})?runtime\allowbreak{}(-\allowbreak{}fips)?\textbackslash{}\allowbreak{}.\allowbreak{}[a-\allowbreak{}z0-\allowbreak{}9-\allowbreak{}]\allowbreak{}+\textbackslash{}\allowbreak{}.\allowbreak{}amazonaws\textbackslash{}\allowbreak{}.\allowbreak{}com\$} & AWS Bedrock \\
model-provider & \code{aip.\allowbreak{}baidubce.\allowbreak{}com} & Baidu ERNIE \\
model-provider & \code{aiplatform.\allowbreak{}googleapis.\allowbreak{}com} & Google Vertex AI \emph{(shared)} \\
model-provider & \code{api.\allowbreak{}ai21.\allowbreak{}com} & AI21 API \\
model-provider & \code{api.\allowbreak{}aleph-\allowbreak{}alpha.\allowbreak{}com} & Aleph Alpha API \\
model-provider & \code{api.\allowbreak{}anthropic.\allowbreak{}com} & Anthropic API \\
model-provider & \code{api.\allowbreak{}baichuan-\allowbreak{}ai.\allowbreak{}com} & Baichuan API \\
model-provider & \code{api.\allowbreak{}cohere.\allowbreak{}ai} & Cohere API \\
model-provider & \code{api.\allowbreak{}cohere.\allowbreak{}com} & Cohere API \\
model-provider & \code{api.\allowbreak{}deepseek.\allowbreak{}com} & DeepSeek API \\
model-provider & \code{api.\allowbreak{}hunyuan.\allowbreak{}cloud.\allowbreak{}tencent.\allowbreak{}com} & Tencent Hunyuan API \\
model-provider & \code{api.\allowbreak{}inceptionlabs.\allowbreak{}ai} & Inception API \\
model-provider & \code{api.\allowbreak{}lingyiwanwu.\allowbreak{}com} & 01.AI API \\
model-provider & \code{api.\allowbreak{}llama.\allowbreak{}com} & Meta Llama API \\
model-provider & \code{api.\allowbreak{}minimax.\allowbreak{}io} & MiniMax API \\
model-provider & \code{api.\allowbreak{}minimaxi.\allowbreak{}com} & MiniMax API (China) \\
model-provider & \code{api.\allowbreak{}mistral.\allowbreak{}ai} & Mistral AI API \\
model-provider & \code{api.\allowbreak{}moonshot.\allowbreak{}ai} & Moonshot Kimi API \\
model-provider & \code{api.\allowbreak{}moonshot.\allowbreak{}cn} & Moonshot Kimi API \\
model-provider & \code{api.\allowbreak{}openai.\allowbreak{}com} & OpenAI API \\
model-provider & \code{api.\allowbreak{}perplexity.\allowbreak{}ai} & Perplexity API \\
model-provider & \code{api.\allowbreak{}reka.\allowbreak{}ai} & Reka API \\
model-provider & \code{api.\allowbreak{}sarvam.\allowbreak{}ai} & Sarvam API \\
model-provider & \code{api.\allowbreak{}stepfun.\allowbreak{}com} & StepFun API \\
model-provider & \code{api.\allowbreak{}upstage.\allowbreak{}ai} & Upstage API \\
model-provider & \code{api.\allowbreak{}writer.\allowbreak{}com} & Writer API \\
model-provider & \code{api.\allowbreak{}x.\allowbreak{}ai} & xAI API \\
model-provider & \code{api.\allowbreak{}z.\allowbreak{}ai} & Z.ai GLM API \\
model-provider & \code{ark.\allowbreak{}cn-\allowbreak{}beijing.\allowbreak{}volces.\allowbreak{}com} & ByteDance Doubao (Volcano Ark) \\
model-provider & \code{auth.\allowbreak{}openai.\allowbreak{}com} & OpenAI sign-in \\
model-provider & \code{chatgpt.\allowbreak{}com} & OpenAI ChatGPT / Codex backend \\
model-provider & \code{cloudcode-\allowbreak{}pa.\allowbreak{}googleapis.\allowbreak{}com} & Google Gemini Code Assist API \\
model-provider & \code{codestral.\allowbreak{}mistral.\allowbreak{}ai} & Mistral Codestral API \\
model-provider & \code{console.\allowbreak{}anthropic.\allowbreak{}com} & Anthropic Console \\
model-provider & \code{dashscope-\allowbreak{}intl.\allowbreak{}aliyuncs.\allowbreak{}com} & Alibaba Qwen (DashScope intl) \\
model-provider & \code{dashscope.\allowbreak{}aliyuncs.\allowbreak{}com} & Alibaba Qwen (DashScope) \\
model-provider & \code{generativelanguage.\allowbreak{}googleapis.\allowbreak{}com} & Google Gemini API \\
model-provider & \code{models.\allowbreak{}github.\allowbreak{}ai} & GitHub Models \\
model-provider & \code{models.\allowbreak{}inference.\allowbreak{}ai.\allowbreak{}azure.\allowbreak{}com} & GitHub Models (legacy) \\
model-provider & \code{open.\allowbreak{}bigmodel.\allowbreak{}cn} & Zhipu GLM API \\
model-provider & \code{platform.\allowbreak{}claude.\allowbreak{}com} & Anthropic Console \\
model-provider & \code{qianfan.\allowbreak{}baidubce.\allowbreak{}com} & Baidu ERNIE (Qianfan) \\
model-provider & \code{spark-\allowbreak{}api-\allowbreak{}open.\allowbreak{}xf-\allowbreak{}yun.\allowbreak{}com} & iFlytek Spark API \\
model-provider & \code{statsig.\allowbreak{}anthropic.\allowbreak{}com} & Anthropic telemetry \\
relay & \code{api.\allowbreak{}chatanywhere.\allowbreak{}org} & ChatAnywhere relay \\
relay & \code{api.\allowbreak{}chatanywhere.\allowbreak{}tech} & ChatAnywhere relay \\
relay & \code{api.\allowbreak{}ohmygpt.\allowbreak{}com} & OhMyGPT relay \\
relay & \code{api.\allowbreak{}openai-\allowbreak{}hk.\allowbreak{}com} & OpenAI-HK relay \\
relay & \code{apic1.\allowbreak{}ohmycdn.\allowbreak{}com} & OhMyGPT relay \\
relay & \code{c-\allowbreak{}z0-\allowbreak{}api-\allowbreak{}01.\allowbreak{}hash070.\allowbreak{}com} & OhMyGPT relay \\
relay & \code{oa.\allowbreak{}api2d.\allowbreak{}net} & API2D relay \\
web-assistant & \code{aistudio.\allowbreak{}google.\allowbreak{}com} & Google AI Studio \\
web-assistant & \code{bolt.\allowbreak{}new} & Bolt \\
web-assistant & \code{character.\allowbreak{}ai} & Character.AI \\
web-assistant & \code{chat.\allowbreak{}deepseek.\allowbreak{}com} & DeepSeek Chat \\
web-assistant & \code{chat.\allowbreak{}mistral.\allowbreak{}ai} & Mistral Le Chat \\
web-assistant & \code{chat.\allowbreak{}openai.\allowbreak{}com} & OpenAI ChatGPT \\
web-assistant & \code{chat.\allowbreak{}qwen.\allowbreak{}ai} & Qwen Chat \\
web-assistant & \code{chat.\allowbreak{}z.\allowbreak{}ai} & Z.ai Chat \\
web-assistant & \code{claude.\allowbreak{}ai} & Claude.ai \\
web-assistant & \code{copilot.\allowbreak{}microsoft.\allowbreak{}com} & Microsoft Copilot \\
web-assistant & \code{duck.\allowbreak{}ai} & Duck.ai \\
web-assistant & \code{gemini.\allowbreak{}google.\allowbreak{}com} & Google Gemini \\
web-assistant & \code{genspark.\allowbreak{}ai} & Genspark \\
web-assistant & \code{grok.\allowbreak{}com} & Grok \\
web-assistant & \code{jules.\allowbreak{}google.\allowbreak{}com} & Google Jules \\
web-assistant & \code{kimi.\allowbreak{}com} & Kimi \\
web-assistant & \code{lmarena.\allowbreak{}ai} & LMArena \\
web-assistant & \code{lovable.\allowbreak{}dev} & Lovable \\
web-assistant & \code{manus.\allowbreak{}im} & Manus \\
web-assistant & \code{meta.\allowbreak{}ai} & Meta AI \\
web-assistant & \code{notebooklm.\allowbreak{}google.\allowbreak{}com} & Google NotebookLM \\
web-assistant & \code{perplexity.\allowbreak{}ai} & Perplexity \\
web-assistant & \code{poe.\allowbreak{}com} & Poe \\
web-assistant & \code{t3.\allowbreak{}chat} & T3 Chat \\
web-assistant & \code{typingmind.\allowbreak{}com} & TypingMind \\
web-assistant & \code{v0.\allowbreak{}app} & v0 \\
web-assistant & \code{www.\allowbreak{}perplexity.\allowbreak{}ai} & Perplexity \\
web-assistant & \code{you.\allowbreak{}com} & You.com \\
\end{longtable}
}

\section{Report Format}\label{app:formats}

\AD{} writes a JSON report whose findings carry their evidence. The excerpt
below is a finding produced during the evaluation, for the synthetic Python
agent of episode A02, with standard-user rights.

\IfFileExists{eval/results/tex/app-finding.tex}{
{\footnotesize
\begin{verbatim}
{
  "tool_id": "probable-agent:python.exe",
  "tool_name": "Probable custom AI agent (python.exe)",
  "category": "custom-agent",
  "level": "active",
  "confidence": "medium",
  "approved": false,
  "user_sid": "S-1-5-21-...",
  "user": "HOST\\user",
  "evidence": [
    {
      "kind": "behavior",
      "detail": "python.exe pid 15656 talks to an LLM and runs cmd.exe",
      "path": "C:\\Python311\\python.exe",
      "pid": 15656,
      "cmdline": "\"C:\\Python311\\python.exe\" C:\\Users\\user\\Documents\\detectorAg
          ent\\paper\\eval\\agents\\agent.py --endpoint https://api.anthropic.com/v1/m
          odels --shell cmd",
      "level": "active",
      "confidence": "medium"
    }
  ]
}
\end{verbatim}
}
}{}

Each Windows event of \cref{tab:events} carries a JSON message. The summary
event (100) contains the product and catalog versions, the host, the account,
the coverage notes, the summary counters and a compact list of all findings;
a finding event (300) contains one finding with its evidence; the
acknowledgement event (410) contains the version of the notice text and the
list of tools, and is written under the identity of the user who accepted.

\section{Default Notice Text}\label{app:disclaimer}

The default text below is a starting point that organisations replace through
the branding file; it must be validated by the organisation's legal counsel
and data protection officer before use. An earlier version told users that
they remain responsible for any action an agent takes on their computer; we
removed that sentence because it may misstate the law of some countries: in
France, for instance, an employee's liability towards the employer is
limited, and the employer answers to third parties for its employees' acts.

\begin{quote}\small
\textbf{An AI agent was detected on your computer.}\\[4pt]
\emph{Detected tools: (list, each marked ``approved'' or ``not approved'')}\\[4pt]
An AI agent can read your files, run commands, and send data to external
services. Its use is governed by the AI policy of \emph{(company)}:
\begin{itemize}
  \item Only use approved tools.
  \item Never share confidential, personal, or customer data with a tool that
    is not approved.
  \item If you are not sure that a tool is approved, ask IT Security before
    using it.
\end{itemize}
This detection and your acceptance are logged by IT Security.\\[4pt]
\fbox{\ } I have read and understood the AI acceptable use policy.
\hfill \fbox{I accept}
\end{quote}

\section{Reproducing the Evaluation}\label{app:bench}

The evaluation bench lives in the \code{paper/eval} folder of the
repository~\cite{canale2026repo}.

\begin{description}[leftmargin=1.2em,style=nextline]
  \item[\code{agents/}] The synthetic agents (\code{agent.py},
    \code{agent.js}, \code{agent.ps1}). They poll real endpoints with an
    invalid key and keep a persistent shell or start one shell per call;
    options add an MCP server, an automated browser, a local LLM-style
    server, an inference-like module or idle connection closing.
  \item[\code{workloads/dev-loop.ps1}] The benign developer workloads (npm,
    git, pip, .NET, Go).
  \item[\code{run-episodes.ps1}] Runs the episodes, the baselines (defined in
    \code{lib.ps1}) and three \AD{} scans per episode, once or several
    times; each run records its provenance in \code{run.json}.
  \item[\code{bench-overhead.ps1}] Runs 100 consecutive scans with cost
    metrics.
  \item[\code{intermittent.ps1}] Runs the intermittent agent of episode A12
    and scans it at random intervals.
  \item[\code{dns-ttl.ps1}] Measures how long connections to CDN-hosted
    providers stay attributable through the DNS cache. It runs alone, at
    least five minutes after any other part of the bench, because a recent
    resolution of the same names would refresh the entries it measures.
  \item[\code{insitu.ps1}, \code{inventory.ps1}, \code{ground-truth.csv}] The
    in-situ cross-check and its expected findings.
  \item[\code{score.py}] Scores everything and writes every table and number
    of \cref{sec:evaluation} as LaTeX files.
  \item[\code{model.py}] Computes the illustrative trajectories of
    \cref{fig:acceleration}.
  \item[\code{run-all.ps1}] Runs the whole bench for the current privilege
    level, with five runs of the episodes by default: started from a
    standard PowerShell window, it reproduces the standard-user results of
    this paper; started from an administrator window, the administrator
    results, which approach the production SYSTEM task
    (\cref{sec:eval-admin}). It refuses to start while processes of an
    interrupted run are still alive.
\end{description}

\end{document}